%% file: main.tex
\documentclass{article}

\usepackage[final, nonatbib]{neurips_2020}

\input{preamble}

\usepackage{ifthen}
\newcounter{SEPARATE_APPENDIX}
\ifthenelse{\value{SEPARATE_APPENDIX} > 0}{
    \myexternaldocument{appendix}
}{}

\title{\papertitle}
\input{author}
\begin{document}

\maketitle
\input{source/abstract}
\input{source/main_content}
\input{source/acknowledgement}
\bibliography{main}

\ifthenelse{\value{SEPARATE_APPENDIX} > 0}{}{
    \newpage
    \input{source/appendix_content}
}

\end{document}

%% file: preamble.tex
\usepackage[]{multicol}
\usepackage{caption}
\usepackage{wrapfig}

\usepackage[utf8]{inputenc} 
\usepackage[T1]{fontenc}    
\usepackage{url}            
\usepackage{booktabs}       
\usepackage{amsfonts}       
\usepackage{nicefrac}       
\usepackage{microtype}      

\usepackage{algorithm}
\usepackage{algorithmic}
\usepackage{color}

\usepackage{amsthm}
\usepackage{amsmath}
\usepackage{amssymb}

\usepackage[pdftex]{graphicx}
\usepackage[toc,page]{appendix}

\newcommand{\Vtrain}{{V_{\mathrm{train}}}}
\newcommand{\Vtest}{{V_{\mathrm{test}}}}
\newcommand{\F}{{\mathrm{F}}}
\newcommand{\op}{{\mathrm{op}}}

\newtheorem{proposition}{Proposition}
\newtheorem{definition}{Definition}
\newtheorem{lemma}{Lemma}
\newtheorem{theorem}{Theorem}
\newtheorem{remark}{Remark}

\newtheorem{assumption}{Assumption}

\newcommand{\papertitle}{{Optimization and Generalization Analysis of Transduction through Gradient Boosting and Application to Multi-scale Graph Neural Networks}}

\allowdisplaybreaks

\input{math_commands}

\usepackage{xr-hyper}
\makeatletter
\newcommand*{\addFileDependency}[1]{
  \typeout{(#1)}
  \@addtofilelist{#1}
  \IfFileExists{#1}{}{\typeout{No file #1.}}
}
\makeatother

\newcommand*{\myexternaldocument}[1]{%
    \externaldocument{#1}%
    \addFileDependency{#1.tex}%
    \addFileDependency{#1.aux}%
}
\usepackage{hyperref}       

%% file: math_commands.tex
\usepackage{amsmath,amsfonts,bm}

\def\eqref#1{Equation~(\ref{#1})}

\def\1{\bm{1}}

\DeclareMathAlphabet{\mathsfit}{\encodingdefault}{\sfdefault}{m}{sl}
\SetMathAlphabet{\mathsfit}{bold}{\encodingdefault}{\sfdefault}{bx}{n}

\newcommand{\E}{\mathbb{E}}

\newcommand{\sigmoid}{\sigma}

\DeclareMathOperator*{\argmax}{arg\,max}
\DeclareMathOperator*{\argmin}{arg\,min}

\DeclareMathOperator{\sign}{sign}

%% file: author.tex
\author{%
  Kenta Oono \\
  The University of Tokyo \\
  Preferred Networks, Inc. \\
  Tokyo, Japan \\
  \texttt{kenta\_oono@mist.i.u-tokyo.ac.jp}
  \And
  Taiji Suzuki \\
  The University of Tokyo \\
  RIKEN Center for Advanced Intelligence Project \\
  Tokyo, Japan\\
  \texttt{taiji@mist.i.u-tokyo.ac.jp}
}

%% file: source/abstract.tex
\renewcommand*{\thefootnote}{\fnsymbol{footnote}}
\begin{abstract}
It is known that the current graph neural networks (GNNs) are difficult to make themselves deep due to the problem known as \textit{over-smoothing}. Multi-scale GNNs are a promising approach for mitigating the over-smoothing problem. However, there is little explanation of why it works empirically from the viewpoint of learning theory. In this study, we derive the optimization and generalization guarantees of transductive learning algorithms that include multi-scale GNNs. Using the boosting theory, we prove the convergence of the training error under weak learning-type conditions. By combining it with generalization gap bounds in terms of transductive Rademacher complexity, we show that a test error bound of a specific type of multi-scale GNNs that decreases corresponding to the number of node aggregations under some conditions. Our results offer theoretical explanations for the effectiveness of the multi-scale structure against the over-smoothing problem. We apply boosting algorithms to the training of multi-scale GNNs for real-world node prediction tasks. We confirm that its performance is comparable to existing GNNs, and the practical behaviors are consistent with theoretical observations. Code is available at \url{https://github.com/delta2323/GB-GNN}
\footnote{Kenta Oono conducted this study at the University of Tokyo.}.
\end{abstract}
\setcounter{footnote}{0}
\renewcommand*{\thefootnote}{\arabic{footnote}}

%% file: source/main_content.tex
\input{source/main/introduction}
\input{source/main/related_work}
\input{source/main/preliminary}
\input{source/main/analysis}

\input{source/main/example}
\input{source/main/practical_consideration}
\input{source/main/experiment}
\input{source/main/discussion}
\input{source/main/conclusion}
\input{source/main/broader_impact}

%% file: source/main/introduction.tex
\section{Introduction}

Graph neural networks (GNNs)~\cite{gori2005new,scarselli2009graph} are an emerging deep learning model for analyzing graph structured-data.
They have achieved state-of-the-art performances in node prediction tasks on a graph in various fields such as biochemistry~\cite{NIPS2015_5954}, computer vision~\cite{yang2018graph}, and knowledge graph analysis~\cite{schlichtkrull2018modeling}.
While they are promising, the current design of GNNs has witnessed a challenge known as \textit{over-smoothing}~\cite{AAAI1816098,Oono2020Graph}.
Typically, a GNN iteratively aggregates and mixes node representations of a graph~\cite{gilmer17a,kipf2017iclr,velickovic2018graph}.
Although it can capture the subgraph information using local operations only, it smoothens the representations and makes them become indistinguishable (over-smoothen) between nodes as we stack too many layers, leading to underfitting of the model.
Several studies suspected that this is the cause of the performance degradation of deep GNNs and devised methods to mitigate it~\cite{Rong2020DropEdge:,zhao2020pairnorm}.
Among others, multi-scale GNNs~\cite{liao2018lanczosnet,nguyen2017semi,pmlr-v80-xu18c} are a promising approach as a solution for the over-smoothing problem.
These models are designed to combine the subgraph information at various scales, for example, by bypassing the output of the middle layers of a GNN to the final layer.  

Although multi-scale GNNs empirically have resolved the over-smoothing problem to some extent, little is known how it works theoretically.
To justify the empirical performance from the viewpoint of statistical learning theory, we need to analyze two factors: \textit{generalization gap} and \textit{optimization}.
There are several studies to guarantee the generalization gaps~\cite{NIPS2019_8809,garg2020generalization,NIPS2018_8076,scarselli2018vapnik,verma2019stability}.
However, to the best of our knowledge, few studies have provided optimization guarantees.
The difficulty partly originates owing to the inter-dependency of predictions. That is, the prediction for a node depends on the neighboring nodes, as well as its feature vector. It prevents us from extending the optimization theory for inductive learning settings to transductive ones.

In this study, we propose the analysis of multi-scale GNNs through the lens of the boosting theory~\cite{pmlr-v80-huang18b,pmlr-v80-nitanda18a}.
Our idea is to separate a model into two types of functions -- aggregation functions $\mathcal{G}$ that mix the representations of nodes and transformation functions $\mathcal{B}$, typically common to all nodes, that convert the representations to predictions.
Accordingly, we can interpret a multi-scale GNN as an ensemble of supervised models and incorporate analysis tools of inductive settings.
We first consider our model in full generality and prove that as long as the model satisfies the \textit{weak learning condition} (w.l.c.), which is a standard type of assumption in the boosting theory, it converges to the global optimum.
By combining it with the evaluation of the transductive version of Rademacher complexity~\cite{el2009transductive}, we give a sufficient condition under which a particular type of multi-scale GNNs has the upper bound of test errors that decreases with respect to depth (the number of node aggregation operations) under the w.l.c.
This is in contrast to usual GNNs suffering from the over-smoothing problem.
Finally, we apply multi-scale GNNs trained with boosting algorithms, termed \textit{Gradient Boosting Graph Neural Network} (GB-GNN), to node prediction tasks on standard benchmark datasets.
We confirm that our algorithm can perform favorably compared with state-of-the-art GNNs, and our theoretical observations are consistent with the practical behaviors.

The contributions of this study can be summarized as follows:
\begin{itemize}
    \item We propose the analysis of transductive learning models via the boosting theory and derive the optimization and generalization guarantees under the w.l.c. (Theorem~\ref{thm:optimization}, Proposition \ref{prop:generalization-gap}).
    \item As a special case, we give the test error bound of a particular type of multi-scale GNNs that monotonically decreases with respect to the number of node aggregations (Theorem~\ref{thm:monotonically-decreasing-test-error}).
    \item We apply GB-GNNs, GNNs trained with boosting algorithms, to node prediction tasks on real-world datasets. We confirm that GB-GNNs perform favorably compared with state-of-the-art GNNs, and theoretical observations are consistent with the empirical behaviors. 
\end{itemize}

\paragraph{Notation}

$\mathbb{N}_+$ denotes the set of non-negative integers.
For $N\in \mathbb{N}_+$, we define $[N]:= \{1, \ldots, N\}$.
For a proposition $P$, $\bm{1}\{P\}$ equals $1$ when $P$ is true and $0$ otherwise.
For $a, b\in \mathbb{R}$, we denote $a\wedge b := \min(a, b)$ and $a\vee b := \max(a, b)$.
For a vector $u, v\in \mathbb{R}^N$ and $p\geq 1$, we denote the $p$-norm of $v$ by $\|v\|_p^p := \sum_{n=1}^N v_n^{p}$ and the Kronecker product by $u\otimes v$.
All vectors are column vectors.
For a matrix $X, Y\in \mathbb{R}^{N\times C}$ and $p\geq 1$, we define the inner product by $\langle X, Y\rangle:=\sum_{n=1}^N\sum_{c=1}^CX_{nc}Y_{nc}$, $(2, p)$-norm of $X$ by 
$\|X\|_{2, p}^p := \sum_{c=1}^C \left(\sum_{n=1}^N X_{nc}^2\right)^{\frac{p}{2}}$, the Frobenius norm by $\|X\|_\F := \|X\|_{2, 2}$, and the operator norm by $\|X\|_{\op} := \sup_{v\in \mathbb{R}^C, \|v\|_2=1} \|Xv\|_2$.
For $y\in \{0, 1\}$, we write $y^{\sharp} := 2y - 1 \in \{\pm 1\}$.
For $a\in \mathbb{R}$, we define $\sign(a) = 1$ if $a \geq 0$ and $-1$ otherwise.

%% file: source/main/related_work.tex
\section{Related Work}

\paragraph{Graph-based Transductive Learning Algorithms}

Graph-based transductive learning algorithms operate on a graph given \textit{a priori} or constructed from the representations of samples. For example, spectral graph transducer~\cite{joachims2003transductive} and the algorithm proposed in~\cite{belkin2004regularization} considered the regularization defined by the graph Laplacian. Another example is the label propagation algorithm~\cite{NIPS2003_2506}, which propagates label information through a graph. The extension of label propagation to deep models achieved state-of-the-art prediction accuracy in semi-supervised tasks appeared in computer vision~\cite{Iscen_2019_CVPR}. Recently, GNNs~\cite{gori2005new,scarselli2009graph} have been used to solve node prediction problems as a transductive learning task, where each sample point is represented as a node on a graph, and the goal is to predict the properties of the nodes. GNNs, especially MPNN-type (message passing neural networks) GNNs~\cite{gilmer17a}, differ from the aforementioned classical transductive learning algorithms because it mixes representations of sample points directly thorough the underlying graph.

\paragraph{Over-smoothing and Multi-scale GNNs}

Multi-scale GNNs~\cite{mixhop,ngcn,busch2020pushnet,liao2018lanczosnet,NIPS2019_9276,nguyen2017semi,pmlr-v80-xu18c} are a promising approach for mitigating the over-smoothing problem using the information of subgraphs at various scales.
For example, the Jumping Knowledge Network ~\cite{pmlr-v80-xu18c} were intentionally designed to solve the over-smoothing problem by aggregating the outputs of the intermediate layers to the final layer. 
However, to the best of our knowledge, there is no theoretical explanation of why multi-scale GNNs can perform well against the over-smoothing problem.
We proved that a specific instantiation of our model has a test error bound that monotonically decreases with respect to depth, thereby providing the evidence for the architectural superiority of multi-scale GNNs for the over-smoothing problem.

\paragraph{Boosting Interpretation of Deep Models}

Boosting~\cite{freund1995boosting,schapire1990strength} is a type of ensemble method for combining several learners to create a more accurate one.
For example, gradient boosting~\cite{friedman2000additive,NIPS1999_1766} is a de-facto boosting algorithm owing to its superior practical performance and easy-to-use libraries~\cite{Chen:2016:XST:2939672.2939785,NIPS2017_6907,NIPS2018_7898}.
Reference~\cite{NIPS2016_6556} interpreted Residual Network (ResNet)~\cite{he2016deep} as a collection of relatively shallow networks.
References~\cite{pmlr-v80-huang18b,pmlr-v80-nitanda18a} gave another interpretation as an ensemble model and evaluated its theoretical optimization and generalization performance.
In particular,~\cite{pmlr-v80-nitanda18a} employed the notion of (functional) gradient boosting.
Similar to these studies, we interpret a GNN as an ensemble model to derive the optimization and generalization guarantees.

AdaGCN (AdaBoosting graph convolutional network), which has been recently proposed by~\cite{sun2019adagcn}, is the closest to our study. They interpreted a multi-scale GCN (graph convolutional network)~\cite{kipf2017iclr} as an ensemble model and trained it using AdaBoost~\cite{freund1995desicion}. Although their research demonstrated the practical superiority of the boosting approach, we would argue that there is room for exploration in their theory. For example, they used the Vapnik–-Chervonenkis (VC) dimension to evaluate the generalization gap. However, it is known that the VC dimension cannot explain the empirical behaviors of AdaBoost~\cite{schapire1998boosting} (see also~\cite[Section 7.3]{mohri2018foundations}). Besides, they did not give optimization guarantees of AdaGCNs. In contrast, our primary goal is to devise methodologies for multi-scale GNNs with a solid theoretical backbone. To realize it, we tackle the non-i.i.d.\ nature of node prediction tasks and derive the optimization and refined generalization guarantees.

\paragraph{Generalization Analysis of GNNs in Transductive Settings}

It is not trivial to define the appropriate notion of generalization in a transductive learning setting because we do not need to consider the prediction accuracy of sample points that are not in a given dataset.
We define the generalization gap as of discrepancy between the training and test errors in terms of the random partition of a full dataset into training and test datasets\cite{el2009transductive,vapnik1982estimation} (see Section~\ref{sec:generalization} for the precise definition). This definition can admit the dependency between sample points.
Furthermore,~\cite{pechyony2009theory} showed that any generalization gap bound in this setting is automatically translated to the bound of the corresponding i.i.d.\ setting.
We employed the transductive version of Rademacher complexity, introduced by~\cite{el2009transductive} to bound generalization gaps.
Similarly to supervised settings, we have the transductive version of model complexities such as the VC dimension and variants of Rademacher complexity~\cite{pmlr-v35-tolstikhin14,tolstikhin2015permutational}.
We also have transductive PAC-Bayes bounds~\cite{pmlr-v33-begin14} and stability-based bounds~\cite{cortes2008stability,el2006stable} for generalization analysis.
Although several research have studied generalization of GNNs~\cite{NIPS2019_8809,garg2020generalization,NIPS2018_8076,scarselli2018vapnik,verma2019stability}, to the best of our knowledge, none of them satisfies for our purpose (see Section ~\ref{sec:more-related-work}).

%% file: source/main/preliminary.tex
\section{Problem Settings}

\input{source/main/preliminary/problem_setting}
\input{source/main/preliminary/functional_gradient_boosting}
\input{source/main/preliminary/algorithm}

%% file: source/main/preliminary/problem_setting.tex
\paragraph{Transductive Learning}\label{sec:problem-setting}

Let $\mathcal{X}$ and $\mathcal{Y}$ be spaces of feature vectors and labels, respectively.
Let $N\in \mathbb{N}_+$ be the sample size and $V:=[N]$ be the set of indices of the sample.
For each sample point $i\in V$, we associate a feature-label pair $(x_i, y_i)\in \mathcal{X} \times \mathcal{Y}$.
Let $\Vtrain$ and $\Vtest$ be the set of training and test samples, respectively, satisfying $\Vtrain \cap \Vtest = \emptyset$ and $\Vtrain \cup \Vtest = V$.
We denote the training and test sample sizes by $M:=|\Vtrain|$ and $U:=|\Vtest|$, respectively.
Given the collection of features $X=(x_i)_{i\in V}$ and labels $(y_i)_{i\in \Vtrain}$ for the training data, the task is to construct a predictor $h:\mathcal{X}\to \widehat{\mathcal{Y}}$ such that $h(x_i)$ is \textit{close} to $y_i$ for all $i\in \Vtest$ (we define it precisely in Section \ref{sec:optimization}).
Here, $\widehat{\mathcal{Y}}$ denotes the range of the predictor.
For later use, we define $Q := \frac{1}{M} + \frac{1}{U}$.

%% file: source/main/preliminary/functional_gradient_boosting.tex
\paragraph{Gradient Boosting}

We briefly present an overview of the gradient boosting method~\cite{friedman2001greedy,NIPS1999_1766}, which is also called the restricted gradient descent~\cite{ICML2011Grubb_626}.
Let $\mathcal{H}$ be a subset of Hilbert space (e.g., a collection of predictors).
Given a functional $\mathcal{L}: \mathcal{H} \to \mathbb{R}$ (e.g., training error), we want to find the minima of $\mathcal{L}$.
Gradient boosting solves this problem by iteratively updating the predictor $h^{(t)}\in\mathcal{H}$ at each iteration $t$ by adding a weak learner $f^{(t)}$ near the steepest direction of $\mathcal{L}$.
Although a general theory can admit that $\mathcal{H}$ is infinite-dimensional (known as \textit{functional} gradient boosting), it is sufficient for our purpose to assume that $\mathcal{H}$ is finite-dimensional.
Let $\mathcal{F}^{(t)}\subset \mathcal{H}$ a hypothesis space of weak learners at iteration $t\in \mathbb{N}_+$.
Gradient boosting attempts to find $f^{(t)} \in \mathcal{F}^{(t)}$ such that
$f^{(t)} \in \argmin_{f\in \mathcal{F}^{(t)}} d(-\nabla \mathcal{L}(h^{(t)}), f)$
holds true, and the step size $\eta^{(t)} > 0$.
Here, $d$ is some distance on $\mathcal{H}$ and $\nabla \mathcal{L}(h)$ is the (Fr\'{e}chet) derivative of $\mathcal{L}$ at $h$.
We update the predictor by
    $h^{(t+1)} = h^{(t)} + \eta^{(t)} f^{(t)}$.
Because we cannot solve the minimization problem above exactly in most cases, we resort to an approximated algorithm that can find the solution near the optimal one (corresponding to Definition \ref{def:weak_learnability_condition} below in our setting).
Several boosting algorithms such as AdaBoost, Arc-x4~\cite{breiman1998arcing}, Confidence Boost~\cite{schapire1999improved}, and Logit Boost~\cite{friedman2000additive} fall into this formulation by appropriately selecting $\mathcal{L}$, $d$ and $\eta^{(t)}$'s~\cite{NIPS1999_1766}.

%% file: source/main/preliminary/algorithm.tex
\paragraph{Models}

\input{image/schematic_view}

Figure \ref{fig:schematic-view} shows a schematic view of the model considered in this study.
It consists of two types of components: \textit{aggregation} functions $g^{(t)}: \mathcal{X}^N\to \mathcal{X}^N$ that mix the representations of sample points and \textit{transformation} functions $b^{(t)}:\mathcal{X}^N \to \widehat{\mathcal{Y}}^N$ that make predictions from representations.
We specify a model by defining the set of aggregation and transformation functions at each iteration $t$, denoted by $\mathcal{G}^{(t)}$ and $\mathcal{B}^{(t)}$, respectively. If we use the same function classes $\mathcal{G}^{(t)}$ and $\mathcal{B}^{(t)}$ for all $t$, we shall omit the superscript $(t)$.
Typically, a transformation function $b\in \mathcal{B}^{(t)}$ is a broadcast of the same function, i.e., $b$ is of the form $b = (b_0, \ldots, b_0)$ for some $b_0:\mathcal{X}\to \widehat{\mathcal{Y}}$.
However, we do not assume this until necessary.
We define the hypothesis space $\mathcal{F}^{(t)}$ at the $t$-th iteration by
    $\mathcal{F}^{(t)} := \{b^{(t)}\circ g^{(t)} \circ \cdots \circ g^{(1)} \mid b^{(t)} \in \mathcal{B}^{(t)}, g^{(s)}\in \mathcal{G}^{(s)} (s \in [t])\}.$
Given $g^{(s)}\in \mathcal{G}^{(s)}$ selected at the $s=1, \ldots, t-1$ iterations, we choose $g^{(t)}\in \mathcal{G}^{(t)}$ and $b^{(t)}\in\mathcal{B}^{(t)}$ to construct a weak learner $f^{(t)}(X) := b^{(t)}(g^{(t)}(X^{(t-1)}))$ and update the representation $X^{(t)}:=g^{(t)}(X^{(t-1)})$. When $t=1$, we define $f^{(1)}(X) := b^{(1)}(X)$ and $X^{(1)} := X$. We do not select $g^{(1)}$, nor do we update the representation.
Algorithm \ref{alg:train} shows the overall training algorithm.

%% file: image/schematic_view.tex
\begin{figure}
  \centering
  \includegraphics[width=0.7\linewidth]{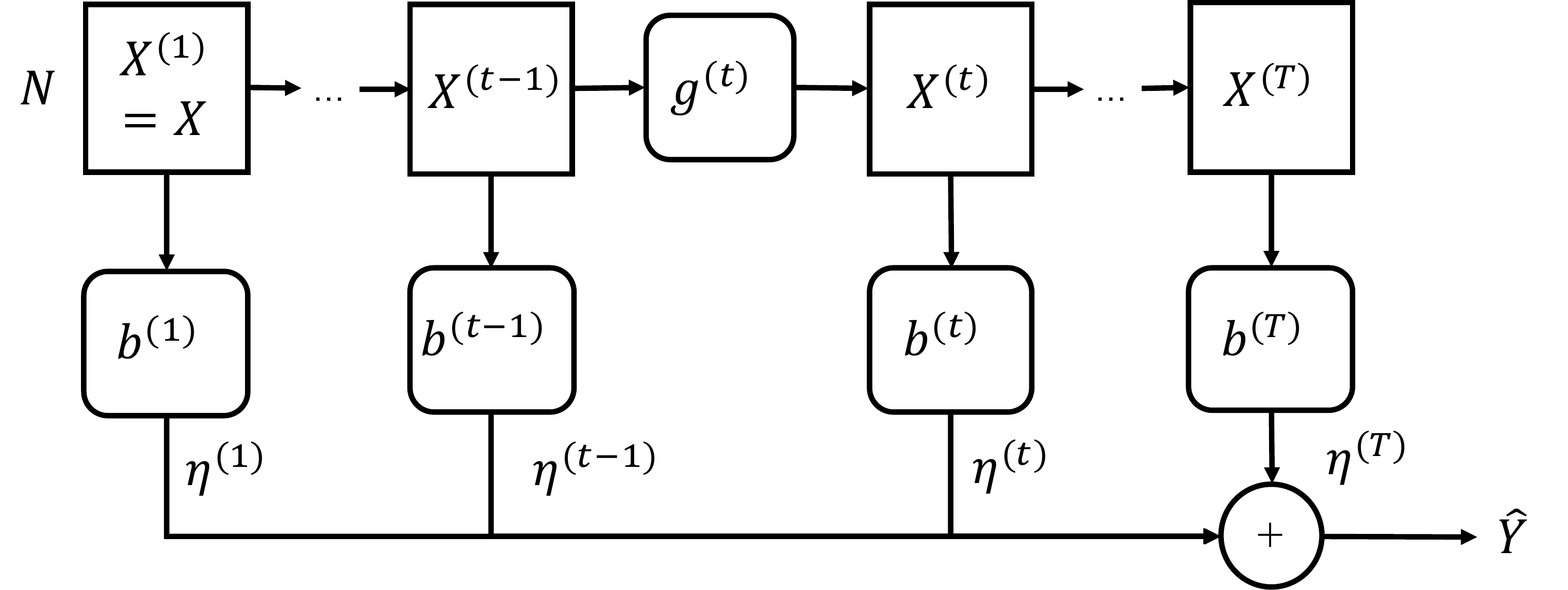}
  \caption{Schematic view of the model. $g^{(t)}:\mathcal{X}^N\to\mathcal{X}^N\in\mathcal{G}^{(t)}$ and $b^{(t)}:\mathcal{X}^N\to\widehat{\mathcal{Y}}^N\in \mathcal{B}^{(t)}$ are aggregation and transformation functions, respectively, and $\eta^{(t)}$ is the learning rate at the $t$-th iteration. We assume $\widehat{\mathcal{Y}}=\mathbb{R}$ in Sections~\ref{sec:analysis} and~\ref{sec:application-to-multi-scale-gnns} and $\mathcal{X}=\mathbb{R}^C$ in Section~\ref{sec:application-to-multi-scale-gnns}.}\label{fig:schematic-view}
\end{figure}

%% file: source/main/analysis.tex
\section{Analysis}\label{sec:analysis}
\input{source/main/analysis/optimization}
\input{table/proposed_algorithm}
\input{source/main/analysis/generalization}

%% file: source/main/analysis/optimization.tex
\subsection{Optimization}\label{sec:optimization}

In this section, we focus on a binary classification problem.
Accordingly, we set $\mathcal{Y} = \{0, 1\}$ and $\widehat{\mathcal{Y}} = \mathbb{R}$.
For $\delta \geq 0$, we define
$\ell_\delta(\hat{y}, y):=\bm{1}[(2p - 1) y^{\sharp} < \delta]$,
where
$p=\mathrm{sigmoid}(\hat{y})=(1 + \exp(-\hat{y}))^{-1}$.
Note that $\ell_{\delta=0}$ is the 0--1 loss.
Because it is difficult to optimize $\ell_\delta$, we define the sigmoid cross entropy loss $\ell_\sigma(\hat{y}, y) := -y \log p-(1-y)\log (1-p)$ as a surrogate function.
For a predictor $h: \mathcal{X} \to \widehat{\mathcal{Y}}$, we define the test error by 
$\mathcal{R}(h) := \frac{1}{U} \sum_{n\in \Vtest} \ell_{\delta}(h(x_n), y_n)$,
and training errors by
$\widehat{\mathcal{R}}(h) := \frac{1}{M} \sum_{n\in \Vtrain} \ell_{\delta}(h(x_n), y_n)$ and
$\widehat{\mathcal{L}}(h) := \frac{1}{M} \sum_{n\in \Vtrain} \ell_{\sigma}(h(x_n), y_n)$.
Because it is sufficient to make predictions of given samples, the values of a predictor outside of the samples do not affect the problem.
Therefore, we can and do identify a predictor $h$ with a vector $\widehat{Y}:=(h(x_1), \ldots, h(x_N))^\top \in \widehat{\mathcal{Y}}^N$.
Accordingly, we represent $\mathcal{R}(\widehat{Y}):= \mathcal{R}(h)$ (same is true for other errors).
Similarly to previous studies~\cite{ICML2011Grubb_626,pmlr-v80-nitanda18a}, we assume the following learnability condition to obtain the optimization guarantee.
\begin{definition}[Weak Learning Condition]\label{def:weak_learnability_condition}
    Let $\alpha > \beta \geq 0$, and $\bm{g}\in \mathbb{R}^N$, we say $Z\in \mathbb{R}^N$ satisfies $(\alpha, \beta, \bm{g})$-weak learning condition (w.l.c.) if it satisfies
        $\left\|Z - \alpha \bm{g}\right\|_{2} \leq \beta \|\bm{g}\|_{2}$.
    We say a weak learner $f:\mathcal{X}^N\to\widehat{\mathcal{Y}}^N=\mathbb{R}^N$ satisfies $(\alpha, \beta, \bm{g})$-w.l.c.\ when $f(X)$ does.
\end{definition}
The following proposition provides a handy way to check the empirical satisfiability of w.l.c. It ensures that the weak learner and negative gradient face the same ``direction".
Using the arugment similar to~\cite{ICML2011Grubb_626}, we show that our w.l.c.\ is equivalent to the AdaBoost-style learnability condition~\cite{pmlr-v80-huang18b}.
\begin{proposition}\label{prop:equivalence-of-wlc}
    Let $Z, \bm{g}\in \mathbb{R}^N$ such that $\bm{g}\not =0$.
    There exists $\alpha > \beta \geq 0$ such that $Z$ satisfies $(\alpha, \beta, \bm{g})$-w.l.c.\ if and only if $\langle Z, \bm{g}\rangle > 0$.
    Further, when $Z\in \{\pm 1\}^N$, this is equivalent to the condition that there exists $\delta \in (0, 1]$ such that $\sum_{n=1}^N w_n\bm{1}\{\sign(\bm{g}_n) \not = Z_n\} \leq \frac{1 - \delta}{2}$ where $w_n = \frac{\bm{g}_n}{\|\bm{g}\|_1}$.
\end{proposition}
See Section~\ref{sec:proof-equivalence-of-wlc} for the proof.
Under the condition, we have the following optimization guarantee.
\begin{theorem}\label{thm:optimization}
    Let $T\in \mathbb{N}_+$ and $\alpha_t > \beta_t \geq0$ ($t\in [T]$).
    Define $\gamma_t := \frac{\alpha_t^2 - \beta_t^2}{\alpha_t^2}$ and $\Gamma_T := \sum_{t=1}^T \gamma_t$.
    If Algorithm~\ref{alg:train} finds a weak learner $f^{(t)}$ for any $t\in [T]$, its output $\widehat{Y} \in \widehat{\mathcal{Y}}^N$ satisfies
        $\widehat{\mathcal{R}}(\widehat{Y})
        \leq  \frac{(1 + e^{\delta})\widehat{\mathcal{L}}(\widehat{Y}^{(1)})}{2M\Gamma_T}$.
    In particular, when $\gamma_t$ is independent of $t$, the right hand side is $O(1/T)$.
\end{theorem}
Note that $\gamma_t$ is the lower bound of the cosine value of the angle between $f^{(t)}(X)$ and $-\nabla \widehat{\mathcal{L}}(\widehat{Y})$.
The proof strategy is similar to that of \cite[Theorem 1]{pmlr-v80-nitanda18a} in that we bound the gradient of the training loss (Lemma~\ref{lem:small-gradient}) and apply a Kurdyka-Łojasiewicz-like inequality (Lemma~\ref{lem:bound-error-by-gradient}).
See Section \ref{sec:proof-of-optimization} for the proof.
We shall confirm that the w.l.c.\ holds empirically in the experiments in Section~\ref{sec:experiment} and discuss the provable satisfiability of the w.l.c.\ in Section~\ref{sec:discussion}.

%% file: table/proposed_algorithm.tex
\begin{algorithm}[t]
\caption{Training Algorithm}
\begin{algorithmic}\label{alg:train}
    \INPUT Features $X\in \mathcal{X}^N$. Labels $y_i \in \mathcal{Y}$ ($i\in \Vtrain$). $\#$iterations $T$. w.l.c.\ params $(\alpha_t, \beta_t)_{t\in [T]}$.
    \OUTPUT A collection of predictions $\widehat{Y}\in \widehat{\mathcal{Y}}^{N}$ for all sample points.
    \STATE Find $b^{(1)} \in \mathcal{B}^{(1)}$ and $\eta^{(1)}>0$.
    \STATE $X^{(1)} \leftarrow X$. 
    \STATE $\widehat{Y}^{(1)} \leftarrow \eta^{(1)}b^{(1)}(X)$.
    \FOR{$t = 2 \ldots T+1$}
        \STATE Find $b^{(t)}\in \mathcal{B}^{(t)}$ and $g^{(t)}\in \mathcal{G}^{(t)}$ and set $f^{(t)}(X)\leftarrow b^{(t)}(g^{(t)}(X^{(t-1)}))$.
        \STATE Ensure $\frac{1}{M}f^{(t)}$ satisfies $(\alpha_t, \beta_t, -\nabla\widehat{\mathcal{L}}(\widehat{Y}^{(t-1)}))$-w.l.c.
        \STATE $X^{(t)}  \leftarrow g^{(t)}(X^{(t-1)})$.
        \STATE $\widehat{Y}^{(t)} \leftarrow \widehat{Y}^{(t-1)} + \eta^{(t)} f^{(t)}(X)$.
    \ENDFOR
    \STATE $t^\ast = \argmin_{t\in [T-1]} \|\nabla \widehat{L}(\widehat{Y}^{(t)})\|_{2, 1}$.
    \STATE $\widehat{Y} \leftarrow \widehat{Y}^{(t^\ast)}$.
\end{algorithmic}
\end{algorithm}

%% file: source/main/analysis/generalization.tex
\subsection{Generalization}\label{sec:generalization}

We follow the problem setting of~\cite{el2009transductive}.
For fixed $M\in\mathbb{N}_+$, we create a training set by uniformly randomly drawing $M$ sample points \textit{without} replacement from $V$ and treating the remaining $U$ sample points as a test set.
We think of training and test errors as random variables with respect to the random partition of $V$.
Reference \cite{el2009transductive} introduced the Rademacher complexity for transductive learning and derived the generalization gap bounds.
We obtain the following proposition by applying it to our setting.
For a hypothesis space $\mathcal{F}\subset \{\mathcal{X}\to \widehat{\mathcal{Y}}\}$, we denote its transductive Rademacher complexity by $\mathfrak{R}(\mathcal{F})$.
We define $S:=\frac{4(M+U)(M\wedge U)}{(2(M+U)-1)(2(M\wedge U)-1)}$, which is close to $1$ when $M$ and $U$ are sufficiently large.
See Section~\ref{sec:proof-of-generalization} for the definition of $\mathfrak{R}(\cdot)$ and the proof of the proposition.
\begin{proposition}\label{prop:generalization-gap}
    There exists a universal constant $c_0 > 0$ such that for any $\delta' > 0$, with a probability of at least $1-\delta'$ over the random partition of samples, the output $\hat{Y}$ of Algorithm~\ref{alg:train} satisfies
    \begin{align*}
        \mathcal{R}(\widehat{Y})
        \leq \widehat{\mathcal{R}}(\widehat{Y}) + \sum_{t=1}^T \eta^{(t)}\mathfrak{R}(\mathcal{F}^{(t)})
        + c_0Q\sqrt{M\wedge U} + \sqrt{\frac{SQ}{2}\log \frac{1}{\delta'}}.
    \end{align*}
\end{proposition}

%% file: source/main/example.tex
\section{Application to Multi-scale GNNs}\label{sec:application-to-multi-scale-gnns}

We shall specialize our model and derive a test error bound for multi-scale GNNs that is monotonically decreasing with respect to $T$. In later sections, we assume $\mathcal{X}= \mathbb{R}^C$ for some $C\in \mathbb{N}_+$.
We continue to assume that $\mathcal{Y}=\{0, 1\}$ and $\widehat{\mathcal{Y}}= \mathbb{R}$.
First, we specialize $\mathcal{B}^{(t)}$ as a parallel application of the same transformation function for a single sample point of the form
\begin{equation}\label{eq:parallel-transformation}
    \mathcal{B}^{(t)} := \{(f_\mathrm{base}, \ldots f_\mathrm{base})^{\top} \mid f_\mathrm{base} \in \mathcal{B}^{(t)}_\mathrm{base} \}
\end{equation}
for some $\mathcal{B}^{(t)}_{\mathrm{base}}\subset \{\mathcal{X} \to \widehat{\mathcal{Y}}\}$.
By assuming this structure, we can evaluate the transductive Rademacher complexity in a similar way to the inductive case.
We take multi layer perceptrons (MLPs) as base functions for an example (see Proposition~\ref{prop:bound-by-inductive-rademacher-complexity} for the general case).
Let $L\in \mathbb{N}_+$ $\bm{C} = (C_1, \ldots, C_{L+1})\in \mathbb{N}_+^{L+1}$ and $B>0$ such that $C_1 = C$ and $C_{L+1}=1$.
We define $\mathcal{B}^{(t)}_\mathrm{base} = \mathcal{B}^{(t)}_\mathrm{base}(\bm{C}, L, \tilde{B}^{(t)}, \sigma)$ as a collection of $L$-layered MLPs with width $\bm{C}$:
\begin{equation}\label{eq:definition-base-mlp}
    \mathcal{B}^{(t)}_\mathrm{base} :=
    \left\{
        \bm{x} \mapsto \sigma ( \cdots \sigma(\bm{x} W^{(1)})\cdots W^{(L-1)})W^{(L)} \mid \|W^{(l)}_{\cdot c}\|_{1} \leq \tilde{B}^{(t)} \text{\ for all\ } c\in [H_{l+1}]
    \right\}.
\end{equation}
Here, $W^{(l)}\in \mathbb{R}^{C_{l}\times C_{l+1}}$ for $l=1, \ldots, L$\footnote{As usual, we can take into account of bias terms by preprocessing the input as $\mathbb{R}^C \ni \bm{x} \mapsto (\bm{x}, 1) \in \mathbb{R}^{C+1}$ } and $\sigma:\mathbb{R}\to \mathbb{R}$ is a $1$-Lipschitz function such that $\sigma(0)=0$ (e.g., ReLU and sigmoid). We apply $\sigma$ to a vector in an element-wise manner.
For $t\in \mathbb{N}_+$, $\tilde{P}^{(t)}\in \mathbb{R}^{N\times N}$, and $\tilde{C}^{(t)} > 0$, we use the aggregation functions $\mathcal{G}^{(t)} = \mathcal{G}^{(t)}(\tilde{P}^{(t)}, \tilde{C}^{(t)})$ defined by
\begin{equation}\label{eq:g-definition}
    \mathcal{G}^{(t)}  := \{X \mapsto \tilde{P}^{(t)}XW \mid W\in \mathbb{R}^{C\times C} \mid     \|W_{\cdot c}\|_1 \leq \tilde{C}^{(t)} \text{\ for all\ } c\in [C]\}.
\end{equation}
When we have a graph $G$ whose nodes are identified with sample points, typical choices of $\tilde{P}^{(t)}$ are the (normalized) adjacency matrix $A$ of $G$, its augmented variant $\tilde{A}$ used in GCN\footnote{Let $D$ be the degree matrix of $G$ and $\tilde{D}=D+I$. We define $\tilde{A}:=\tilde{D}^{-\frac{1}{2}}(A+I)\tilde{D}^{-\frac{1}{2}}$~\cite{kipf2017iclr}.}, (normalized) graph Laplacian, or their polynomial used in e.g., LanczosNet~\cite{liao2018lanczosnet}.
We can evaluate Rademacher complexity as follows. 
By combining it with Propositions~\ref{prop:generalization-gap} and Theorem \ref{thm:optimization}, we obtain test error bounds for multi-scale GNNs. See Sections~\ref{sec:proof-rademacher-complexity-of-parallel-mlp} and~\ref{sec:proof-test-error-for-multi-scale-gnns} for the proof.
\begin{proposition}\label{prop:rademacher-complexity-of-mlp}
    Suppose we use $\mathcal{B}^{(t)}$ and $\mathcal{G}^{(t)}$ defined above.
    Let $D^{(t)} = 2\sqrt{2}(2\tilde{B}^{(t)})^{L-1}\prod_{s=2}^{t}\tilde{C}^{(s)}$ and $P^{(t)}:=\prod_{s=2}^{t}\tilde{P}^{(s)}$. We have
$\mathfrak{R}(\mathcal{F}^{(t)})\leq \frac{1}{\sqrt{MU}}D^{(t)}\|P^{(t)}X\|_{\F}$.
\end{proposition}
\begin{theorem}\label{thm:monotonically-decreasing-test-error}
    Suppose we use $\mathcal{B}^{(t)}$ and $\mathcal{G}^{(t)}$ defined above.
    Let $T\in \mathbb{N}_+$ and $\alpha_t > \beta_t\geq0$ ($t\in [T]$).  
    Suppose Algorithm~\ref{alg:train} with the learning rate $\eta^{(t)} = \frac{4}{\alpha_t}$ finds a weak learner $f^{(t)}$ for any $t\in [T]$.
    Then, for any $\delta'>0$, with a probability of at least $1-\delta'$, its output satisfies
    \begin{equation}\label{eq:test-error-bound}
        \mathcal{R}(\widehat{Y})
        \leq \frac{(1 + e^\delta)\widehat{\mathcal{L}}(\widehat{Y}^{(1)})}{2M\Gamma_T}
        + \frac{4}{\sqrt{MU}}\sum_{t=1}^T\frac{D^{(t)}\|P^{(t)}X\|_{\F}}{\alpha_t}
        + c_0Q\sqrt{M\wedge U} + \sqrt{\frac{SQ}{2}\log \frac{1}{\delta'}}.
    \end{equation}
    In particular, if $\Gamma_T = \Omega(T^{\varepsilon})$ for some $\varepsilon > 0$, the first term is asymptotically monotonically decreasing with respect to $T$.
    If $\alpha_t^{-1}D^{(t)}\|P^{(t)}\|_{\op} = O(\tilde{\varepsilon}^{t})$ for some $\tilde{\varepsilon}\in (0, 1)$ independent of $T$, the second term is bounded by a constant independent of $T$.
\end{theorem}

\paragraph{Analogy to AdaBoost Bounds}
By interpreting $T$ as the depth of a GNN, this theorem clarifies how the information of intermediate layers helps to mitigate the over-smoothing problem, that is, \textit{the deeper a model is, the better it is}.
Theorem \ref{thm:monotonically-decreasing-test-error} is similar to typical test error bounds for AdaBoost in that it consists of monotonically decreasing training error terms and model complexity terms independent of $T$ (e.g.,~\cite[Corollay 7.5]{mohri2018foundations}). While hypothesis spaces are fixed for all iterations $t$ in the AdaBoost case, they can vary in our case due to the representation mixing caused by $\mathcal{G}^{(t)}$'s. The condition on $\|P\|_\op$ ensures that the hypothesis space does not grow significantly.

\paragraph{Trade-off between Model Complexity and Weak Learning Condition}
There is a trade-off between the model complexity and the satisfiability of the w.l.c.
Suppose we use the normalized adjacency matrix $A$ as $\tilde{P}^{(t)}$ for all $t$ (we can alternatively use the augmented version $\tilde{A}$).
If the underlying graph is connected and non-bipartite, then, it is known that the eigenvalues of $A$ satisfies $1 = \lambda_1 > \lambda_2 \geq \cdots \geq \lambda_N > -1$ (e.g.,~\cite[Lemma 1.7]{chung1997spectral},~\cite[Proposition 1]{Oono2020Graph}).
Let $(\xi_n)_{n\in [N]}$ be the orthonormal basis consisting of eigenvectors of $A$ and 
we decompose $X$ as $X_c = \sum_{n=1}^N a_{nc}\xi_n$ ($a_{nc}\in \mathbb{R}$). We denote $\tilde{X}^{(t)} :=P^{(t)}X$.
Assume that $|\lambda_n|$'s are small for $n\geq 2$.
On the one hand, we have $\|\tilde{X}^{(t)}\|^2_{\F} = \|X\|^2_{\F} - \sum_{n\geq 2}\sum_{c=1}^C (1-\lambda_n^{2t})a_{nc}^2$.
Therefore, the model complexity terms decrease rapidly with respect to $t$.
On the other hand, we have $\tilde{X}^{(t)}_c = \xi_{1c} a_{1c} + \sum_{n=2}^N \lambda_n^{t}\xi_n a_{nc}$. Therefore, $\tilde{X}^{(t)}$ degenerates to a rank-one vector of the form $\xi_1\otimes v$ ($v\in \mathbb{R}^C$) quickly under the condition.
Since it is known that $(\xi_{1})_i \propto \deg(i)^{\frac{1}{2}}$ where $\deg(\cdot)$ is the node degree (e.g., see~\cite{chung1997spectral}), $\tilde{X}^{(t)}$ has little information for distinguishing nodes other than node degrees (corresponding to the information-less space $\mathcal{M}$ in~\cite{Oono2020Graph}).
Therefore, it is hard for weak learners to satisfy w.l.c.\ using the smoothened representations $\tilde{X}^{(t)}$.
We shall discuss the large model complexity case in Section~\ref{sec:discussion}.

\paragraph{General Transformation Functions}
We have used MLPs as a specific choice of $\mathcal{B}^{(t)}$.
More generally, by using the proposition below, we can reduce the computation of the transductive Rademacher complexity to that of the inductive counterpart without any structural assumption on $\mathcal{B}^{(t)}$ other than the parallel function application of the form~\eqref{eq:parallel-transformation}.
See Section~\ref{sec:proof-bound-by-inductive-rademacher-complexity} for the proof.
Note that if the order of training and test sample sizes are same, this bound does not worsen the dependency on sample sizes. This assumption corresponds to the case where the ratio $r$ defined below satisfies $r = \Theta(1)$ as a function of $N$.
\begin{proposition}~\label{prop:bound-by-inductive-rademacher-complexity}
    Let $r:=\frac{U}{M}$.
    Suppose $\mathcal{B}^{(t)}$ is of the form~\eqref{eq:parallel-transformation} for some $\mathcal{B}^{(t)}_{\mathrm{base}}$. Use~\eqref{eq:g-definition} as $\mathcal{G}^{(s)}$ for $s\in [t]$. Define $\mathcal{F}^{(t)}_\mathrm{base} \subset \{\mathcal{X} \to \widehat{\mathcal{Y}}\}$ by
    \begin{equation*}
        \mathcal{F}^{(t)}_\mathrm{base} :=
        \{\bm{x} \mapsto f(\bm{x}W^{(2)} \cdots W^{(t)}) \mid f\in \mathcal{B}_{\mathrm{base}}^{(t)}, \|W^{(s)}_{c\cdot}\|_1 \leq \tilde{C}^{(s)}, \forall c\in [C], s=2, \ldots, t\}.
    \end{equation*}    
    We denote the (non-transductive) empirical Rademacher complexity of $\mathcal{F}^{(t)}_{\mathrm{base}}$ conditioned on $P^{(t)}X$ by $\widehat{\mathcal{\mathfrak{R}}}_{\mathrm{ind}}(\mathcal{F}^{(t)}_{\mathrm{base}}; P^{(t)}X)$ (see Definition~\ref{def:inductive-rademacher-complexity}).
    Then, we have 
    $\mathfrak{R}(\mathcal{F}^{(t)})
        < \frac{(1+r)^2}{r}\widehat{\mathcal{\mathfrak{R}}}_{\mathrm{ind}}(\mathcal{F}^{(t)}_{\mathrm{base}}; P^{(t)}X)$.
\end{proposition}

%% file: source/main/practical_consideration.tex
\section{Practical Considerations}\label{sec:model-variants}

\paragraph{Learning Kernels}

The one-layer transformation of a GNN $X\mapsto AXW$ can be considered as a kernelized linear model whose Gram matrix is $\mathcal{K} = AXX^\top A$.
This interpretation motivates us to select aggregate functions by learning appropriate kernels from data.
We employ kernel target alignment (KTA)~\cite{JMLR:v13:cortes12a,NIPS2001_1946} typically used in the context of kernel methods.
Specifically, for $Z\in \mathbb{R}^{N\times C}$, we denote its Gram matrix $\mathcal{K}[Z]\in \mathbb{R}^{M\times M}$ of the training data by $\mathcal{K}[Z]_{ij} := Z_i Z_j^\top$ for $i, j\in \Vtrain$.
We define the correlation $\rho$ by $\rho(Z, Z') :=\frac{\langle \mathcal{K}[Z], \mathcal{K}[Z'] \rangle}{\|\mathcal{K}[Z]\|_\F \|\mathcal{K}[Z']\|_\F}$.
Given a set of aggregation functions $\mathcal{G}^{(t)}_{\mathrm{KTA}}$, we choose the aggregation function $g^{(t)}$ such that
$g^{(t)} \in \argmax_{g\in \mathcal{G}^{(t)}_{\mathrm{KTA}}} \rho(g(X^{(t-1)}), Y)$ is approximately valid\footnote{When the task is a classification in which $\mathcal{Y} = [K]$, we identify $Y\in \mathcal{Y}^N$ with the matrix consisting of one-hot vectors: $\tilde{Y}=(\tilde{y}_1, \ldots, \tilde{y}_N)^\top \in\mathbb{R}^{N\times K}$ with $\tilde{y}_{nk}=\bm{1}\{y_{n} = k\}$.}.
If we can assume graph structures that represent the relationships of sample points, we can utilize them to define $\mathcal{G}^{(t)}_{\mathrm{KTA}}$.
For example, we used the linear combinations of various powers of the adjacency matrix in our experiments.

\paragraph{Fine Tuning}

After the training using boosting algorithms, we can optionally fine-tune the whole model.
For example, if each component of the model is differentiable, we can train it in an end-to-end manner using backpropagation.
Because fine-tuning does not increase the Rademacher complexity, it does not worsen the generalization gap bound.
Therefore, if fine-tuning does not increase the training error, it does not worsen the test error theoretically. However, because it is not true in all cases, we should compare the model with and without fine-tuning and select the better of the two in practice.

\paragraph{Computational Complexity}
The memory-efficiency is an advantage of the boosting algorithm.
When we train the model without fine-tuning, we do not have to retain intermediate weights and outputs.
Therefore, its memory usage is constant w.r.t. $T$, assuming that the memory usage of transformation functions $b^{(t)}$ are the same.
This is in contrast to the ordinal GNN models trained in an end-to-end manner.
Fine-tuned models use memory proportional to the depth $T$.

%% file: source/main/experiment.tex
\section{Experiments}\label{sec:experiment}

To confirm that boosting algorithms can train multi-scale GNNs practically and our theoretical observations reflect practical behaviors, we applied our models, coined \textit{Gradient Boosting Graph Neural Networks} (GB-GNN), to node classification tasks on citation network. We used Cora~\cite{mccallum2000automating,sen2008collective}, CiteSeer~\cite{giles1998citeseer, sen2008collective}, and PubMed~\cite{sen2008collective} datasets.
We used SAMME~\cite{hastie2009multi}, an extension of AdaBoost for multi-class classification tasks, as a boosting algorithm.
We considered two variants as aggregation functions $\mathcal{G}$: the multiplication model $\mathcal{G}_{\tilde{A}}$ by the augmented normalized adjacency matrix $\tilde{A}$ consisting of a singleton $\mathcal{G}_{\tilde{A}}:=\{X\mapsto \tilde{A}X\}$ and the KTA model $\mathcal{G}_{\mathrm{KTA}}$ (we refer to them as GB-GNN-Adj and GB-GNN-KTA, respectively).
We employed MLPs with single hidden layers as transformation functions $\mathcal{B}$.
See Section~\ref{sec:experiment-details} regarding the further experiment setups.
In Section~\ref{sec:experiment-additional-results}, we report the performance of three types of model variants that use (1) MLPs with different layer size, (2) Input Injection, which is another node aggregation strategy similar to GCNII~\cite{icml2020_2172}, and (3) SAMME.R, a different boosting algorithm.

Table~\ref{tbl:node_classification_model_variation} presents the prediction accuracy. It is noteworthy that boosting algorithms greedily train the models and achieve comparable performance to existing GNNs trained in an end-to-end manner by backpropagation.
Fine-tuning enhanced the performance of GB-GNN-KTA in the Cora dataset. However, whether it works well depends on model--dataset combinations.
One fine-tuned model failed due to the memory error.
There are two reasons. First, our implementation naively processes all nodes at once. Second, memory consumption of fine-tuning models increase proportionally to the depth $T$. These problems are not specific to our model but common to end-to-end deep GNN models. We can solve them by node mini-batching.

Figure~\ref{fig:train-test-cosine} shows the transition of loss values and angle between the obtained weak learners $f^{(t)}$ and negative gradients $-\nabla\widehat{\mathcal{L}}(\widehat{Y}^{(t)})$ during the training of GB-GNN-Adj using the CiteSeer dataset.
Both training and test errors keep decreasing until GB-GNN has grown up to be a deep model with as many as 40 weak learners. Accordingly, the angle is acute within this period, meaning that the w.l.c.\ is satisfied by Proposition~\ref{prop:equivalence-of-wlc}.
In later iterations, the training and test errors saturate, and the angle fluctuates. These behaviors are consistent with Theorem~\ref{thm:monotonically-decreasing-test-error}, which implies that the training and test error bounds monotonically decrease under the w.l.c. We observed similar behaviors for MLPs with various layer sizes.

\input{table/citation_variants}
\input{image/cosine_loss}

%% file: table/citation_variants.tex
\begin{table}
  \caption{Accuracy of node classification tasks. Numbers denote the $(\textrm{mean})\pm (\textrm{standard deviation}) (\%)$ of ten runs. $(\ast)$ All runs failed due to GPU memory errors. $(\ast\ast)$ We have cited the result of~\cite{kipf2017iclr}. See Section \ref{sec:model-comparison} for more comprehensive comparisons with other GNNs.}
  \label{tbl:node_classification_model_variation}
  \centering
  \begin{tabular}{lllll}
    \toprule
       Model & & Cora & CiteSeer & PubMed\\
    \midrule
GB-GNN & Adj.
& $79.9 \pm 0.8$ & $70.5 \pm 0.8$ & $79.4 \pm 0.2$ \\
 & Adj. + Fine Tuning
& $80.4 \pm 0.8$ & $70.8 \pm 0.8$ & $79.0 \pm 0.5$ \\
 & KTA
& $80.9 \pm 0.9$ & $73.1 \pm 1.1$ & $79.1 \pm 0.4$ \\
 & KTA + Fine Tuning
& $82.3 \pm 1.1$ & $70.8 \pm 1.0$ & N.A.$^{(\ast)}$ \\
\midrule
GCN$^{(\ast\ast)}$ & -- & $81.5$ & $70.3$ & $79.0$\\
    \bottomrule
  \end{tabular}
\end{table}

%% file: image/cosine_loss.tex
\begin{figure}
  \includegraphics[width=0.33\linewidth]{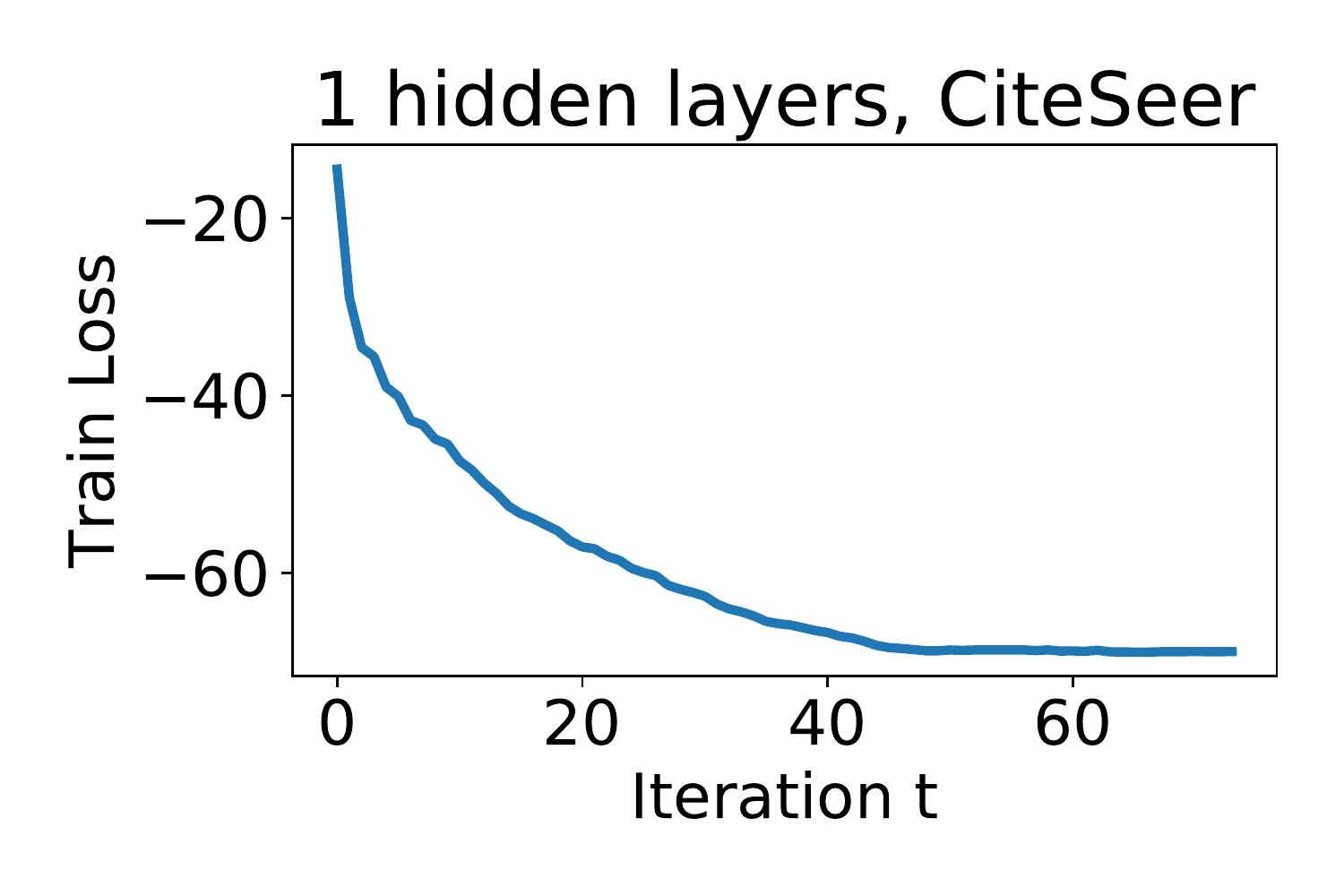}
  \includegraphics[width=0.33\linewidth]{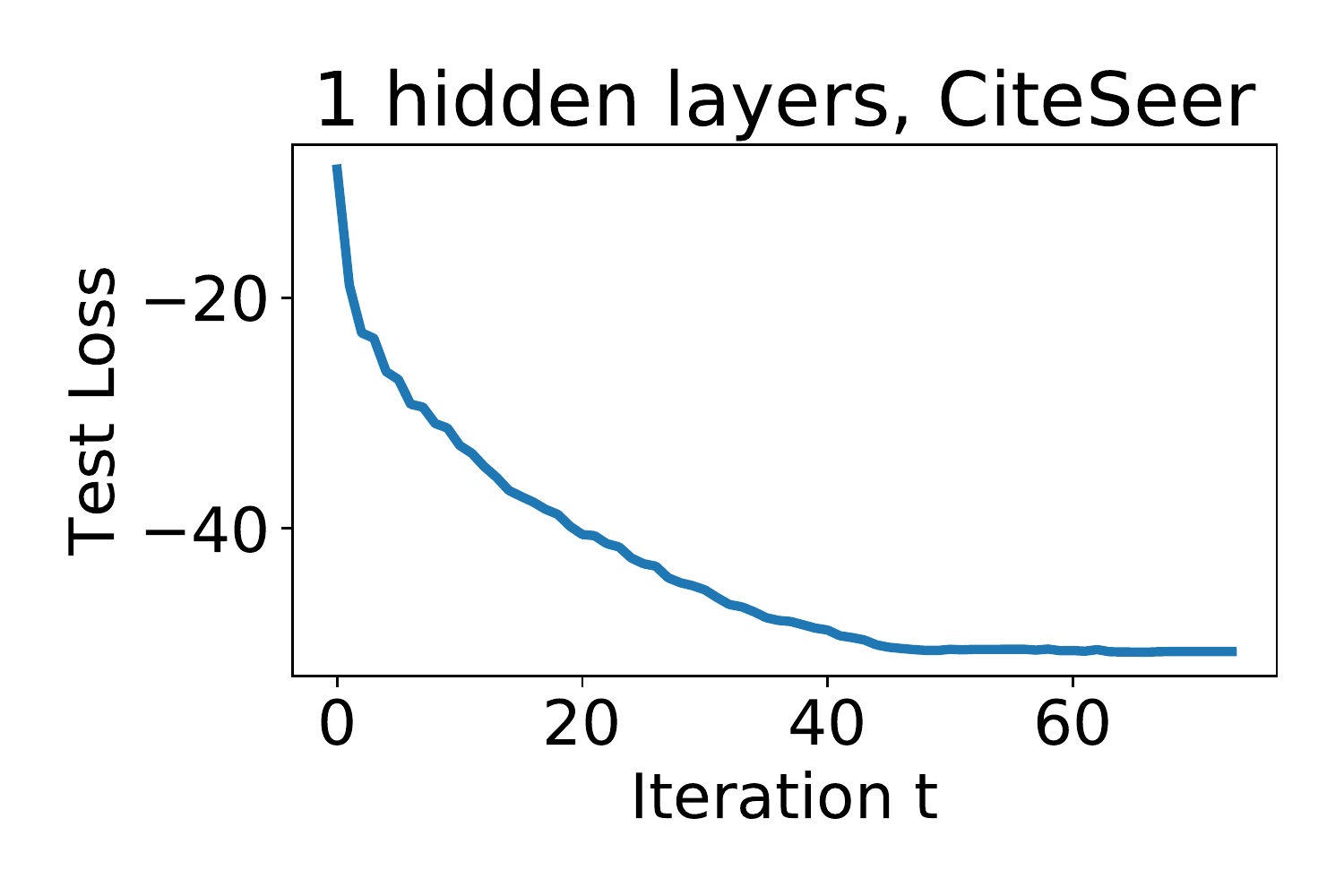}
  \includegraphics[width=0.33\linewidth]{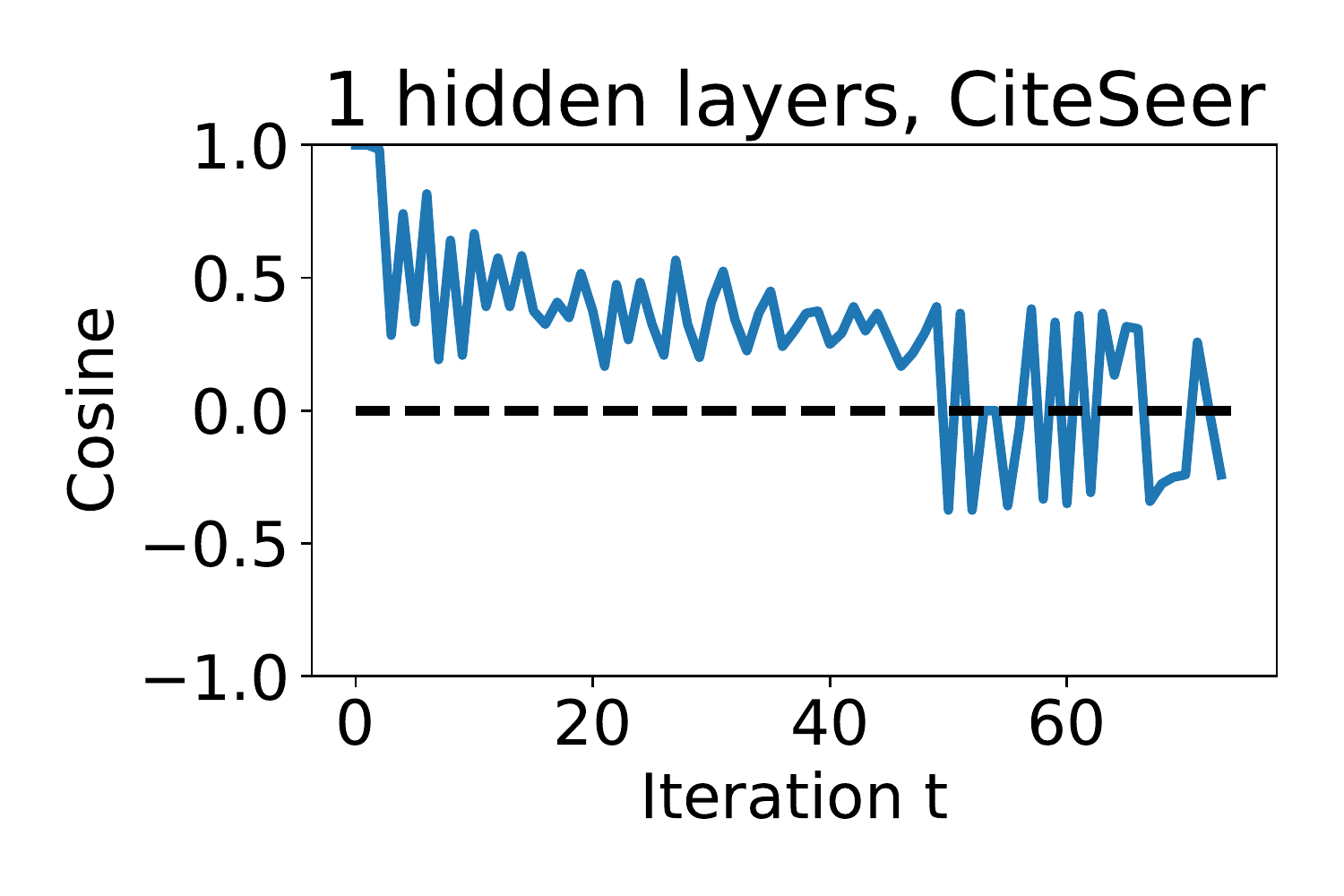}
  \caption{Results of GB-GNN-Adj trained with the CiteSeer dataset. (Left) The transition of the training loss, (Middle) test loss, (Right) angle $\cos \theta^{(t)}$ between weak learners and negative gradients.} \label{fig:train-test-cosine}
\end{figure}

%% file: source/main/discussion.tex
\section{Discussion}\label{sec:discussion}

\paragraph{Satisfiability of Weak Learning Condition}

The key assumption of our theory is the w.l.c. (Definition~\ref{def:weak_learnability_condition}). 
Although we have observed in Section~\ref{sec:experiment} that the w.l.c.\ empirically holds, it is a natural question whether we can \textit{provably} ensure it.
To obtain the guaranteed w.l.c., the model must be sufficiently expressive so that it can approximate all possible values of the negative gradient.
We can show that gradient descent can find a weak learner made of an overparameterized MLP that the w.l.c.\ holds with high probability, by leveraging the optimization analysis in the NTK regime~\cite{pmlr-v97-arora19a,du2018gradient} (see Section~\ref{sec:wlc-by-overparameterized-model} for details).
However, the guaranteed w.l.c.\ comes at the cost of large model complexity, as evident the following proposition (see Section~\ref{sec:proof-of-trade-off-between-wlc-and-rademacher-complexity} for the proof).
\begin{proposition}\label{prop:trade-off-between-wlc-and-rademacher-complexity}
    Let $\mathcal{V}, \mathcal{V}_{g}\subset \mathbb{R}^N$, and $\alpha > \beta \geq 0$ such that $\{-1, 0, 1\}^N \subset \mathcal{V}_{g}$.
    If for any $\bm{g}\in \mathcal{V}_{g}$, there exists $Z\in \mathcal{V}$ such that $Z$ satisfies $(\alpha, \beta, \bm{g})$-w.l.c., then, we have $\mathfrak{R}(\mathcal{V}) \geq \frac{\alpha^2 - \beta^2}{\alpha}$.
\end{proposition}
Let $\mathcal{V}_{g}$ be the set of possible values that can be taken by the negative gradient and $\mathcal{V}$ be the space of outputs of weak learners at a specific iteration.
Then, if we want weak learners to satisfy the w.l.c., its Rademacher complexity is inevitably as large as $\Omega(1)$ (assuming that $\alpha$ and $\beta$ are independent of $M$).
We leave the problem for future work whether there exists a setting that simultaneously satisfies the following conditions: (1) the w.l.c.\ (or similar conditions) provably holds, (2) training of weak learners is tractable, and (3) the model has a small complexity (such as the Rademacher complexity).

\paragraph{Choice of Transformation Functions}

In Section~\ref{sec:application-to-multi-scale-gnns}, we used an $L$-layered MLP with $L_1$ norm constraints as a transformation function $\mathcal{B}^{(t)}$.
The test error bound in Theorem~\ref{thm:monotonically-decreasing-test-error} can exponentially depend on $L$ via the constant $D^{(t)}$.
Since this problem occurs in inductive MLPs, too, many studies derived generalization bounds that avoid this exponential dependency~\cite{pmlr-v80-arora18b,nagarajan2018deterministic,NIPS2019_9166}.
We can incorporate them to obtain tighter bounds using Proposition~\ref{prop:bound-by-inductive-rademacher-complexity}.

\paragraph{Choice of Node Aggregation Functions}

We considered a \textit{linear} aggregation model as $\mathcal{G}^{(t)}$ in Section~\ref{sec:experiment} because GNNs that consist of linear node aggregations and non-linear MLPs is practically popular in the GNN research, such as SGC~\cite{wu2019simplifying}, gfNN~\cite{nt2019revisiting}, and APPNP~\cite{klicpera2018combining}.
Theoretically,~\cite{nt2019revisiting,Oono2020Graph} claimed that non-linearity between aggregations is not essential for predictive performance.

We can alternatively use \textit{non-linear} aggregation models.
For example, consider the model $X \mapsto \sigma(\tilde{P}^{(t)}XW)$ with the same $L_1$ constraints as \eqref{eq:g-definition} as $\mathcal{G}^{(t)}$,
where $\sigma: \mathbb{R}\to \mathbb{R}$ is the ReLU function.
Adding the non-linearity $\sigma$ changes two things.
First, $\|P^{(t)}X\|_{\mathrm{F}}$ in the bound of Theorem~\ref{thm:monotonically-decreasing-test-error} is replaced with $\prod_{s=2}^t \|\tilde{P}^{(s)}\|_{\mathrm{op}}\|X\|_{\mathrm{F}}$.
It makes the interpretation of the trade-off discussed in Section~\ref{sec:application-to-multi-scale-gnns} impossible, and the bound looser, essentially because the bound loses the information of eigenvectors.
Second, the bound for Rademacher complexity of $\mathcal{F}^{(t)}$ is multiplied by $2^t$.
It changes the condition for the monotonically decreasing test error bound with respect to $T$ from $\alpha_t^{-1}D^{(t)}\|P^{(t)}\|_{\op} = O(\tilde{\varepsilon}^{t})$ to a stricter one $\alpha_t^{-1}2^tD^{(t)}\prod_{s=2}^t \|\tilde{P}^{(s)}\|_{\mathrm{op}} = O(\tilde{\varepsilon}^t)$.

With that being said, we do not have a definitive answer whether linear aggregation models are truly superior to non-linear ones --- we may be able to use techniques similar to \cite{Oono2020Graph} for the first problem and refined analyses could eliminate the $2^t$ term for the second problem.

%% file: source/main/conclusion.tex
\section{Conclusion}

In this study, we analyzed a certain type of transductive learning models and derived their optimization and generalization guarantees under the weak learnability condition (w.l.c.). Our idea was to interpret multi-scale GNNs as an ensemble of weak learners and apply boosting theory. As a special case, we showed that a particular type of multi-scale GNNs has a generalization bound that is decreasing with respect to the number of node aggregations under the condition. To the best of our knowledge, this is the first result that multi-scale GNNs provably avoid the over-smoothing from the viewpoint of learning theory. We confirmed that our models, coined GB-GNNs, worked comparably to existing GNNs, and that their empirical behaviors were consistent with theoretical observations. We believe that exploring deeper relationships between the w.l.c.\ and the underlying graph structures such as graph spectra is a promising direction for future research.

%% file: source/main/broader_impact.tex
\section*{Broader Impact}

\paragraph{Benefits}

Deepening the theoretical understanding of machine learning models motivates people to explore their advanced usage. For example, the universality of MLPs boosted their usage as a building block of more advanced models as a function approximator (e.g., deep reinforcement learning). In this study, we investigated the probable optimization and generalization guarantees of GNNs, especially multi-scale GNNs. We expect that this research to broaden the applicability of GNNs in both theoretical and practical situations.

\paragraph{Potential Risks and Associated Mitigations}

The theoretical understanding of GNNs could result in their misuse, either intentionally or unintentionally. For example, if a GNN is used in social networks, the contamination of biased information on the networks such as fake news could affect the prediction of GNNs, thereby resulting in the promotion of social disruption or unfair treatment to minority groups. The methods devised in the study of fairness could mitigate such misuse of GNNs.

%% file: source/acknowledgement.tex
\begin{ack}
We would like to thank the following people for insightful discussions:
Kohei Hayashi, Masaaki Imaizumi, Masanori Koyama, Takanori Maehara, Kentaro Minami, Atsushi Nitanda, Akiyoshi Sannai, Sho Sonoda, and Yuuki Takai.
TS was partially supported by JSPS KAKENHI (18K19793, 18H03201, and 20H00576), Japan Digital Design, and JST CREST.
\end{ack}

%% file: source/appendix_content.tex
\begin{appendices}
This is the supplemental material for \textit{\papertitle}.

\input{source/appendix/proofs}
\input{source/appendix/more_related_works}
\input{source/appendix/wlc_by_overparameterized_model}
\input{source/appendix/more_model_variants}
\input{source/appendix/experiment_details}
\input{source/appendix/experiment_additional_results}
\end{appendices}

%% file: source/appendix/proofs.tex
\section{Proof of Theorems and Propositions}

We give proofs for the theorems and propositions in the order they appeared in the main paper.

\input{source/appendix/proofs/proof_equivalence_of_wlc_extended}
\input{source/appendix/proofs/proof_optimization}
\input{source/appendix/proofs/proof_generalization}
\input{source/appendix/proofs/proof_rademacher_complexity_of_parallel_mlp}
\input{source/appendix/proofs/proof_generalization_bound_for_multi_scale_gnns}
\input{source/appendix/proofs/proof_bound_by_inductive_rademacher_complexity}
\input{source/appendix/proofs/proof_trade_off_between_wlc_and_rademacher_complexity}

%% file: source/appendix/proofs/proof_equivalence_of_wlc_extended.tex
\subsection{Proof of Proposition~\ref{prop:equivalence-of-wlc}}\label{sec:proof-equivalence-of-wlc}

We prove the more detailed claim.
Proposition~\ref{prop:equivalence-of-wlc} is a part of the following proposition.
The third condition below means that the prediction $Z$ is better than a random guess as a solution to the binary classification problem on the training dataset weighed by $w_n$'s.
The proof for the equivalence of the second and third conditions are similar to that of \cite[Theorem 1]{ICML2011Grubb_626}

\begin{proposition}
    Let $Z, \bm{g}\in \mathbb{R}^N$ such that $\bm{g}\not =0$.
    The followings are equivalent
    \begin{enumerate}
        \item There exist $\alpha, \beta$ such that $\alpha > \beta \geq 0$ and $Z$ satisfies $(\alpha, \beta, \bm{g})$-w.l.c.
        \item $\langle Z, \bm{g}\rangle > 0$.
    \end{enumerate}
    Under the condition, for any $r \in [\sin^2\theta, 1)$, we can take $\alpha := C$, $\beta := rC$, and $\|Z - \alpha \bm{g} \|_2 = \beta \|\bm{g}\|_2$, where 
    \begin{equation*}
        \cos \theta = \frac{\langle Z, \bm{g}\rangle}{\|Z\|_2\|\bm{g}\|_2},
        \quad C := \frac{\langle Z, \bm{g}\rangle \pm \sqrt{\langle Z, \bm{g} \rangle^2 - (1-r^2)\|Z\|^2\|\bm{g}\|^2 } }{(1-r^2)\|\bm{g}\|^2_2}.
    \end{equation*}
    Suppose further $Z \in \{\pm 1\}^{N}$, then, the conditions 1 and 2 are equivalent to
    \begin{enumerate}
        \setcounter{enumi}{2}
        \item There exists $\delta \in (0, 1]$ such that $\sum_{n=1}^N w_n\bm{1}\{\sign(\bm{g}_n) \not = Z_n\} \leq \frac{1 - \delta}{2}$ where $w_n = \frac{\bm{g}_n}{\|\bm{g}\|_1}$.
    \end{enumerate}
\end{proposition}

\begin{proof}
    $(1.\implies2.)$ We have
    \begin{align}
    \|Z - \alpha \bm{g}\|_2 \leq \beta \|\bm{g}\|_2
    \Longleftrightarrow \ &
        \|Z\|_{2}^2
            - 2\alpha \langle Z, \bm{g}\rangle
            + \alpha^2 \|\bm{g}\|^2_2
        \leq \beta^2 \|\bm{g}\|^2_2 \notag\\
    \Longleftrightarrow \ &
         \langle Z, \bm{g}\rangle \geq \frac{\|Z\|^2_{2}}{2\alpha} + \frac{\alpha^2 - \beta^2}{2\alpha}\|\bm{g}\|^2_{2} > 0. \label{eq:inner-prod-lower-bound}
    \end{align}
    
    $(2.\implies1.)$ For $k>0$, we define
    \begin{equation*}
        \tilde{r}(k) := \frac{\|Z - k\bm{g}\|_2}{k\|\bm{g}\|_2}.
    \end{equation*}
    Then, by direct computation, we have 
    \begin{align*}
        \tilde{r}(k)^2
        = \frac{\|Z\|_2^2}{\|\bm{g}\|_2^2}\left(\gamma - \frac{\langle Z, \bm{g} \rangle}{\|Z\|_2^2}\right)^2 + 1 - \frac{\langle Z, \bm{g} \rangle^2}{\|Z\|_2^2\|\bm{g}\|_2^2}.
    \end{align*}
    where $\gamma := k^{-1}$.
    Therefore, $\tilde{r}(k)$ is a quadratic function of $\gamma$ that takes the minimum value $\sin^2\theta$ at $\gamma = \frac{\langle Z, \bm{g}\rangle}{\|Z\|_2^2} > 0$.
    Therefore, for any $r\in [\sin^2\theta, 1) $ there exists $k_0 > 0$ such that $\tilde{r}(k_0) = r$.
    Then, by setting $\alpha:= k_0$ and $\beta := r k_0$, we have $\alpha > \beta \geq 0$ and
    \begin{equation*}
        \|Z - \alpha\bm{g}\|_2
        = \tilde{r}(k_0) \alpha \|\bm{g}\|_2
        = r\alpha\|\bm{g}\|_2
        = \beta \|\bm{g}\|_2.
    \end{equation*}
    By solving $\tilde{r}(k_0) = r$, we obtain $\alpha=C$ and $\beta=rC$.
    
    $(1.\implies3.)$
    Define $w^+ := \sum_{n=1}^N w_n \bm{1}\{Z_n = \sign(\bm{g}_n)\}$ and $w^- := \sum_{n=1}^N w_n \bm{1}\{Z_n \not= \sign(\bm{g}_n)\}$.
    By definition, we have $w^+ + w^- = \|w\|_1 = 1$.
    \begin{align}
        \langle Z, \bm{g}\rangle
        &= \sum_{n=1}^N Z_n \bm{g}_n \notag\\
        &= \sum_{n=1}^N Z_n w_n \|\bm{g}\|_1 \sign(\bm{g}_n)\notag\\
        &= \|\bm{g}\|_1 (w^+ - w^-).\label{eq:from-inner-prod-to-classification-error}
    \end{align}
    Therefore, using the reformulation \eqref{eq:inner-prod-lower-bound} of the assumption, we have
    \begin{align*}
        w^+ - w^-
        &= \frac{\langle Z, \bm{g} \rangle}{\|\bm{g}\|_1} \\
        &\geq \frac{\langle Z, \bm{g} \rangle}{\|\bm{g}\|_2\sqrt{N}} 
            \quad \text{($\because$ Cauchy--Schwraz inequality))}\\
        &\geq \frac{\sqrt{N}}{2\alpha\|\bm{g}\|_2}
            + \frac{\alpha^2 - \beta^2}{2\alpha}\frac{\|\bm{g}\|_{2}}{\sqrt{N}}
            \quad \text{($\because$ \eqref{eq:inner-prod-lower-bound} and $\|Z\|_2^2  = N$)}\\
        &\geq 2\sqrt{\frac{1}{2\alpha} \frac{\alpha^2 - \beta^2}{2\alpha}} 
            \quad \text{($\because$ AM--GM inequality)}\\
        &= \sqrt{1 - \frac{\beta^2}{\alpha^2}}.
    \end{align*}
    Set $\delta := \sqrt{1 - \frac{\beta^2}{\alpha^2}}$.
    By the assumption $\alpha > \beta \geq 0$, we have $\delta \in (0, 1]$.
    Therefore, we have
    \begin{align*}
        w^+
        &= \frac{1}{2}(w^+ + w^-) + \frac{1}{2}(w^+ - w^-)\\
        &\geq \frac{1}{2}(w^+ + w^-) + \frac{\delta}{2}\\
        &= \frac{1+\delta}{2},
    \end{align*}
    which is equivalent to $w^- \leq \frac{1}{2}(1-\delta)$.    
    
    \paragraph{$(3.\implies2.)$}
    Using the same argument as \eqref{eq:from-inner-prod-to-classification-error}, we have 
    \begin{align*}
        \langle Z, \bm{g}\rangle = \|\bm{g}\|_1 (w^+ - w^-).
    \end{align*}
    By the assumption, we have
    \begin{align*}
        & w^- \leq \frac{1-\delta}{2}(w^+ + w^-)\\
        & \Longleftrightarrow w^+ \geq \frac{1+\delta}{2}(w^+ + w^-)\\
        & \Longleftrightarrow w^+ - \frac{w^++w^-}{2}\geq \frac{1+\delta}{2}(w^+ + w^-) - \frac{w^++w^-}{2}\\
        & \Longleftrightarrow \frac{1}{2}(w^+ - w^-) \geq \frac{\delta}{2}(w^+ + w^-)\\        
    \end{align*}
    Therefore, we have
    \begin{align*}
        \langle Z, \bm{g}\rangle
        &= \|\bm{g}\|_1 (w^+ - w^-)\\
        &= \|\bm{g}\|_1 \delta(w^+ + w^-)\\
        &= \|\bm{g}\|_1 \delta > 0.
    \end{align*}
\end{proof}

%% file: source/appendix/proofs/proof_optimization.tex
\subsection{Proof of Theorem \ref{thm:optimization}}\label{sec:proof-of-optimization}

\begin{assumption}\label{asm:loss}
    $\ell: \widehat{\mathcal{Y}}\times \mathcal{Y} \to \mathbb{R}$ is a non-negative $C^2$ convex funxtion with respect to the first variable and satisfies $|\nabla^2_{\hat{y}} \ell(\hat{y}, y) | \leq A$ for all $\hat{y} \in \widehat{\mathcal{Y}}$ and $y\in \mathcal{Y}$.
\end{assumption}
\begin{proposition}\label{prop:sce-smoothness}
    The sigmoid cross entropy loss $\ell_\sigmoid$ satisfies Assumption \ref{asm:loss} with $A=\frac{1}{4}$. 
\end{proposition}

\begin{lemma}\label{lem:small-gradient}
    Suppose the loss function $\ell$ satisfies Assumption \ref{asm:loss} with $A>0$.
    Define the training error $\widehat{\mathcal{L}}$ by $\widehat{\mathcal{L}}(\widehat{Y}):=\frac{1}{M}\sum_{n=1}^M \ell (\hat{y}_n, y_n)$ for $\widehat{Y}^{\top} = (\hat{y}_1, \ldots, \hat{y}_N)^{\top}$.
    Suppose Algorithm~\ref{alg:train} with the learning rate $\eta^{(t)} = \frac{1}{A\alpha_t}$ finds a weak learner $f^{(t)}$ for any $t\in [T]$.
    Then, we have
    \begin{align*}
        \sum_{t=1}^{T} \gamma_t \|\nabla \widehat{\mathcal{L}} (\widehat{Y}^{(t)})\|^2_{\F}
        \leq \frac{2A\widehat{\mathcal{L}}(\widehat{Y}^{(1)})}{M}.
    \end{align*}
    \begin{proof}
        First, we define $C_f^{(t)} := (2\alpha_t)^{-1}$ and $C_{\mathcal{L}}^{(t)} := \frac{\alpha_t^2 - \beta_t^2}{2\alpha_t}$.
        Note that we have by definition 
        \begin{align}
            \gamma_t = 4C_f^{(t)}C_{\mathcal{L}}^{(t)}.\label{eq:gamma-t}
        \end{align}
        We denote $Z^{(t)\top} = (z_1^{(t)}, \ldots, z_N^{(t)})^\top := \frac{1}{M}f^{(t)}(X)^\top$ and $(\hat{y}^{(t)}_1, \ldots, \hat{y}^{(t)}_N)^{\top}:=\widehat{Y}^{(t)\top}$.
        Since $Z^{(t)}$ satisfies $(\alpha_t, \beta_t, -\nabla \widehat{\mathcal{L}}(\widehat{Y}^{(t-1)}))$-w.l.c., we have 
        \begin{align}
            & \|Z^{(t)} + \alpha_t \nabla \widehat{\mathcal{L}}(\widehat{Y}^{(t-1)})\|_\F
                \leq \beta_t \|\nabla \widehat{\mathcal{L}}(\widehat{Y}^{(t-1)})\|_\F \notag\\
            \Longleftrightarrow \ & \|Z^{(t)}\|_{\F}^2 + 2\alpha_t \langle Z^{(t)}, \nabla \widehat{\mathcal{L}}(\widehat{Y}^{(t-1)})\rangle + \alpha_t^2 \|\nabla \widehat{\mathcal{L}}(\widehat{Y}^{(t-1)})\|^2_\F
                \leq \beta_t^2 \|\nabla \widehat{\mathcal{L}}(\widehat{Y}^{(t-1)})\|^2_\F \notag\\
            \Longleftrightarrow \ & \langle Z^{(t)}, \nabla \widehat{\mathcal{L}}(\widehat{Y}^{(t-1)})\rangle + C_f^{(t)} \|Z^{(t)}\|^2_{\F}
                \leq -C_{\mathcal{L}}^{(t)}\|\nabla \widehat{\mathcal{L}}(\widehat{Y}^{(t-1)})\|^2_{\F} \label{eq:thm1-wlc}.
        \end{align}
        Since $\eta^{(t)} = \frac{2C_f^{(t)}}{A}$, we have
        \begin{equation*}
            \langle \nabla\widehat{\mathcal{L}}(\widehat{Y}^{(t-1)}), Z^{(t)} \rangle + \frac{A\eta^{(t)}}{2} \|Z^{(t)}\|_\F^2 \leq -C_{\mathcal{L}}^{(t)} \|\nabla \widehat{\mathcal{L}}(\widehat{Y}^{(t-1)})\|_\F^2.
        \end{equation*}
        By Taylors' theorem, and Assumption \ref{asm:loss}, we have
        \begin{align*}
            \ell(\hat{y}^{(t)}_n, y_n)
            &\leq \ell(\hat{y}^{(t-1)}_n, y_n) + \langle \nabla_{\hat{y}}\ell(\hat{y}_n^{(t-1)}, y_n), \hat{y}^{(t)}_n - \hat{y}^{(t-1)}_n\rangle + \frac{A}{2} \|\hat{y}^{(t)}_n - \hat{y}^{(t-1)}_n\|_2^2.
        \end{align*}
        By taking the average in terms of $n$, we have
        \begin{align*}
            \widehat{\mathcal{L}}(\widehat{Y}^{(t)}) \leq \widehat{\mathcal{L}}(\widehat{Y}^{(t-1)}) + \frac{1}{M}\sum_{n=1}^M \langle \nabla_{\hat{y}}\ell(\hat{y}_n^{(t-1)}, y_n), y^{(t)}_n - y^{(t-1)}_n\rangle + \frac{A}{2M} \sum_{n=1}^M \|y^{(t)}_n - y^{(t-1)}_n\|_2^2.
        \end{align*}
        By the definition of $\widehat{Y}^{(t)}$'s, we have
        \begin{align*}
            \hat{y}^{(t)}_n - \hat{y}^{(t-1)}_n = \eta^{(t)} Mz^{(t)}_n.
        \end{align*}
        Therefore, we have
        \begin{align*}
            \widehat{\mathcal{L}}(\widehat{Y}^{(t)})
            &\leq \widehat{\mathcal{L}}(\widehat{Y}^{(t-1)})
                + \eta^{(t)}\sum_{n=1}^M \langle \nabla_{\hat{y}}\ell(\hat{y}_n^{(t-1)}, y_n), z^{(t)}_n \rangle
                + \frac{1}{2}\eta^{(t)2}AM \sum_{n=1}^M \|z^{(t)}_n\|_2^2\\
            &\leq \widehat{\mathcal{L}}(\widehat{Y}^{(t-1)})
                + \eta^{(t)}\sum_{n=1}^N \langle \nabla_{\hat{y}}\ell(\hat{y}_n^{(t-1)}, y_n), z^{(t)}_n \rangle
                + \frac{1}{2}\eta^{(t)2}AM \sum_{n=1}^N \|z^{(t)}_n\|_2^2\\            
            &\leq \widehat{\mathcal{L}}(\widehat{Y}^{(t-1)})
                + \eta^{(t)}M \langle \nabla_{\hat{y}} \widehat{\mathcal{L}} (\widehat{Y}^{(t-1)}), Z^{(t)}\rangle
                + \frac{1}{2}\eta^{(t)2}AM \|Z^{(t)}\|_\F^2\\
            &\leq \widehat{\mathcal{L}}(\widehat{Y}^{(t-1)})
                - \eta^{(t)} MC_{\mathcal{L}}^{(t)} \|\nabla \widehat{\mathcal{L}}(\widehat{Y}^{(t-1)})\|_\F^2
                \quad \text{($\because$ \eqref{eq:thm1-wlc})}\\ 
            &= \widehat{\mathcal{L}}(\widehat{Y}^{(t-1)})
                - \frac{2MC_f^{(t)}C_{\mathcal{L}}^{(t)}}{A} \|\nabla \widehat{\mathcal{L}}(\widehat{Y}^{(t-1)})\|_\F^2
                \quad \text{($\because$ Definition of $\eta^{(t)}$ )}\\
            &= \widehat{\mathcal{L}}(\widehat{Y}^{(t-1)})
                - \frac{M\gamma_t}{2A} \|\nabla \widehat{\mathcal{L}}(\widehat{Y}^{(t-1)})\|_\F^2
                \quad \text{($\because$ \eqref{eq:gamma-t})}
        \end{align*}
        Rearranging the term, we get
        \begin{align*}
            \gamma_t \|\nabla \widehat{\mathcal{L}} (\widehat{Y}^{(t-1)})\|_{\F}^2
            \leq \frac{2A}{M}\left( \widehat{\mathcal{L}}(\widehat{Y}^{(t-1)}) - \widehat{\mathcal{L}}(\widehat{Y}^{(t)})\right).
        \end{align*}
        By taking the summation in terms of $t$, we have 
        \begin{align*}
            \sum_{t=1}^{T} \gamma_t \|\nabla \widehat{\mathcal{L}} (\widehat{Y}^{(t)})\|^2_{\F}
            \leq \frac{2A}{M} \left(\widehat{\mathcal{L}}(\widehat{Y}^{(1)}) - \widehat{\mathcal{L}}(\widehat{Y}^{(T+1)})\right)
            \leq \frac{2A\widehat{\mathcal{L}}(\widehat{Y}^{(1)})}{M}.
        \end{align*}
        We used the non-negativity of the loss function $\widehat{\mathcal{L}}$ in the final inequality.
    \end{proof}
\end{lemma}

\begin{lemma}\label{lem:bound-error-by-gradient}
    Assume the loss function is the cross entropy loss.
    Let $\delta\geq0$. For any $\widehat{Y} \in \mathcal{\widehat{Y}}^N$, we have
    \begin{align*}
        \widehat{\mathcal{R}}(\widehat{Y}) \leq (1 + e^{\delta}) \|\nabla \widehat{\mathcal{L}}(\widehat{Y})\|_{2, 1}
    \end{align*}
    \begin{proof}
        Same as \cite[Proposition C]{pmlr-v80-nitanda18a}.
    \end{proof}
\end{lemma}

    \begin{proof}[Proof of Theorem \ref{thm:optimization}]
        Since we use the cross entropy loss, by Lemma \ref{lem:bound-error-by-gradient}, we have 
        \begin{align}\label{eq:r-to-nabla-ell}
            \widehat{\mathcal{R}}(\widehat{Y}) \leq (1 + e^{\delta}) \|\nabla \widehat{\mathcal{L}}(\widehat{Y})\|_{2, 1}
        \end{align}
        By the definition of $\widehat{Y}$, $t^\ast$, and $\Gamma$, we have
        \begin{align}
            \Gamma_T \|\nabla \widehat{\mathcal{L}} (\widehat{Y})\|_{2, 1}
            &= \left(\sum_{t=1}^T \gamma_t\right) \|\nabla \widehat{\mathcal{L}} (\widehat{Y}^{(t^{\ast})})\|_{2, 1}\notag\\
            &\leq \sum_{t=1}^{T} \gamma_{t}\|\nabla \widehat{\mathcal{L}} (y^{(t)})\|^2_{2, 1} \notag\\
            &\leq \sum_{t=1}^{T} \gamma_{t} \|\nabla \widehat{\mathcal{L}} (y^{(t)})\|^2_{\F} \label{eq:nabla-ell-to-nabla-ell-average}
        \end{align}
        From Lemma \ref{lem:small-gradient} with $A = \frac{1}{4}$ (Proposition \ref{prop:sce-smoothness}), we have
        \begin{align}\label{eq:nabla-ell-average-to-ell}
            \sum_{t=1}^{T} \gamma_t \|\nabla \widehat{\mathcal{L}} (\widehat{Y}^{(t)})\|^2_{\F}
            \leq \frac{\widehat{\mathcal{L}}[\widehat{Y}^{(1)}]}{2M}
        \end{align}
        Combining \eqref{eq:r-to-nabla-ell}, \eqref{eq:nabla-ell-to-nabla-ell-average}, and \eqref{eq:nabla-ell-average-to-ell}, we have
        \begin{align*}
            \widehat{\mathcal{R}}(\widehat{Y})
            \leq \frac{(1 + e^{\delta})\widehat{\mathcal{L}}[\widehat{Y}^{(1)}]}{2M\Gamma_T}.
        \end{align*}
    \end{proof}

%% file: source/appendix/proofs/proof_generalization.tex
\subsection{Proof of Proposition \ref{prop:generalization-gap}}\label{sec:proof-of-generalization}

The proof is basically the application of~\cite[Corollary 1]{el2009transductive} to our setting.
We recall the definition of the transductive Rademacher complexity introduced by \cite{el2009transductive}.
\begin{definition}[Transductive Rademacher Complexity]\label{def:transductive-rademacher-complexity}
    For $p\in [0, \frac{1}{2}]$ and $\mathcal{V} \subset \mathbb{R}^N$, we define
    \begin{align*}
        \mathfrak{R}(\mathcal{V}, p) := Q \E_{\bm{\sigma}}\left[\sup_{v\in V} \bm{\sigma} \cdot v\right],
    \end{align*}
    Here, $Q = \frac{1}{M} + \frac{1}{U}$ and $\bm{\sigma} = (\sigma_1, \ldots, \sigma_N)$ is an sequence of i.i.d. random variables whose distribution is $\mathbb{P}(\sigma_i = 1) = \mathbb{P}(\sigma_i = -1) = p$ and $\mathbb{P}(\sigma_i = 0) = 1-2p$.
    In particular, we denote $\mathfrak{R}(\mathcal{V}):=\mathfrak{R}(\mathcal{V}, p_0)$ where $p_0 = \frac{MU}{(M+U)^2}$.
\end{definition}
We introduce the notion of the (weighed) sum of sets.
\begin{definition}
For $\mathcal{V}_1, \ldots, \mathcal{V}_T \subset \mathbb{R}^N$ and $\alpha_1, \ldots, \alpha_T \in \mathbb{R}$, we define their (weighted) sum $\sum_{t=1}^T \alpha_t \mathcal{V}_t$ by
\begin{equation*}
    \sum_{t=1}^T \alpha_t\mathcal{V}_t := \left\{\sum_{t=1}^T \alpha_t v_t \,\middle\vert\,  v_t \in \mathcal{V}_t\right\}.
\end{equation*}
\end{definition}
We define the hypothesis space $\mathcal{H}$ by $\mathcal{H} := \sum_{t=1}^T \eta^{(t)} \mathcal{F}^{(t)}$. Note that any output $\widehat{Y}$ of Algorithm \ref{alg:train} satisfies $\widehat{Y}\in \mathcal{H}$.
We can compute the Rademacher complexity of the sum similarly to the inductive case.
\begin{proposition}\label{prop:sum_of_rademacher_complexity}
    For $\mathcal{V}_1, \mathcal{V}_2\subset \mathbb{R}^N$, $a_1, a_2\in \mathbb{R}$ and $p\in [0, \frac{1}{2}]$, we have $\mathfrak{R}(a_1\mathcal{V}_1 + a_2\mathcal{V}_2) \leq |a_1|\mathfrak{R}(\mathcal{V}_1) + |a_2|\mathfrak{R}(\mathcal{V}_2)$.
    \begin{proof}
        Take any realization $\bm{\sigma}$ of the $N$ i.i.d. transductive Rademacher variable of parameter $p$.
        For any $v = a_1v_1 + a_2v_2\in a_1\mathcal{V}_1 + a_2\mathcal{V}_2$ ($v_1 \in \mathcal{V}_1$, $v_2 \in \mathcal{V}_2$), we have
        \begin{equation*}
            \langle \bm{\sigma}, v\rangle
            = \langle \bm{\sigma}, a_1v_1\rangle + \langle \bm{\sigma}, a_2v_2\rangle
            \leq |a_1|\sup_{v_1\in \mathcal{V}_1} \langle \bm{\sigma}, v_1\rangle + |a_2|\sup_{v_1\in \mathcal{V}_2} \langle \bm{\sigma}, v_2\rangle.
        \end{equation*}
        By taking the supremum of $v$, we have
        \begin{equation*}
            \sup_{v\in \mathcal{V}_1 + \mathcal{V}_2}\langle \bm{\sigma}, v\rangle
            \leq |a_1|\sup_{v_1\in \mathcal{V}_1} \langle \bm{\sigma}, v_1\rangle + |a_2|\sup_{v_2\in \mathcal{V}_2} \langle \bm{\sigma}, v_2\rangle.
        \end{equation*}
        The proposition follows by taking the expectation with respect to $\bm{\sigma}$.
    \end{proof}
\end{proposition}

\begin{proof}[Proof of Proposition \ref{prop:generalization-gap}]
    Let $\mathcal{H} = \sum_{t=1}^T \eta^{(t)} \mathcal{F}^{(t)}$.
    By \cite[Corollary 1]{el2009transductive}, with probability of at least $1-\delta'$ for all $\widehat{Y}' \in \mathcal{H}$, we have
    \begin{align*}
        \mathcal{R}(\widehat{Y}')
        \leq \widehat{\mathcal{R}}(\widehat{Y}') + \mathfrak{R}(\mathcal{H})
        + c_0Q\sqrt{M\wedge U} + \sqrt{\frac{SQ}{2}\log \frac{1}{\delta'}}.
    \end{align*}
    Since the output $\widehat{Y}$ of Algorithm~\ref{alg:train} satisfies $\widehat{Y}\in \mathcal{H}$, we have
    \begin{align}\label{eq:thm2-eq1}
        \mathcal{R}(\widehat{Y})
        \leq \widehat{\mathcal{R}}(\widehat{Y}) + \mathfrak{R}(\mathcal{H})
        + c_0Q\sqrt{M\wedge U} + \sqrt{\frac{SQ}{2}\log \frac{1}{\delta'}}.
    \end{align}
    By Proposition \ref{prop:sum_of_rademacher_complexity}, we have
    \begin{align}\label{eq:thm2-eq2}
        \mathfrak{R}(\mathcal{H}) \leq \sum_{t=1}^T \eta^{(t)}\mathfrak{R}(\mathcal{F}^{(t)}).
    \end{align}
    Combining \eqref{eq:thm2-eq1} and \eqref{eq:thm2-eq2} concludes the proof.
\end{proof}

%% file: source/appendix/proofs/proof_rademacher_complexity_of_parallel_mlp.tex
\subsection{Proof of Proposition~\ref{prop:rademacher-complexity-of-mlp}}
\label{sec:proof-rademacher-complexity-of-parallel-mlp}

We shall prove Proposition~\ref{prop:rademacher-complexity-of-mlp-general}, which is more general than Proposition~\ref{prop:rademacher-complexity-of-mlp}.
To formulate it, we first introduce the variant of the transductive Rademacher complexity.
\begin{definition}[(Symmetrized) Transductive Rademacher Complexity]\label{def:symmetrized-transductive-rademacher-complexity}
For $\mathcal{V} \subset \mathbb{R}^N$ and $p\in [0, 1/2]$, we define the symmetrized transductive Rademacher complexity $\overline{\mathfrak{R}}(\mathcal{V}, p)$ by
\begin{equation*}
    \overline{\mathfrak{R}}(\mathcal{V}, p) := Q \E_{\bm{\sigma}} \left[ \sup_{v\in \mathcal{V}} \left|\left\langle\bm{\sigma}, v\right\rangle\right| \right].
\end{equation*}
We denote $\overline{\mathfrak{R}}(\mathcal{V}) := \overline{\mathfrak{R}}(\mathcal{V}, p_0)$ for $p_0 = \frac{MU}{(M+U)^2}$. For $\mathcal{F}\subset \{\mathcal{X}\to \widehat{\mathcal{Y}}\}$, we denote $\overline{\mathfrak{R}}(\mathcal{F}, p) := \overline{\mathfrak{R}}(\mathcal{V}, p)$ where $\mathcal{V} = \{(f(X_1), \ldots, f(X_N)))^{\top}\mid f\in \mathcal{F}\}$, where $X_1, \ldots X_N$ are feature vectors of the given training dataset defined in Section~\ref{sec:optimization}.
\end{definition}
We refer to the transductive Rademacher complexity defined in Definition~\ref{def:transductive-rademacher-complexity} as the \textit{unsymmetrized} transductive Rademacher complexity if necessary\footnote{We are not aware the standard notion used to tell apart the complexities defined in Definitions~\ref{def:transductive-rademacher-complexity} and ~\ref{def:symmetrized-transductive-rademacher-complexity}. The notion of \textit{(un)symmetrized} is specific to this paper.}.
Note that we have by definition
\begin{align}
    \mathfrak{R}(\mathcal{V}, p) &\leq \overline{\mathfrak{R}}(\mathcal{V}, p) \label{eq:inequality-between-rademacher-complexity-variants}.
\end{align}
Using the concept of the symmetrized transductive Rademacher complexity, we state the main proposition of this section.
\begin{proposition}\label{prop:rademacher-complexity-of-mlp-general}
    Let $p\in [0, 1/2]$.
    Suppose we use $\mathcal{G}^{(t)}$ and $\mathcal{B}^{(t)}$ defined by~\eqref{eq:parallel-transformation} and~\eqref{eq:definition-base-mlp} as a model.
    Define $D^{(t)} = 2\sqrt{2}(2\tilde{B}^{(t)})^{L-1}\prod_{s=2}^{t}\tilde{C}^{(s)}$ and $P^{(t)}:=\prod_{s=2}^{t}\tilde{P}^{(s)}$.
    Then, we have
    \begin{equation*}
        \overline{\mathfrak{R}}(\mathcal{F}^{(t)}, p)\leq \sqrt{2p}QB^{(t)}\|P^{(t)} X\|_{\F}.
    \end{equation*}
\end{proposition}
We shall prove this proposition in the end of this section.
Reference \cite{el2009transductive} proved the contraction property of the unnsymmetrized Rademacher complexity. We prove the contraction property for the symmetrized variant.
\begin{proposition}\label{prop:contraction-of-rademacher-complexity}
    Let $\mathcal{V} \subset \mathbb{R}^{N}$, $p\in [0, 1/2]$.
    Suppose $\rho:\mathbb{R}\to\mathbb{R}$ is $L_\rho$-Lipschitz and $\rho(0)=0$. Then, we have
    \begin{equation*}
        \overline{\mathfrak{R}}(\rho\circ \mathcal{V}, p) \leq 2L_{\rho}\overline{\mathfrak{R}}(\mathcal{V}, p),
    \end{equation*}
    where $\rho \circ \mathcal{V} := \{(\rho(v_1), \ldots, \rho(v_N))^\top\mid \bm{v} = (v_1, \ldots, v_N)^\top\in \mathcal{V}\}$.
    \begin{proof}
        First, by the definition $\overline{\mathfrak{R}}$ and $\rho(0) = 0$, we have
        \begin{align*}
            &\overline{\mathfrak{R}}(\mathcal{V}\cup \{0\}, p)
            = \overline{\mathfrak{R}}(\mathcal{V}, p),\\
            &\overline{\mathfrak{R}}(\rho \circ (\mathcal{V}\cup \{0\}), p)
            = \overline{\mathfrak{R}}((\rho \circ \mathcal{V}) \cup \{0\}, p)
            = \overline{\mathfrak{R}}(\rho \circ \mathcal{V}, p).
        \end{align*}
        Therefore, we can assume without loss of generality that $0\in \mathcal{V}$.
        Then, we have
        \begin{align}
            \overline{\mathfrak{R}}(\rho \circ \mathcal{V}, p)
            &= Q\E_{\bm{\sigma}} \sup_{v\in \mathcal{V}} |\langle \bm\sigma, \rho(v)\rangle| \notag\\
            &\leq Q\E_{\bm{\sigma}} \sup_{v\in \mathcal{V}} \langle \bm\sigma, \rho(v)\rangle
                + Q\E_{\bm{\sigma}} \sup_{v\in \mathcal{V}} \langle \bm\sigma, -\rho(v)\rangle \notag\\
            &=\mathfrak{R}(\rho\circ \mathcal{V}, p)
                + \mathfrak{R}(-\rho\circ \mathcal{V}, p),\label{eq:contraction-1}
        \end{align}
        where $\bm{\sigma} = (\sigma_1, \ldots, \sigma_N)$ are the i.i.d. transducive Rademacher variables of parameter $p$.
        We used $0\in \mathcal{V}$ and $\rho(0) = 0$ in the inequality above.
        By the contraction property of the unsymmetrized transductive Rademacher complexity (\cite[Lemma 1]{el2009transductive}), we have
        \begin{align*}
            &\mathfrak{R}(\rho\circ \mathcal{V}, p) \leq L_\rho \mathfrak{R}(\mathcal{V}, p)\\
            &\mathfrak{R}(-\rho\circ \mathcal{V}, p) \leq L_\rho \mathfrak{R}(\mathcal{V}, p)
        \end{align*}
        By combining them with \eqref{eq:inequality-between-rademacher-complexity-variants} and \eqref{eq:contraction-1}, we have
        \begin{align*}
            \overline{\mathfrak{R}}(\rho \circ \mathcal{V}, p)
            \leq 2L_\rho \mathfrak{R}(\mathcal{V}, p)
            \leq 2L_\rho \overline{\mathfrak{R}}(\mathcal{V}, p).
        \end{align*}
    \end{proof}
\end{proposition}

\begin{proof}[Proof of Proposition~\ref{prop:rademacher-complexity-of-mlp-general}]
The proof is the extension of~\cite[Exercises 3.11]{mohri2018foundations} to the transductive and multi-layer setting.
See also the proof of \cite[Theorem 3]{pmlr-v80-nitanda18a}.
First, we note that the multiplication $X\mapsto \tilde{P}^{(s)}X$ in $\mathcal{G}^{(s)}$ and the multiplication $X\mapsto XW^{(s-1)}\in \mathbb{R}^{C\times C}$ in $\mathcal{G}^{(s-1)}$ are commutative operations.
Therefore, we have
\begin{equation*}
    \mathcal{F}^{(t)}(X) := \{(\bm{z}_1,\ldots, \bm{z}_N) \mid f\in \mathcal{B}_{\mathrm{base}}^{(t)}, \|W^{(s)}_{\cdot c}\|_1 \leq C^{(s)} \text{\ for all\ } c\in [C] \text{\ and\ } s=2, \ldots, t \},
\end{equation*}
where $\bm{z}_n := f(\bm{x}_{n}W^{(2)} \cdots W^{(t)})$ and $\bm{x_n} := (P^{(t)}X)_n \in \mathbb{R}^C$.
Therefore, it is sufficient that we first prove the proposition by assuming $\tilde{P}^{(s)} = I_N$ for all $s=2, \ldots, t$ and then replace $X$ with $P^{(t)}X$.

We define $\mathcal{J}^{(s)} \subset \mathbb{R}^N$ be the set of possible values of any channel of the $s$-th representations and $\mathcal{H}^{(l)} \subset \mathbb{R}^N$ be the set of possible values of any output channel of the $l$-th layer of an MLP. More concretely, we define
\begin{align*}
    \mathcal{J}^{(1)} &:= \{X_{\cdot c} \mid c\in [C]\},\\
    \mathcal{J}^{(s+1)} &:= \left\{\sum_{c=1}^C \bm{z}_c w_c
        \,\middle\vert\, \bm{z}_c \in \mathcal{J}^{(s)}, \|w\|_1 \leq \tilde{C}^{(s+1)}\right\},
\end{align*}
for $s=1, \ldots t-1$.
Similarly, we define
\begin{align*}
    \mathcal{H}^{(1)} &:= \mathcal{J}^{(t)},\\
    \tilde{\mathcal{H}}^{(l+1)} &:=
        \left\{ \sum_{c=1}^{C_{l+1}} \bm{z}_cw_{c} 
            \,\middle\vert\, \bm{z}_c \in \mathcal{H}^{(l)}, \|w\|_{1} \leq \tilde{B}^{(t)} \right\}, \\
    \mathcal{H}^{(l+1)} &:= \sigma \circ \tilde{\mathcal{H}}^{(l+1)} = \{\sigma(\bm{z}) \mid \bm{z} \in \tilde{\mathcal{H}}^{(l+1)} \}.
\end{align*}
for $l=1, \ldots, L$.
By the definition of $\mathcal{F}^{(t)}$, we have $\{f(X) \mid f \in \mathcal{F}^{(t)}\} = \tilde{\mathcal{H}}^{(L+1)}$.
On one hand, we can bound the Rademacher complexity of $\tilde{\mathcal{H}}^{(l)}$ as
\begin{align}
    Q^{-1}\overline{\mathfrak{R}}(\tilde{\mathcal{H}}^{(l+1)}, p)
    &= \E_{\bm{\sigma}}
        \left[
            \sup_{\|w\|_1 \leq \tilde{B}^{(l)}, Z_{\cdot c}\in \mathcal{H}^{(l)}} \left|\sum_{n=1}^N \sigma_n \sum_{c=1}^{C_{l+1}} Z_{nc} w_c\right|
        \right] \notag\\
    &= \E_{\bm{\sigma}}
        \left[
            \sup_{\|w\|_1 \leq \tilde{B}^{(l)}, Z_{\cdot c}\in \mathcal{H}^{(l)}} \left|\sum_{c=1}^{C_{l+1}} w_c \sum_{n=1}^N \sigma_n  Z_{nc}\right|
        \right] \notag\\        
    &= \tilde{B}^{(t)} \E_{\bm{\sigma}}
        \left[
            \sup_{Z\in \mathcal{H}^{(l)}} \left|\sum_{n=1}^N \sigma_n Z_{n}\right|
        \right] \notag\\
    &= \tilde{B}^{(t)} Q^{-1}\overline{\mathfrak{R}}(\mathcal{H}^{(l)}, p).\label{eq:h-tilde-to-h}
\end{align}

On the other hand, since $\sigma$ is $1$-Lipschitz, by the contraction property (Proposition~\ref{prop:contraction-of-rademacher-complexity}), we bound the Rademacher complexity of $\mathcal{H}^{(l+1)}$ as
\begin{equation}\label{eq:h-to-h-tilde}
    \overline{\mathfrak{R}}(\mathcal{H}^{(l+1)}, p) \leq 2\overline{\mathfrak{R}}(\tilde{\mathcal{H}}^{(l+1)}, p).
\end{equation}
By combining \eqref{eq:h-tilde-to-h} and \eqref{eq:h-to-h-tilde}, we have the inductive relationship.
\begin{align}
    \overline{\mathfrak{R}}(\mathcal{H}^{(l+1)}, p)
    \leq 2\tilde{B}^{(t)} \overline{\mathfrak{R}}(\mathcal{H}^{(l)}, p).\label{eq:rademacher-complexity-inductive-step}
\end{align}

Using the similar argument to $\mathcal{J}^{(s)}$'s, for $s\in 2, \ldots, t-1$, we have
\begin{equation}\label{eq:rademacher-complexity-inductive-step-for-g}
    \overline{\mathfrak{R}}(\mathcal{J}^{(s+1)}, p)
    \leq \tilde{C}^{(s)} \overline{\mathfrak{R}}(\mathcal{J}^{(s)}, p).
\end{equation}

Let $P_c\in \mathbb{R}^{C}$ be the projection matrix onto the $c$-th coordinate.
Then, for the base step, we can evaluate the Rademacher complexity of $\mathcal{J}^{(1)}$ as
\begin{align}
    Q^{-1}\overline{\mathfrak{R}}(\mathcal{J}^{(1)}, p)
    & = \E_{\bm{\sigma}} \left[ \max_{c\in [C]} \left|\sum_{n=1}^N \sigma_n X_{nc} \right|\right] \notag\\
    & = \E_{\bm{\sigma}} \left[ \max_{c\in [C]}  \left|\left(\sum_{n=1}^N \sigma_n X_{n}\right)P_c\right| \right] \notag\\
    & \leq \E_{\bm{\sigma}} \left[ \max_{c\in [C]} \|P_c\|_{\op}\left\| \sum_{n=1}^N \sigma_n X_{n}\right\| _2 \right] \notag\\    
    & \leq \E_{\bm{\sigma}} \left\| \sum_{n=1}^N \sigma_n X_{n}\right\|_2 \notag\\
    & \leq \sqrt{\E_{\bm{\sigma}} \sum_{c=1}^C \left( \sum_{n=1}^N \sigma_n X_{nc}\right)^2 } \text{($\because$ Jensen's inequality)} \notag\\
    & = \sqrt{\E_{\bm{\sigma}} \sum_{c=1}^C \sum_{n, m=1}^N \sigma_n\sigma_m  X_{nc}X_{mc}} \notag\\
    & = \sqrt{\sum_{c=1}^C \sum_{n=1}^N 2p (X_{nc})^2 } \label{eq:reduction-of-rademacher-variable}\\
    & = \sqrt{2p}\|X\|_{\F}.\label{eq:rademacher-complexity-base-step}
\end{align}
Here, we used in \eqref{eq:reduction-of-rademacher-variable} the equality
\begin{equation}\label{eq:expectation-of-rademacher-variable}
    \E_{\bm{\sigma}} \sigma_m \sigma_n = 2p\delta_{mn},
\end{equation}
which is shown by the independence of transductive Rademacher variables.
By using the inequalities we have proved so far, we obtain
\begin{align*}
    \overline{\mathfrak{R}}(\mathcal{F}^{(t)}, p)
    &= \overline{\mathfrak{R}}(\tilde{\mathcal{H}}^{(L+1)}, p)\\
    &= 2\overline{\mathfrak{R}}(\mathcal{H}^{(L)}, p)
        \quad \text{($\because$ \eqref{eq:h-to-h-tilde} )}\\
    &\leq 2(2\tilde{B}^{(t)})^{L-1} \overline{\mathfrak{R}}(\mathcal{H}^{(1)}, p)
        \quad \text{($\because$ \eqref{eq:rademacher-complexity-inductive-step})}\\
    &\leq 2(2\tilde{B}^{(t)})^{L-1} \overline{\mathfrak{R}}(\tilde{\mathcal{J}}^{(t)}, p)
        \quad \text{($\because$ Definition of $\mathcal{H}^{(1)}$) }\\
    &\leq 2(2\tilde{B}^{(t)})^{L-1} \left(\prod_{s=2}^{t}\tilde{C}^{(s)}\right) \overline{\mathfrak{R}}(\tilde{\mathcal{J}}^{(1)}, p)
        \quad \text{($\because$ \eqref{eq:rademacher-complexity-inductive-step-for-g})}\\
    &\leq \sqrt{p}QD^{(t)}\|X\|_{\F},
        \quad \text{($\because$ \eqref{eq:rademacher-complexity-base-step})}\\
\end{align*}
where we used $D^{(t)} = 2\sqrt{2}(2\tilde{B}^{(t)})^{L-1}\prod_{s=2}^{t}\tilde{C}^{(s)}$.
Therefore, the proposition is true for $\tilde{P}^{(s)} = I_N$ for all $s=2, \ldots, t$.
As stated in the beginning of the proof, we should replace $X$ with $P^{(t)}X$ in the general case.
\end{proof}

\begin{proof}[Proof of Proposition~\ref{prop:rademacher-complexity-of-mlp}]
    By applying Proposition~\ref{prop:rademacher-complexity-of-mlp-general} with $p = p_0= \frac{MU}{(M+U)^2}$ and using \eqref{eq:inequality-between-rademacher-complexity-variants}, we have
    \begin{align*}
        \mathfrak{R}(\mathcal{F}^{(t)}) 
        \leq \overline{\mathfrak{R}}(\mathcal{F}^{(t)}, p_0)
        \leq \sqrt{\frac{MU}{(M+U)^2}}QD^{(t)}\|P^{(t)}X\|_\F
        = \frac{D^{(t)}\|P^{(t)}X\|_\F}{\sqrt{MU}}.
    \end{align*}
\end{proof}

%% file: source/appendix/proofs/proof_generalization_bound_for_multi_scale_gnns.tex
\subsection{Proof of Theorem~\ref{thm:monotonically-decreasing-test-error}}\label{sec:proof-test-error-for-multi-scale-gnns}

\begin{proof}
By Proposition~\ref{prop:generalization-gap}, with probability $1-\delta'$, we have 
\begin{align}\label{eq:thm2-1}
    \mathcal{R}(\widehat{Y})
    \leq \widehat{\mathcal{R}}(\widehat{Y}) + \sum_{t=1}^T \eta^{(t)}\mathfrak{R}(\mathcal{F}^{(t)})
        + c_0Q\sqrt{M\wedge U} + \sqrt{\frac{SQ}{2}\log \frac{1}{\delta'}}.
\end{align}
By Theorem~\ref{thm:optimization}, we have
\begin{align}\label{eq:thm2-2}
    \widehat{\mathcal{R}}(\widehat{Y})
    \leq  \frac{(1 + e^{\delta})\widehat{\mathcal{L}}(\widehat{Y}^{(1)})}{2M\Gamma_T}.
\end{align}
By Proposition\ref{prop:rademacher-complexity-of-mlp}, we have
\begin{equation}\label{eq:thm2-4}
    \mathfrak{R}(\mathcal{F}^{(t)})
    \leq \frac{D^{(t)}\|P^{(t)}X\|_{\F}}{\sqrt{MU}}
\end{equation}
By applying \eqref{eq:thm2-2} and \eqref{eq:thm2-4} to \eqref{eq:thm2-1} and substituting the learning rate $\eta^{(t)} = \frac{4}{\alpha_{t}}$, we obtained~\eqref{eq:test-error-bound}.
In particular, when $\Gamma_T = O(T^\varepsilon)$, the first term of \eqref{eq:test-error-bound} is $O(T^{-\varepsilon})$, which is asymptotically monotonically decreasing (assuming $\delta$, $\widehat{Y}^{(1)}$, and $M$ is independent of $T$).
When $\alpha_t^{-1}B^{(t)}\|P^{(t)}\|_{\op}^t = O(\tilde{\varepsilon}^{t})$, the second term of~\eqref{eq:test-error-bound} is bounded by
\begin{align*}
    \frac{4}{\sqrt{MU}}\sum_{t=1}^T\frac{D^{(t)}\|P^{(t)}X\|_{\F}}{\alpha_t}
    &\leq \frac{4\sqrt{2}\|X\|_\F}{\sqrt{MU}}\sum_{t=1}^T\frac{B^{(t)}\|P^{(t)}\|_{\op}}{\alpha_t}\\
    &\lesssim \frac{\|X\|_\F}{\sqrt{MU}}\frac{1}{1-\tilde{\varepsilon}}.
\end{align*}
The upper bound is independent of $T$ (assuming that $\|X\|_{\F}$, $M$, and $U$ are independent of $T$).
\end{proof}

%% file: source/appendix/proofs/proof_bound_by_inductive_rademacher_complexity.tex
\subsection{Proof of Proposition~\ref{prop:bound-by-inductive-rademacher-complexity}}\label{sec:proof-bound-by-inductive-rademacher-complexity}

First, we recall the usual (i.e., inductive) version of the Rademacher complexity.
We employ the following definition.
\begin{definition}[(Inductive) Empirical Rademacher Complexity]~\label{def:inductive-rademacher-complexity}
    For $\mathcal{F}_{\mathrm{base}} \subset \{ \mathcal{X} \to \widehat{\mathcal{Y}} \}$ and $Z = (\bm{z}_1, \ldots, \bm{z}_N) \in \mathcal{X}^N$, we define the (inductive) empirical Rademacher complexity $\widehat{\mathcal{\mathfrak{R}}}_{\mathrm{ind}}(\mathcal{F}_{\mathrm{base}})$  conditioned on $Z$ by
    \begin{equation*}
        \widehat{\mathcal{\mathfrak{R}}}_{\mathrm{ind}}(\mathcal{F}_{\mathrm{base}}; Z) := \frac{1}{N} \E_{\bm{\varepsilon}}\left[\sup_{f\in \mathcal{F}_{\mathrm{base}}} \sum_{n=1}^N \varepsilon_n f(\bm{z}_n) \right],
    \end{equation*}
    where $\bm{\varepsilon} = (\varepsilon_1, \ldots, \varepsilon_N)$ is the i.i.d. Rademacher variables defined by $\mathbb{P} (\varepsilon_i = 1) = \mathbb{P} (\varepsilon_i = -1) = 1/2$.
\end{definition}

    \begin{proof}[Proof of Proposition~\ref{prop:bound-by-inductive-rademacher-complexity}]
        Similarly to Proposition~\ref{prop:rademacher-complexity-of-mlp-general} it is sufficient that we first prove the proposition by assuming $\tilde{P}^{(s)} = I_N$ for all $s=2, \ldots, t$ and then replace $X$ with $P^{(t)}X$.
        By definition, the transductive Rademacher variable of parameter $p=1/2$ equals to the (inductive) Rademacher variable. Therefore, we have
        \begin{align}
            Q^{-1} \mathfrak{R}(\mathcal{F}^{(t)}, 1/2)
            &= \E_{\bm{\sigma}} \left[\sup_{f\in \mathcal{F}^{(t)}} \sum_{n=1}^N \sigma_n f(X)_n \right] \notag\\
            &= \E_{\bm{\sigma}} \left[\sup_{f_{\mathrm{base}}\in \mathcal{B}^{(t)}_{\mathrm{base}}, \|W^{(s)}\|_1\leq \tilde{C}^{(s)}} \sum_{n=1}^N \sigma_n f_{\mathrm{base}}(XW^{(2)}\cdots W^{(t)}) \right] \notag\\
            &= N\widehat{\mathcal{\mathfrak{R}}}_{\mathrm{ind}}(\mathcal{F}_{\mathrm{base}}^{(t)}; X). \label{eq:transductive-to-inductive}
        \end{align}
        Since $p_0 := \frac{MU}{(M+U)^2} < 1/2$, by the monotonicity of the transductive Rademacher complexity (see ~\cite{el2009transductive} Lemma 1), we have 
        \begin{align}\label{eq:p-zero-to-one-half}
            \mathfrak{R}(\mathcal{F}^{(t)})
            = \mathfrak{R}(\mathcal{F}^{(t)}, p_0)
            < \mathfrak{R}(\mathcal{F}^{(t)}, 1/2).
        \end{align}
        The proposition follows from \eqref{eq:transductive-to-inductive} and \eqref{eq:p-zero-to-one-half} as follows
        \begin{align*}
            \mathfrak{R}(\mathcal{F}^{(t)})
            < QN\widehat{\mathcal{\mathfrak{R}}}_{\mathrm{ind}}(\mathcal{F}^{(t)}_{\mathrm{base}}; X)
            = \frac{(1+r)^2}{r} \widehat{\mathcal{\mathfrak{R}}}_{\mathrm{ind}}(\mathcal{F}^{(t)}_{\mathrm{base}}; X)
        \end{align*}
    \end{proof}

%% file: source/appendix/proofs/proof_trade_off_between_wlc_and_rademacher_complexity.tex
\subsection{Proof of Proposition \ref{prop:trade-off-between-wlc-and-rademacher-complexity}}\label{sec:proof-of-trade-off-between-wlc-and-rademacher-complexity}

\begin{proof}
We denote $p_0 = \frac{MU}{(M+U)^2}$.
Let $\sigma_1, \ldots, \sigma_N$ be the i.i.d. transductive Rademacher variable of parameter $p_0$.
Since $\{-1, 0, 1\}^N\subset \mathcal{V}_{g}$, for any realization of $\bm{\sigma} = (\sigma_1, \ldots, \sigma_N)$, we have $\bm{\sigma} \in \mathcal{V}_{g}$.
By the assumption, there exists $Z_{\bm{\sigma}}\in\mathcal{V}$ such that
\begin{equation*}
    \|Z_{\bm{\sigma}} - \alpha\bm{\sigma}\|_2 \leq \beta \|\bm{\sigma}\|_2.
\end{equation*}
Set $C_f := (2\alpha)^{-1}$ and $C_\mathcal{L} := \frac{\alpha^2 - \beta^2}{2\alpha}$.
Similarly to the proof of Theorem \ref{thm:optimization}, we have
\begin{align*}
    & \|Z_{\bm{\sigma}} - \alpha \bm{\sigma}\|_2 \leq \beta \|\bm{\sigma}\|_2\\
    \Longleftrightarrow \ &
        \|Z_{\bm{\sigma}}\|_{2}^2
            - 2\alpha \langle Z_{\bm{\sigma}}, \bm{\sigma}\rangle
            + \alpha^2 \|\bm{\sigma}\|^2_2
        \leq \beta^2 \|\bm{\sigma}\|^2_2\\
    \Longleftrightarrow \ &
         C_f \|Z_{\bm{\sigma}}\|^2_{2} + C_{\mathcal{L}}\|\bm{\sigma}\|^2_{2}
        \leq \langle Z_{\bm{\sigma}}, \bm{\sigma}\rangle.
\end{align*}
Therefore, we have
\begin{align*}
    Q^{-1} \mathfrak{R}(\mathcal{V})
    &= \E_{\bm{\sigma}} \left[ \sup_{Z\in \mathcal{F}} \langle \bm{\sigma}, Z \rangle \right] \\
    &\geq \E_{\bm{\sigma}} \left[ \langle \bm{\sigma}, Z_{\bm{\sigma}} \rangle \right] \\
    &\geq \E_{\bm{\sigma}} \left[ C_{\mathcal{L}} \|\bm{\sigma}\|^2_2 + C_f \|f_{\bm{\sigma}}\|^2_2 \right]\\
    &\geq \E_{\bm{\sigma}} \left[ C_{\mathcal{L}} \|\bm{\sigma}\|^2_2 \right]\\
    &= C_{\mathcal{L}} \cdot 2Np_0.
\end{align*}
In the last equality, we used \eqref{eq:expectation-of-rademacher-variable}.
Therefore, we have $\mathfrak{R}(\mathcal{F})\geq 2C_{\mathcal{L}}Np_0Q = \frac{\alpha^2 - \beta^2}{\alpha}$.
\end{proof}

%% file: source/appendix/more_related_works.tex
\section{More Related Work}\label{sec:more-related-work}

\paragraph{Generalization Gap Bounds of GNNs}

Reference \cite{scarselli2018vapnik} derived the upper bound of the VC dimension of GNNs. However, the derivation is specific to their model and does not apply to other GNNs.
Reference \cite{NIPS2019_8809} incorporated the idea of Neural Tangent Kernels~\cite{NIPS2018_8076} and derived a generalization gap by reducing it to a kernel regression problem. However, they considered graph prediction problems, where each sample point itself is represented as a graph drawn from some distribution, while our problem is a node prediction problem.
Reference \cite{verma2019stability} derived the generalization gap bounds for node prediction tasks using the stability argument.
However, they only considered a GNN with a single hidden layer. It is not trivial to extend their result to multi-layered and multi-scale GNNs.
Similarly to our study,~\cite{garg2020generalization} employed the (inductive) Rademacher complexity. However, because they did not discuss the optimization guarantee, we cannot directly derive the test error bounds from their analysis.

%% file: source/appendix/wlc_by_overparameterized_model.tex
\section{Provable Satisfiability of Weak Learning Condition using Over-parameterized Models}\label{sec:wlc-by-overparameterized-model}

In this section, we show that there exists a model that for any w.l.c.\ parameters $\alpha$ and $\beta$, we can find a weak learner which \textit{probably} satisfies the w.l.c.\ using the gradient descent algorithm.
To ensure the w.l.c., the set of transformation functions $\mathcal{B}$ must be sufficiently large so that it can approximate all possible values of the negative gradient $-\nabla \widehat{\mathcal{L}}$ can take.
We can accomplish it by leveraging the universal approximation property of MLPs, similarly to graph isomorphism networks (GIN)~\cite{xu2018how}, but for a different purpose.
We adopt the recent studies that proved the global convergence of over-parameterized MLPs trained by a tractable algorithm (e.g.,~\cite{pmlr-v97-arora19a,du2018gradient}).

Let $R\in \mathbb{N}_+$.
For the parameter $\Theta = (\theta_{ri}) \in \mathbb{R}^{R \times N}$, we consider an MLP with a single hidden layer $f_\Theta: \mathbb{R}^C\to \mathbb{R}$ defined by
\begin{equation*}
    f_{\Theta}(\bm{x}) := \frac{1}{\sqrt{R}} \sum_{r=1}^R a_{r} \sigma (\theta_{r\cdot}^\top \bm{x}),
\end{equation*}
where $a_{r}\in \{-1, 1\}$ and $\sigma$ is the ReLU activation function: $\sigma(x) := x\vee 0$ (we apply ReLU in an element-wise manner for a vector input).
We define the set of transformation functions $\mathcal{B}:= \{(f_\Theta, \ldots, f_\Theta) \mid \Theta \in \mathbb{R}^{R\times N} \}$.
At the $t$-th iteration, given a gradient $\nabla\widehat{\mathcal{L}}(\widehat{Y}^{(t-1)})$, we initialize the model with $a_{r}\overset{\text{i.i.d.}}{\sim}\mathrm{Unif}(\{-1, 1\})$ and $\theta_{ri} \overset{\text{i.i.d.}}{\sim} \mathcal{N}(0, I)$ independently and train it with the gradient descent to optimize $\Theta$ by minimizing the mean squared error between the output of the model and the properly normalized negative gradient.
Using the result of~\cite{du2018gradient}, we obtain the following guarantee.
\begin{proposition}\label{prop:wlc_by_overparameterized_model}
    Suppose Algorithm \ref{alg:train} finds $g^{(s)} \in \mathcal{G}^{(s)}$ ($s\in [t]$) such that $X^{(t)} = g^{(t)} \circ \cdots \circ g^{(1)}(X) = \begin{bmatrix}\bm{x}_1 & \cdots & \bm{x}_N \end{bmatrix}^\top\in \mathbb{R}^{N\times C}$ ($\bm{x}_i\in \mathbb{R}^C$) satisfies the conditions that $\bm{x}_i\not =0$ for all $i\in [N]$ and $\bm{x}_{i} \nparallel \bm{x}_{j}$ for all $i\not = j \in [N]$.
    Let $\delta, \alpha, \beta > 0$.
    Then, there exists $R = O(N^6\delta^{-3})$ such that for all $\nabla \widehat{\mathcal{L}}(\widehat{Y}^{(t-1)})$, with a probability of at least $1-\delta$, the gradient descent algorithm finds $b^{(t)}\in \mathcal{B}^{(t)}$ such that the $t$-th weak learner $f^{(t)}$ satisfies the $(\alpha, \beta, -\nabla \widehat{\mathcal{L}}(\widehat{Y}^{(t-1)}))$-w.l.c.
\end{proposition}

\begin{proof}
    We define $H^\infty\in \mathbb{R}^{N\times N}$ by $H^\infty_{ij} := \mathbb{E}_{\bm{w}\sim \mathcal{N}(0, I)}[\bm{x}_i^\top \bm{x}_j \bm{1}\{\bm{w}^\top \bm{x}_i\geq 0\} \bm{1}\{\bm{w}^\top \bm{x}_j \geq 0\} ]$.
    Let $\lambda_0$ be the lowest eigenvalue of $H^\infty$.
    Under the assumption, we know $\lambda_0 > 0$ by Theorem 3.1 of~\cite{du2018gradient}.
    We train the parameter $\Theta_{\cdot c\cdot}$ using the dataset $((\bm{x}_1, -\alpha[\widehat{\mathcal{L}}(\widehat{Y}^{(t-1)}]_{1}), \ldots, (\bm{x}_N, -\alpha[\widehat{\mathcal{L}}(\widehat{Y}^{(t-1)})]_{N})$.
    We denote the parameter of the MLP at the $k$-th iteration of the gradient descent by $\Theta^{(k)}$.
    We denote the output of the model $f_{\Theta^{(k)}}$ by $\bm{u}^{(k)} := (f_{\Theta^{(k)}}(\bm{x}_1), \ldots, f_{\Theta^{(k)}}(\bm{x}_N))^{\top}$.
    By~\cite[Theorem 4.1]{du2018gradient}, with probability $1-\delta$, we have
    \begin{equation*}
        \|\bm{u}^{(k)} + \alpha\nabla\widehat{\mathcal{L}}(\widehat{Y}^{(t-1)})\|_2
        \leq \left(1 - \frac{\eta\lambda_0}{2}\right)^k
        \|\bm{u}^{(k)} + \alpha\nabla \widehat{\mathcal{L}}(\widehat{Y}^{(t-1)}) \|_2,
    \end{equation*}
    where $\eta = O\left(\frac{\lambda_0}{N^2}\right)$.
    Set
    \begin{equation*}
        k := \log\left(\frac{\beta \|\nabla \widehat{\mathcal{L}}(\widehat{Y}^{(t-1)})\|_{2}}{\|\bm{u}^{(0)} + \alpha\nabla \widehat{\mathcal{L}}(\widehat{Y}^{(t-1)})\|_2\vee 1}\right)
        \left(\log \left(1 - \frac{\eta\lambda_0}{2}\right)\right)^{-1}.
    \end{equation*}
    Then, with probability $1-\delta$, we have
    \begin{equation*}
        \|\bm{u}^{(k)} + \alpha \nabla \widehat{\mathcal{L}}(\widehat{Y}^{(t-1)})\|_{2} \leq \beta \|\nabla \widehat{\mathcal{L}}(\widehat{Y}^{(t-1)})\|_{2},
    \end{equation*}
    which means $\bm{u}^{(k)}$ satisfies $(\alpha, \beta, -\nabla\widehat{\mathcal{L}}(\widehat{Y}^{(t-1)}))$-w.l.c.
\end{proof}

\begin{remark}
    Reference \cite{du2018gradient} assumed that any feature vector $\bm{x}$ of the training data satisfies $\|\bm{x}\| = 1$. 
    However, as commented in~\cite{du2018gradient}, we can loosen this condition as follows: there exists $c_{low}, c_{high} > 0$ such that any feature vector $\bm{x}$ satisfies $c_{low} \leq \|\bm{x}\|\leq c_{high}$.
\end{remark}
Although this instantiation provably satisfies the w.l.c.\ with high probability, its model complexity is extremely large because it has as many as $O(N^6)$ parameters.
As we saw in Section~\ref{sec:discussion}, such a large model complexity is inevitable as long as the gradient can take arbitrary values.

%% file: source/appendix/more_model_variants.tex
\section{More Model Variants}\label{sec:more-model-variants}

\paragraph{Input Injection}

A matrix $P\in \mathbb{R}^{N\times N}$ defines the aggregation model $\mathcal{G}_P:=\{X\mapsto PX\}$ as we did in Section~\ref{sec:experiment}.
Typical choices of $P$ are the (normalized) adjacency matrix, a GCN-like augmented normalized adjacency matrix, or the (normalized) graph Laplacian.
If we choose $P':= \rho P + (1-\rho)I_N$ for some $\rho\in[0, 1]$, it aggregates the representations in a lazy manner using $P$.
In the similar spirit of~\cite{pmlr-v108-nitanda20a,zhang2019gresnet}, there is another type of lazy aggregation that allows us to inject the information of unmixed features directly to the representations.
Specifically, for $\rho \in [0, 1]$, we define the \textit{input injection} model $\mathcal{G}_{\mathrm{II}}(\rho, P)$ by
\begin{equation*}
    \mathcal{G}_{\mathrm{II}}(\rho, P) := \{X \mapsto \rho PX + (1-\rho)X^{(1)}\}.
\end{equation*}
On one hand, $\mathcal{G}_{\mathrm{II}}(\rho, P)$ equals $\mathcal{G}_{P}$ when $\rho = 1$.
On the other hand, when $\rho=0$, $\mathcal{G}_{\mathrm{II}}(\rho, P)$ ignores the effect of the representation mixing and employs the original features.
We can identify $\mathcal{G}_\mathrm{II}(\rho, P)$ with $\{(X, X') \mapsto (\rho PX + (1-\rho)X', X') \}$.
Therefore, if we redefine a new input space as $\mathcal{X}' := \mathcal{X} \times \mathcal{X}$, which means that we double the input channel size, and preprocess features as $x_i \mapsto (x_i, x_i)$ for each $i\in V$, we can think the input injection model as an example of our model.
For the augmented normalized adjacency matrix $\tilde{A}$, we refer to the model that uses $\mathcal{G}_{\mathrm{II}}(\rho, \tilde{A})$ as the set of aggregation functions $\mathcal{G}^{(t)}$ and the set of MLPs as $\mathcal{B}^{(t)}$ for all $t$ as GB-GNN-II.
We conducted the same experiment as the one we did in Section~\ref{sec:experiment} using GB-GNN-II. The result is reported in Section~\ref{sec:experiment-additional-results}.

%% file: source/appendix/experiment_details.tex
\section{Details of Experiment Settings}\label{sec:experiment-details}

\subsection{Dataset}
We used the Cora~\cite{mccallum2000automating,sen2008collective}, CiteSeer~\cite{giles1998citeseer, sen2008collective}, and PubMed~\cite{sen2008collective} datasets.
Each dataset represents scientific papers as the nodes and citation relationships as the edges of a graph.
For each paper, the genre of this paper is associated as a label.
The task is to predict the genre of papers from word occurrences and the citation relationships.
Table~\ref{tbl:dataset_specifications} shows the statistics of datasets.
We obtained the preprocessed dataset from the code repository of~\cite{kipf2017iclr} (\url{https://github.com/tkipf/gcn}) and split each dataset into train, validation, and test datasets in the same way as experiments in~\cite{kipf2017iclr}.

\input{table/dataset_specification}

\subsection{Model}~\label{sec:model-details}

As shown in Table~\ref{tbl:node_classification_on_citation_network} in Section~\ref{sec:experiment-additional-results}, we have tested four base models: GB-GNN-Adj, GB-GNN-KTA, GB-GNN-II, and GB-GNN-SAMME.R.
For each model, we consider three types of variants: (1) the base model, (2) the model with fine tuning, and (3) models with different layer sizes ($L=0, 2, 3, 4$),
We have shown the result of variants (1) and (2) of GB-GNN-Adj and GB-GNN-KTA in the main paper.
See Section~\ref{sec:experiment-additional-results} for the results of other models.

\subsubsection{Node Aggregation Functions}

For the aggregation functions $\mathcal{G}$, we used the matrix multiplication model with the augmented normalized adjacency matrix $\tilde{A}$ of the underlying graph $\mathcal{G}_{\tilde{A}}$ (for GB-GNN-Adj) and the KTA model $\mathcal{G}_{\mathrm{KTA}}$ (for GB-GNN-KTA).
We also employed the input injection model with the augmented normalized adjacency matrix $\mathcal{G}_{\mathrm{II}}(\rho, \tilde{A})$ defined in Section~\ref{sec:more-model-variants} (for GB-GNN-II).
For the KTA models, we used $\mathcal{G}_{\mathrm{KTA}} := \{g: X\mapsto wX + \sum_{k=0}^{N_{\mathrm{deg}}} w_k\tilde{A}^{2^{k}}X \mid w, w_k\in \mathbb{R} \}$.
We treat weights $w$ and $w_k$'s in $\mathcal{G}_{\mathrm{KTA}}$ as learnable parameters and the mixing parameter $\rho$ of $\mathcal{G}_{\mathrm{II}}(\rho, \tilde{A})$ as a hyperparameter.

\subsubsection{Boosting Algorithms}

We used two boosting algorithms SAMME and SAMME.R.
SAMME is the default boosting algorithm and is applied to GB-GNN-Adj, GB-GNN-KTA, and GB-GNN-II.
We used SAMME.R in combination with the matrix multiplication model $\mathcal{G}_{\tilde{A}}$ only (for GB-GNN-SAMME.R).

\subsubsection{Transformation Functions}

For the transformation functions $\mathcal{B}$, we used MLPs with ReLU activation functions that have $L=0, \ldots, 4$ hidden layers followed by the argmax operation.
We showed results for the $L=1$ model in the main paper.
See Section~\ref{sec:experiment-additional-results} for other models.

The SAMME algorithm assumes that each weak learner outputs one of label categories, while SAMME.R assumes that the probability distribution over the set of categorical labels.
Therefore, we added the argmax operation to the MLPs when we used SAMME and the softmax operation to the MLPs when SAMME.R,

We only imposed soft restrictions on MLPs in the models using regularization methods such as Dropout and weight decay.
This is different from the MLP model defined in Section~\ref{sec:application-to-multi-scale-gnns} in the main paper, which hard-thresholded the norms of weights and bias.

We treat weights in the MLP as trainable parameters and treat architectural parameters (e.g., unit size) other than the layer size $L$ as hyperparameters (see Table~\ref{tbl:hyperparameters} for the complete hyperparameters).

\subsection{Training}\label{sec:training-setting}

We used the SAMME or SAMME.R algorithm to train the model.
At the $t$-th iteration, we give $X^{(t-1)}$ and $Y$ as a set of feature vectors and labels, respectively to the model.
We picked $B$ training sample points randomly and trained the transformation functions $\mathcal{B}^{(t)}$ using them.
We used a gradient-based optimization algorithm to minimize the cross entropy between the prediction of the weak learner and the ground truth labels.
We initialized the model (i.e., MLP) using the default initialization method implemented in PyTorch.

For the aggregation model $\mathcal{G}^{(t)}$, if it does not have a learnable parameter, that is, if $\mathcal{G}^{(t)}$ consists of a single function, we just applied the function to convert $X^{(t-1)}$ into $X^{(t)}$.
For the KTA model, which has learnable parameters $w$ and $w_k$'s, we trained the model $g$ using a gradient-based optimization to maximize the correlation between gram matrices created from transformed features $g(X^{(t-1)})$ and labels $Y$.
The correlation is defined as follows:
\begin{equation*}
    \frac{\langle \mathcal{K}[g(X^{(t-1)})], \mathcal{K}[Y] \rangle}{\|\mathcal{K}[g(X^{(t-1)})]\|_\F \|\mathcal{K}[Y]\|_\F},
\end{equation*}
where $\mathcal{K}$ is the operator that takes the outer product of training sample points defined in Section~\ref{sec:model-variants}.
We initialized weights $w$ and $w_k$'s with $1$.

After the training using boosting algorithm, we optionally trained the whole model as fine-tuning.
When we used SAMME, we replaced the argmax operation in the transformation functions $\mathcal{B}$ with the softmax function along class labels to make the model differentiable.
When we used SAMME.R, we did not change the same architecture in the training and fine tuning phases.
We trained the whole model in an end-to-end manner using a gradient-based optimization algorithm to minimize the cross entropy between the prediction of the model and the ground truth label.

\subsection{Evaluation}

We split the dataset into training, validation, and test datasets.
For each hyperparameter, we trained a model using the training dataset and evaluated it using the validation dataset. 
We defined the performance of a set of hyperparameters as the accuracy on the validation dataset at the iteration that maximizes the validation accuracy.
If a model has a fine-tuning phase, we used the accuracy after the fine-tuning as the performance.
We chose the set of hyperparameters that maximizes the performance using a hyperparameter optimization algorithm.
We employed Tree-structured Parzen Estimator~\cite{NIPS2011_4443} and for hyperparameter optimization and the median stopping rule implemented in Optuna for pruning unpromising sets of hyperparameters.
Table~\ref{tbl:hyperparameters} shows the set of hyperparameters.
We define the final performance of the model as the accuracy on the test dataset attained by the optimized set of hyperparameters.

For each pair of the dataset and the model, we ran the above evaluation ten times and computed the mean and standard deviation of the performance.

\input{table/hyperparameters}

\subsection{Implementation and Computational Resources}

Experimental code is written in Python3.
We used PyTorch~\cite{NEURIPS2019_9015} and Ignite for the implementation and training of models, Optuna~\cite{optuna} for the hyperparameter optimization, NetworkX~\cite{SciPyProceedings_11} for preprocessing graph objects, and SciPy~\cite{2020SciPy-NMeth} for miscellaneous machine learning operations.
We ran each experiment on a docker image (OS: Ubuntu18.04) built on a cluster.
The image has two CPUs
and single GPGPUs (NVIDIA Tesla V100).

%% file: table/dataset_specification.tex
\begin{table}
  \caption{Dataset specifications.}
  \label{tbl:dataset_specifications}
  \centering
  \begin{tabular}{lllll}
    \toprule
             & \#Node & \#Edge & \#Class ($K$) & Chance Rate \\
    \midrule
    Cora     & 2708  & 5429  & 6      & 30.2\%\\
    CiteSeer & 3312  & 4732  & 7      & 21.1\%\\
    PubMed   & 19717 & 44338 & 3      & 39.9\%\\
    \bottomrule
  \end{tabular}
\end{table}

%% file: table/hyperparameters.tex
\begin{table}
  \caption{Hyperparameters of experiments in Section \ref{sec:experiment}. $X \sim \mathrm{LogUnif}[a, b]$ means the random variable $\log_{10} X$ obeys the uniform distribution over $[a, b]$. $(\ast)$ For KTA + Fine Tuning setting, we reduce the number of weak learners to 40 due to GPU memory constraints. $(\ast\ast)$ Learning rate corresponds to $\alpha$ when Optimization algorithm is Adam~\cite{kingma2014iclr}.
  }
  \label{tbl:hyperparameters}
  \centering
  \begin{tabular}{lll}
    \toprule
    Category & Name     & Value \\
    \midrule
    Boosting
     & \#Weak learners & $\{1, 2, \ldots, 100\ (40^{(\ast)}) \}$ \\
     & Minibatch size $B$ & $\{1, 2, \ldots, |\Vtrain|\}$ \\
     & Clipping value & $\mathrm{LogUnif}[-10, -5]$ \\
    Model
     & Epoch & $\{10, 20, \ldots, 100\}$ \\
     & Optimization algorithm & $\{\mathrm{SGD}, \mathrm{Adam}, \mathrm{RMSProp}\}$ \\
     & Learning rate$^{(\ast\ast)}$  & $\mathrm{LogUnif}[-5, -1]$ \\
     & Momentum  & $\mathrm{LogUnif}[-10, -1]$ \\
     & Weight decay  & $\mathrm{LogUnif}[-10, -1]$ \\    
     & Unit size & $\{10, 11,\ldots, 200\}$ \\
     & Dropout & $\{\mathrm{ON} (\text{ratio=0.5}), \mathrm{OFF}\}$ \\
    Input Injection
     & Mixing ratio $\rho$ & $\mathrm{Unif}[0, 1]$ \\
    Kernel Target Alignment
     & Epoch & $\{5, 6, \ldots, 30\}$ \\
     & Optimization algorithm & $\{\mathrm{SGD}, \mathrm{Adam}, \mathrm{RMSProp}\}$ \\ 
     & Learning rate$^{(\ast\ast)}$ & $\mathrm{LogUnif}[-5, -1]$ \\
     & Degree $N_{\mathrm{deg}}$ & $3$ \\
    Fine Tuning
     & Epoch & $\{1, 2, \ldots, 100\}$ \\
     & Optimization algorithm & $\{\mathrm{SGD}, \mathrm{Adam}, \mathrm{RMSProp}\}$ \\
     & Learning rate$^{(\ast\ast)}$  & $\mathrm{LogUnif}[-5, -1]$ \\
     & Momentum  & $\mathrm{LogUnif}[-10, -1]$ \\
     & Weight decay  & $\mathrm{LogUnif}[-10, -1]$ \\
    \bottomrule
  \end{tabular}
\end{table}

%% file: source/appendix/experiment_additional_results.tex
\section{Additional Experiment Results}

\input{source/appendix/experiment_additional_results/experiment_architecture_variants_results}
\input{source/appendix/experiment_additional_results/model_comparison}

%% file: source/appendix/experiment_additional_results/experiment_architecture_variants_results.tex
\subsection{More Results for Model Variants}\label{sec:experiment-additional-results}

Table~\ref{tbl:node_classification_on_citation_network} shows the result of the prediction accuracies of models that use MLPs with various layer size $L$ as transformation functions $\mathcal{B}^{(t)}$.
It also shows the results for the input injection model (GB-GNN-II) we have introduced in Section~\ref{sec:more-model-variants} and the SAMME.R model (GB-GNN-SAMME.R) in Section~\ref{sec:model-details}.
Figures~\ref{fig:train-loss-all-cora}--\ref{fig:train-loss-all-pubmed} show the transition of the training loss for $L=0, \ldots, 4$ (Figure~\ref{fig:train-loss-all-cora}: Cora, Figure \ref{fig:train-loss-all-citeseer}: CiteSeer, Figure~\ref{fig:train-loss-all-pubmed}: PubMed).
Figures~\ref{fig:test-loss-all-cora}--\ref{fig:test-loss-all-pubmed} show the transition of the training loss for $L=0, \ldots, 4$ (Figure \ref{fig:test-loss-all-cora}: Cora, Figure \ref{fig:test-loss-all-citeseer}: CiteSeer, Figure~\ref{fig:test-loss-all-pubmed}: PubMed).
Figures~\ref{fig:cosine-all-cora}--\ref{fig:cosine-all-pubmed} show the transition of the cosine values between the negative gradient $-\nabla \widehat{\mathcal{L}}(\widehat{Y}^{(t-1)})$ and the weak learner $f^{(t)}$ at the $t$-th iteration for models that has $L=0$ to $4$ layers (Figure~\ref{fig:cosine-all-cora}: Cora, Figure~\ref{fig:cosine-all-citeseer}: CiteSeer, Figure~\ref{fig:cosine-all-pubmed}: PubMed).

\input{image/train_loss_all}
\input{image/test_loss_all}
\input{image/cosine_all}
\input{table/citation_for_deep_mlps}

%% file: image/train_loss_all.tex
\begin{figure}[p]
  \includegraphics[width=0.33\linewidth]{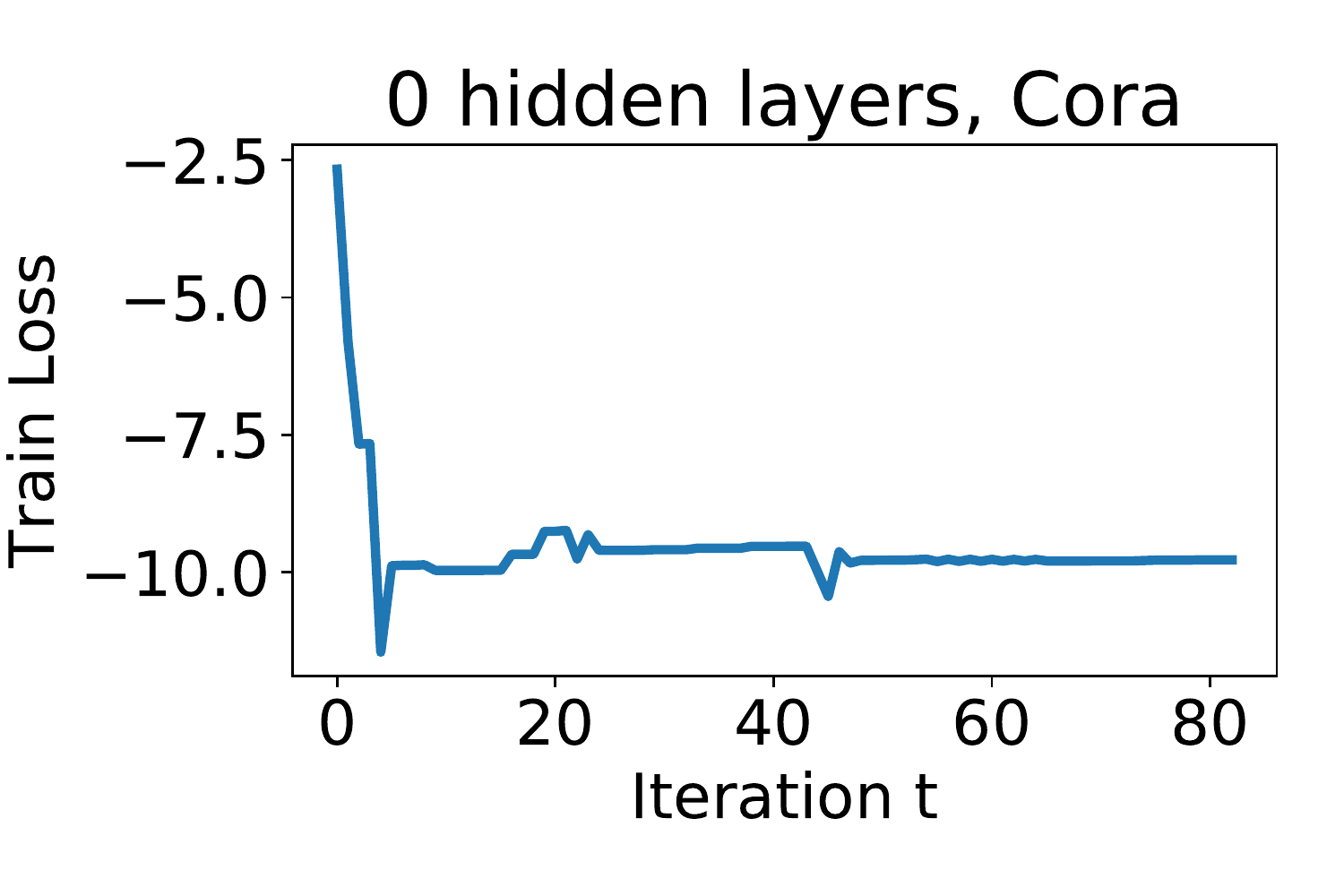} 
  \includegraphics[width=0.33\linewidth]{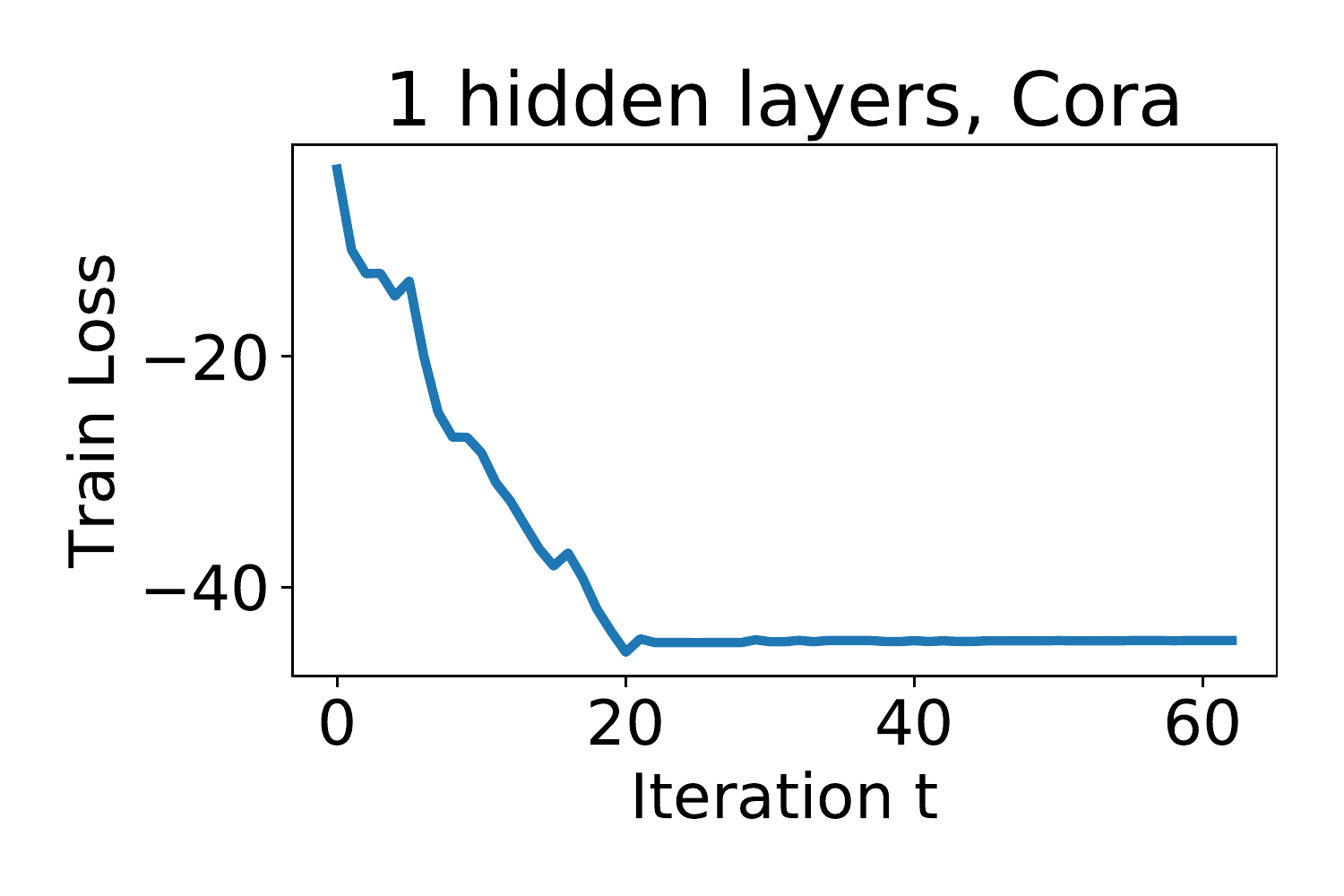}
  \includegraphics[width=0.33\linewidth]{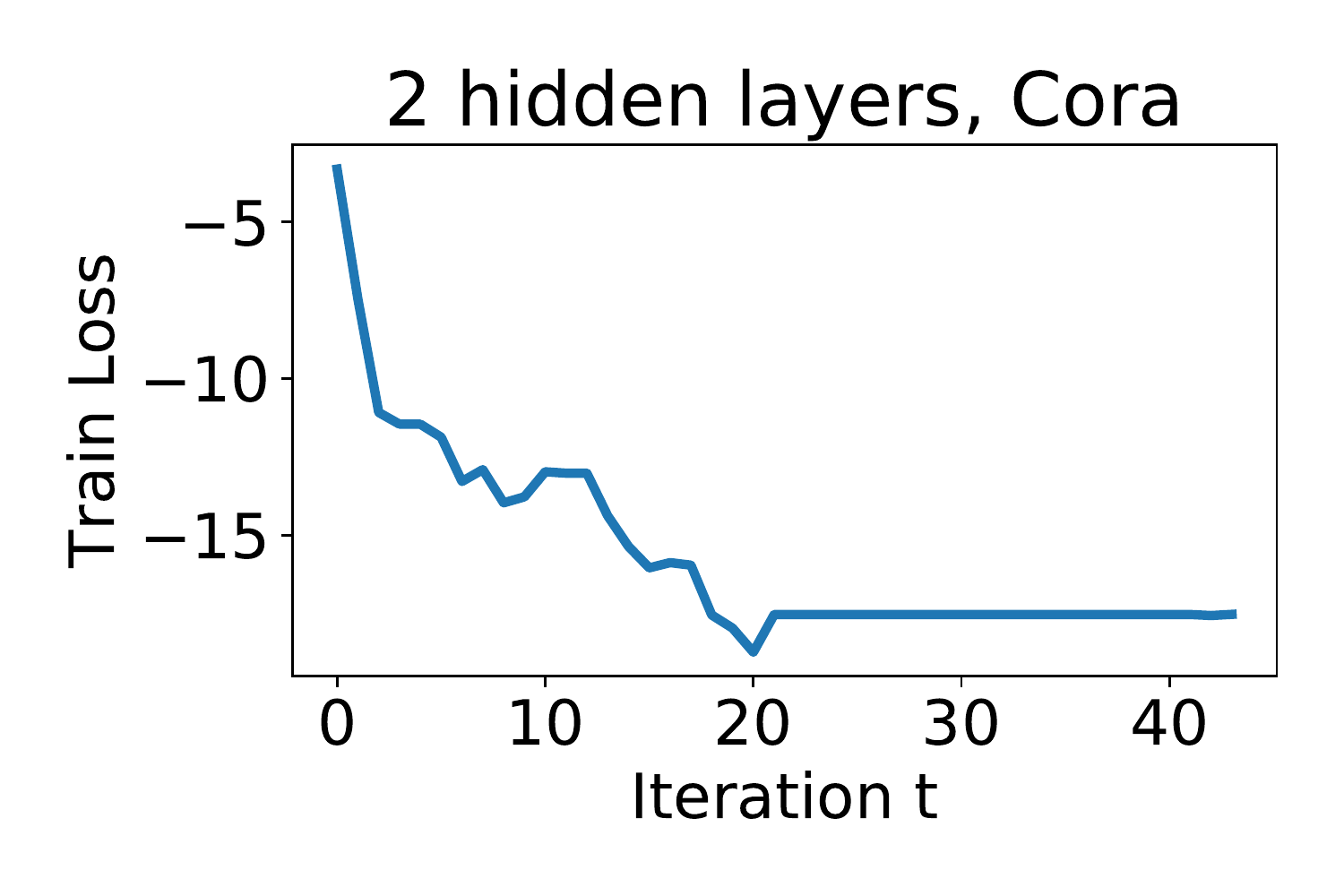}\\
  \includegraphics[width=0.33\linewidth]{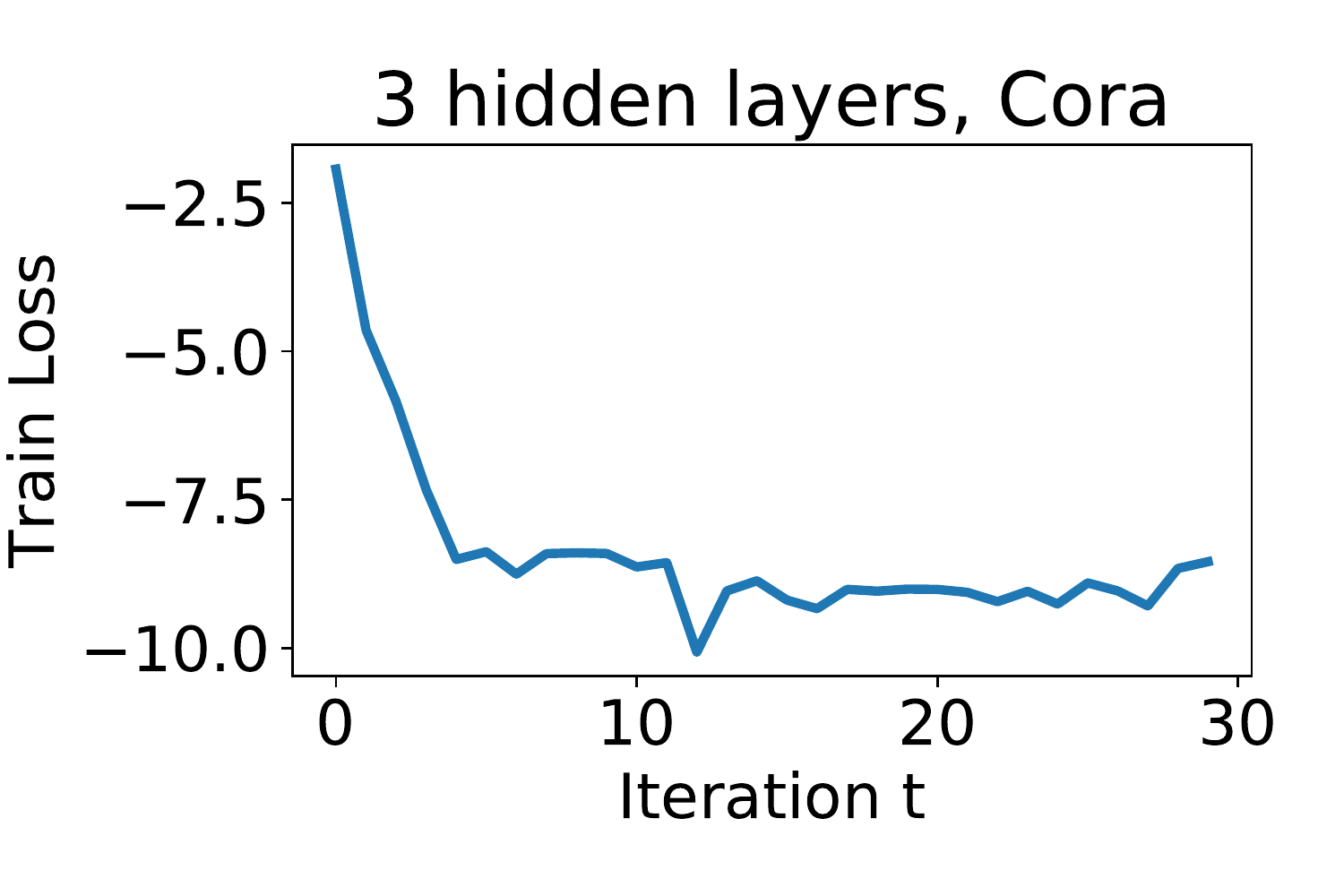}
  \includegraphics[width=0.33\linewidth]{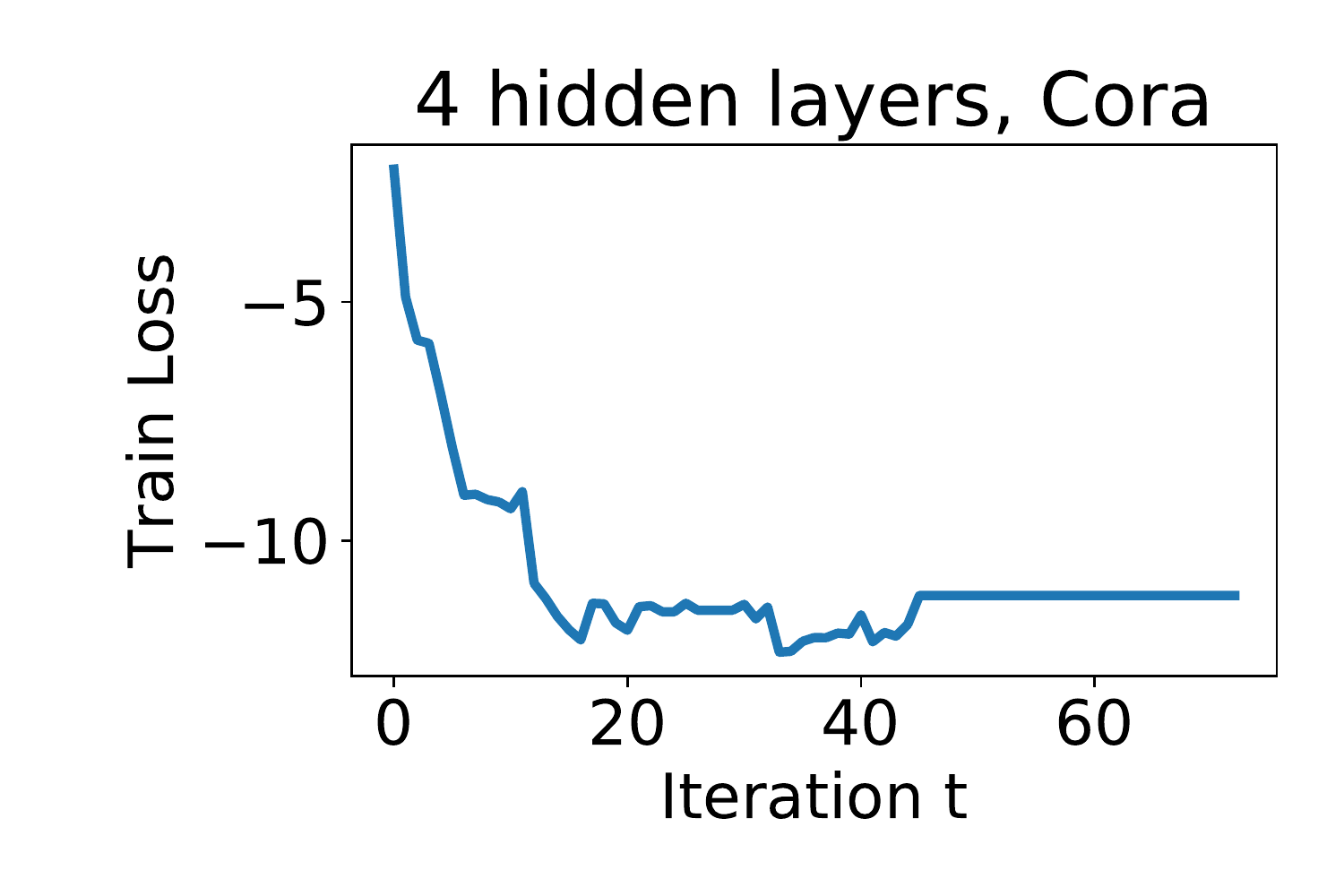}
  \caption{Train loss transition for the Cora dataset.}\label{fig:train-loss-all-cora}
\vspace{\baselineskip}
  \includegraphics[width=0.33\linewidth]{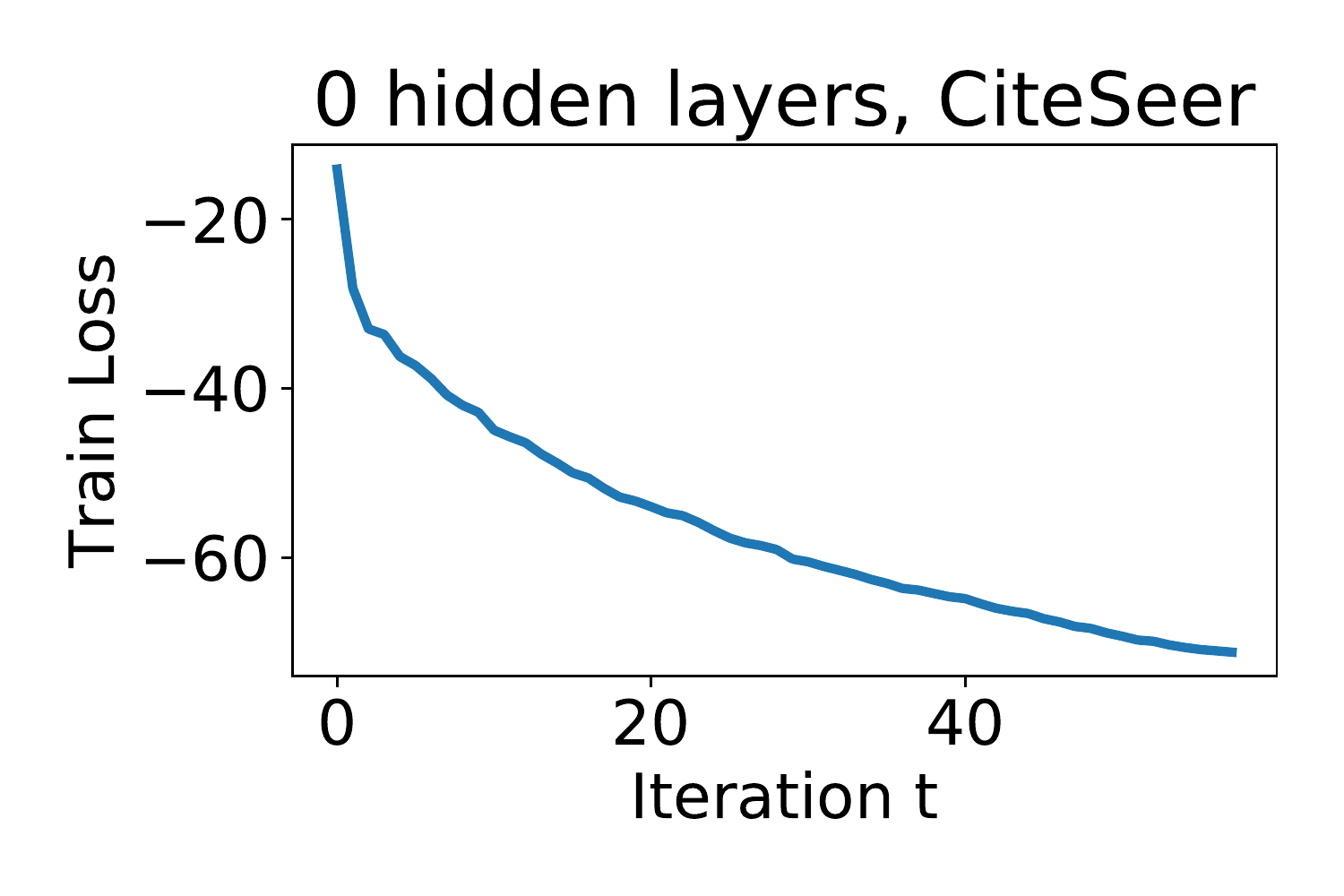} 
  \includegraphics[width=0.33\linewidth]{image/train_loss/citeseer/1.pdf}
  \includegraphics[width=0.33\linewidth]{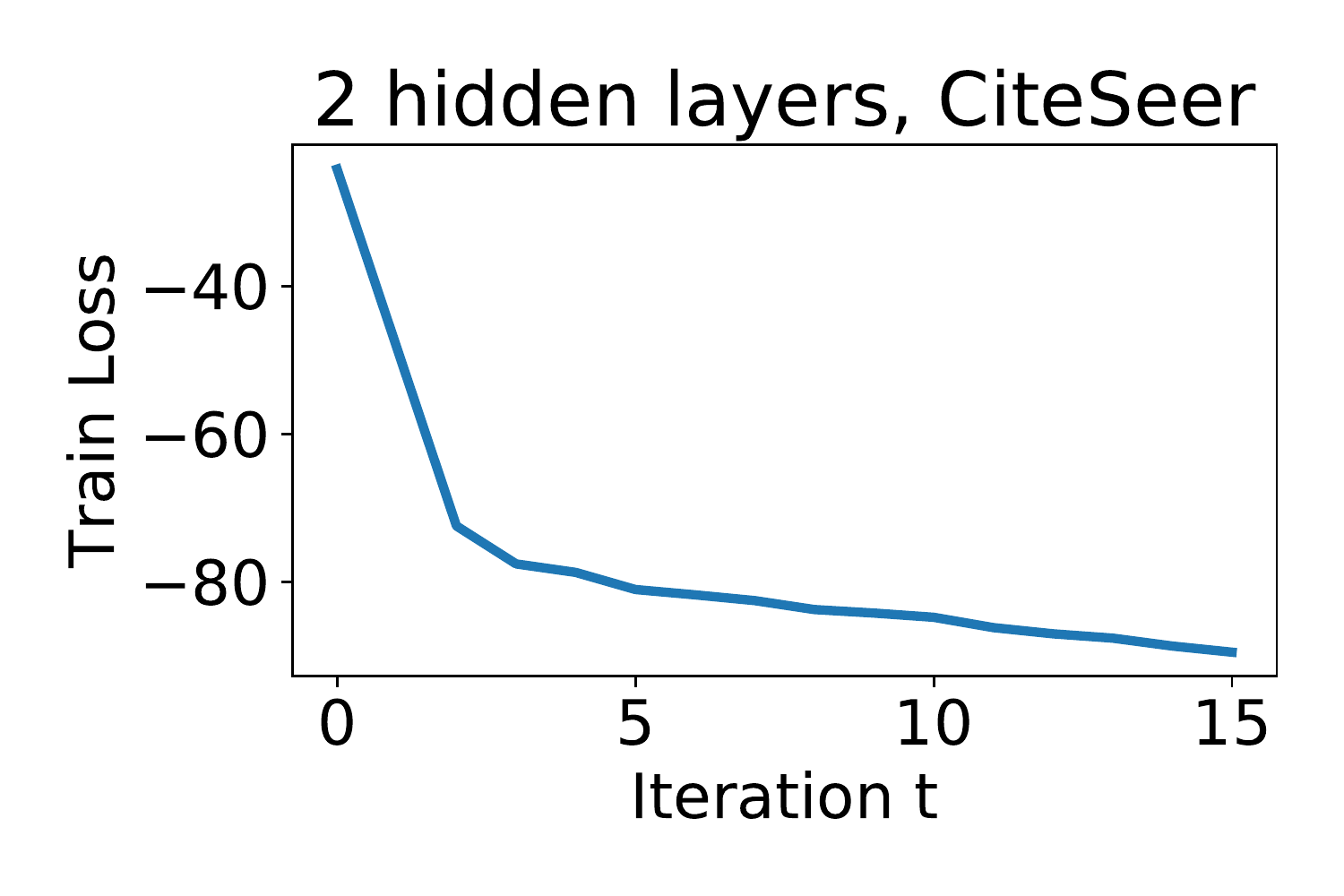}\\
  \includegraphics[width=0.33\linewidth]{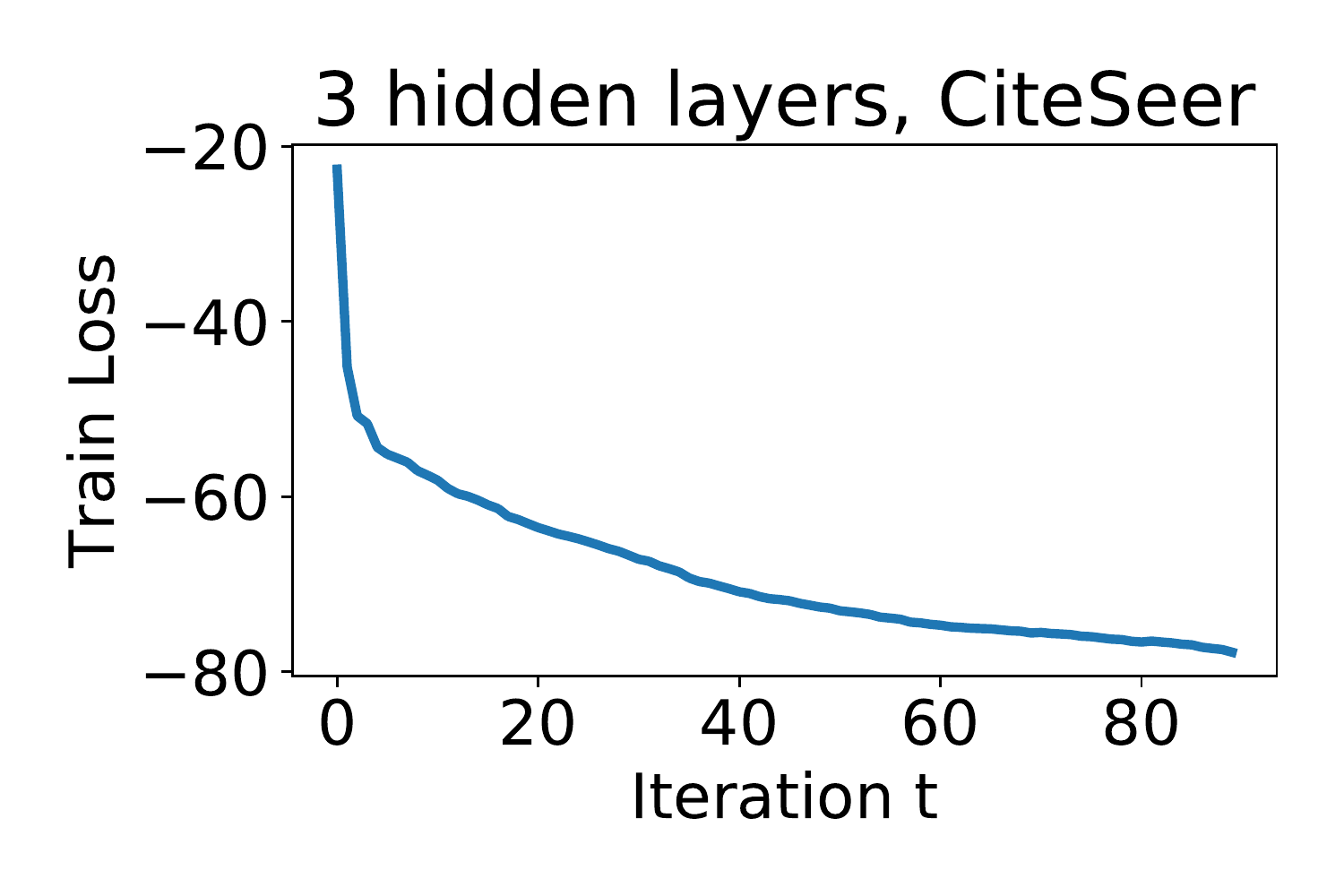}
  \includegraphics[width=0.33\linewidth]{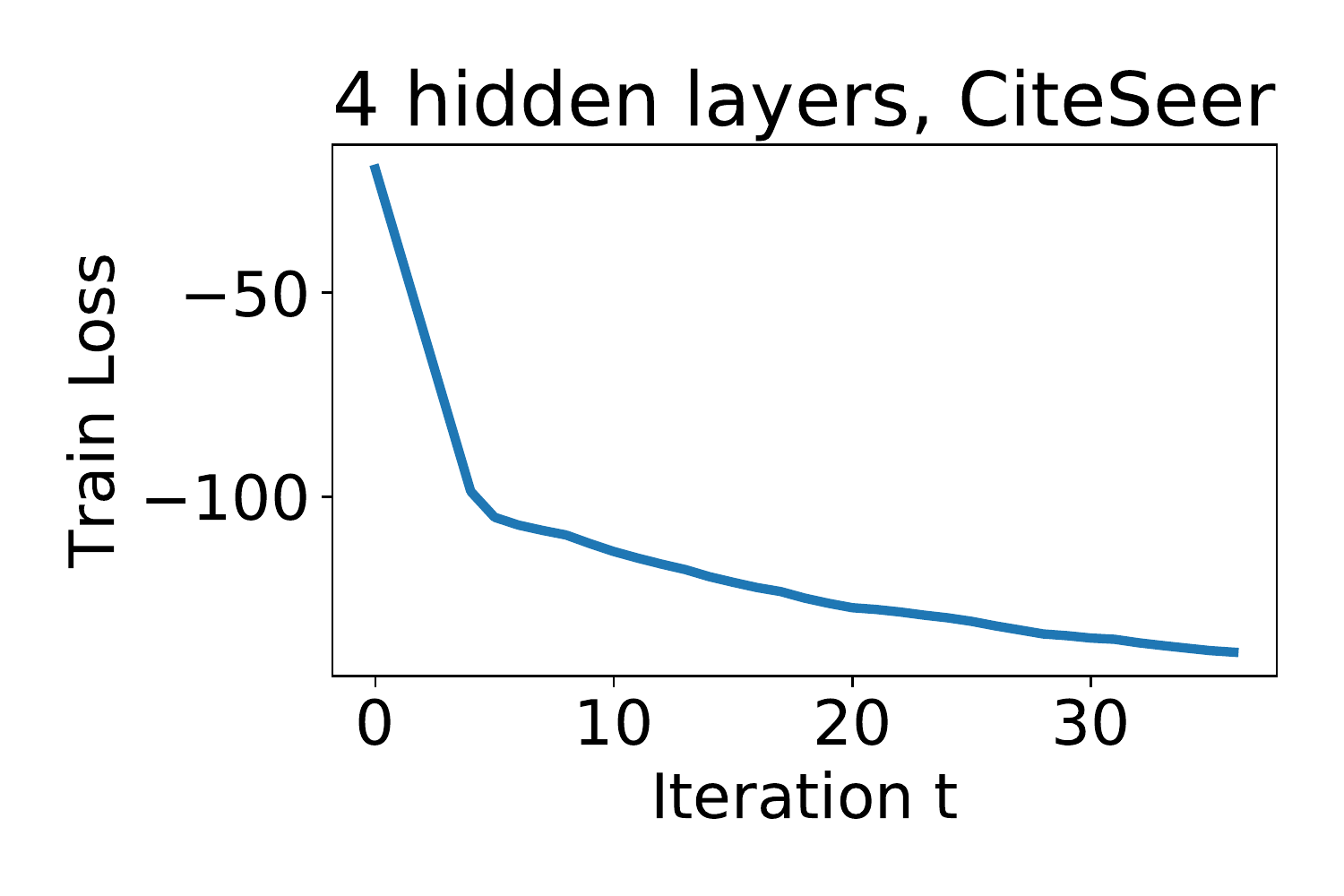}
  \caption{Train loss transition for the CiteSeer dataset.}\label{fig:train-loss-all-citeseer}
\vspace{\baselineskip}
  \includegraphics[width=0.33\linewidth]{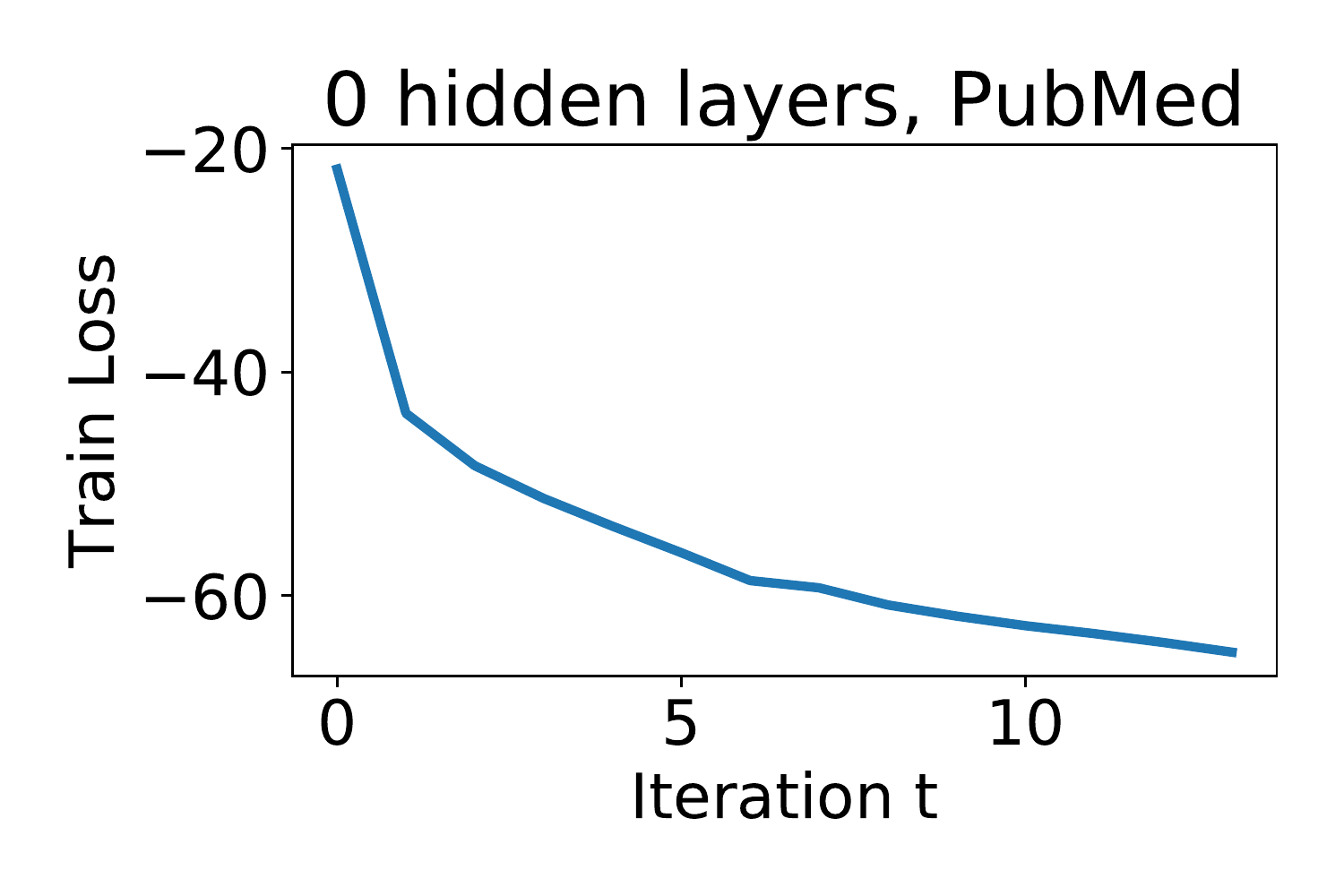} 
  \includegraphics[width=0.33\linewidth]{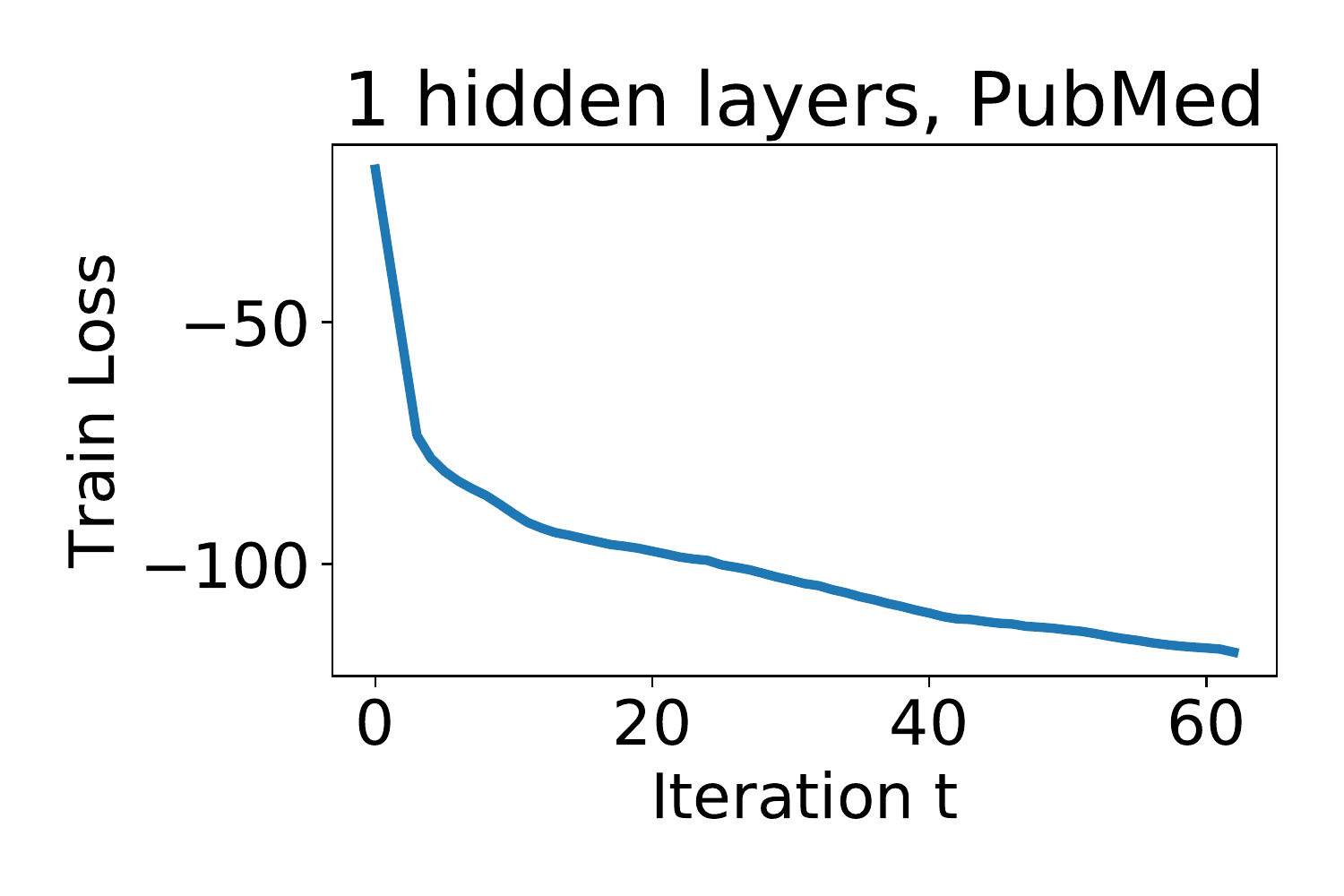}
  \includegraphics[width=0.33\linewidth]{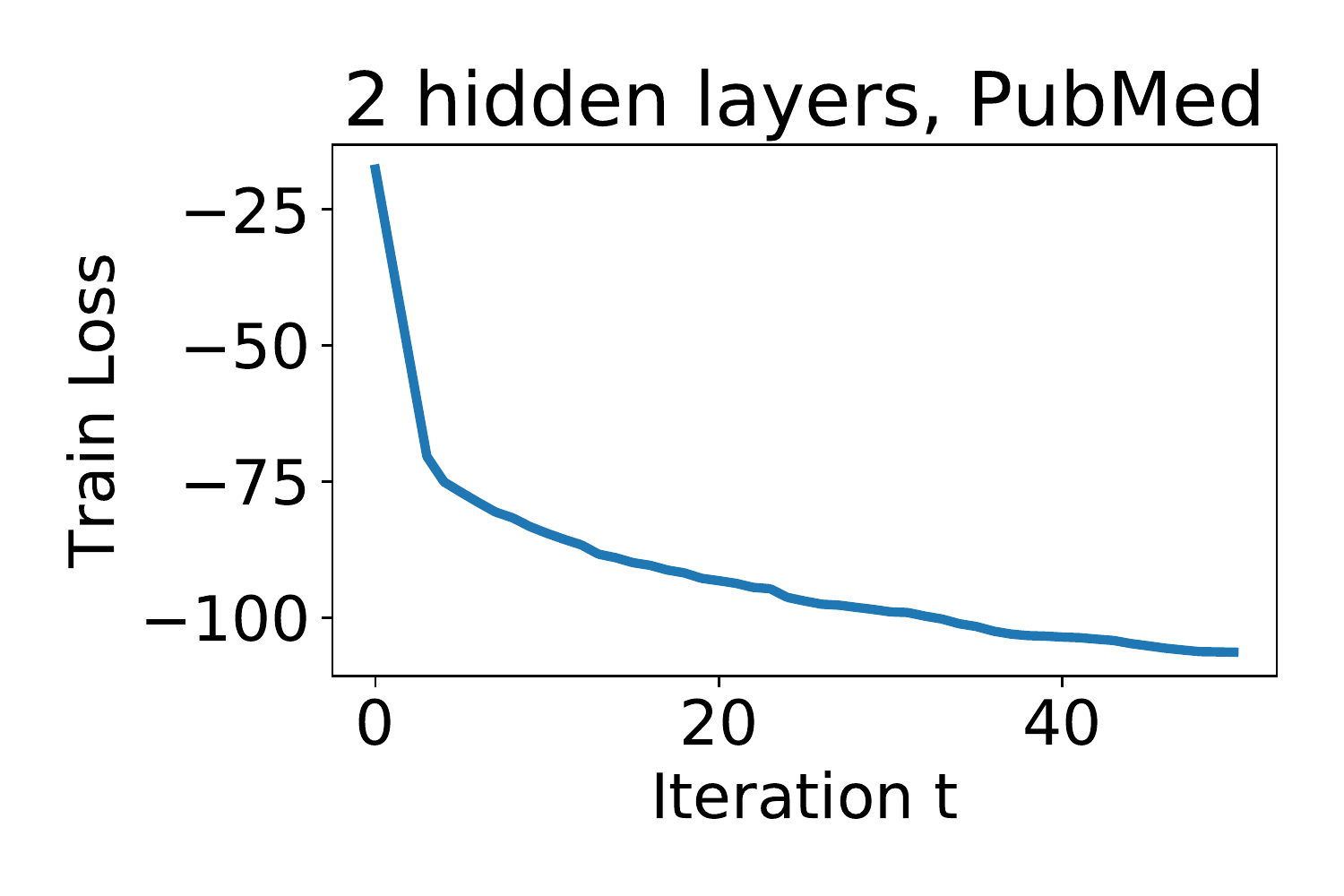}\\
  \includegraphics[width=0.33\linewidth]{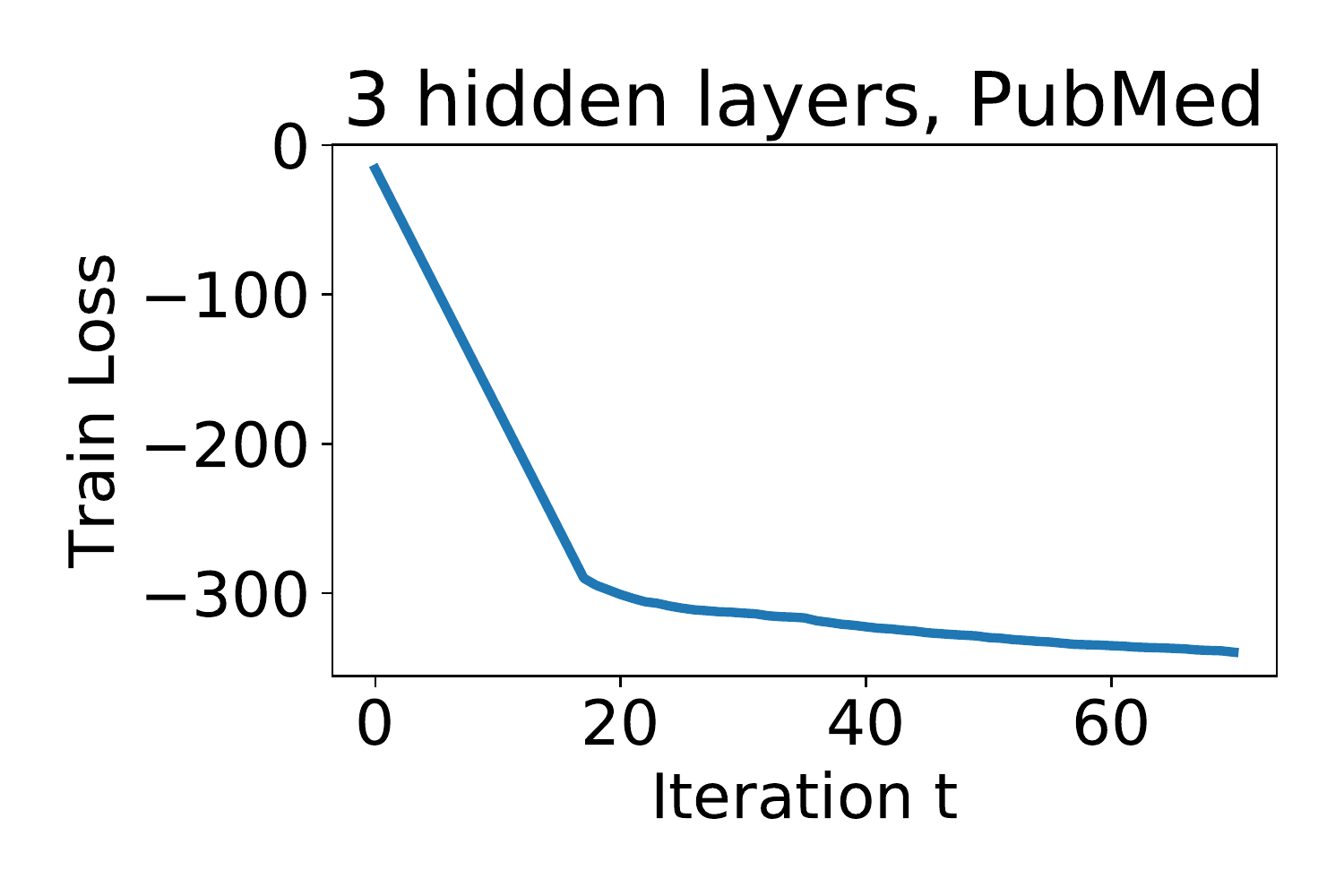}
  \includegraphics[width=0.33\linewidth]{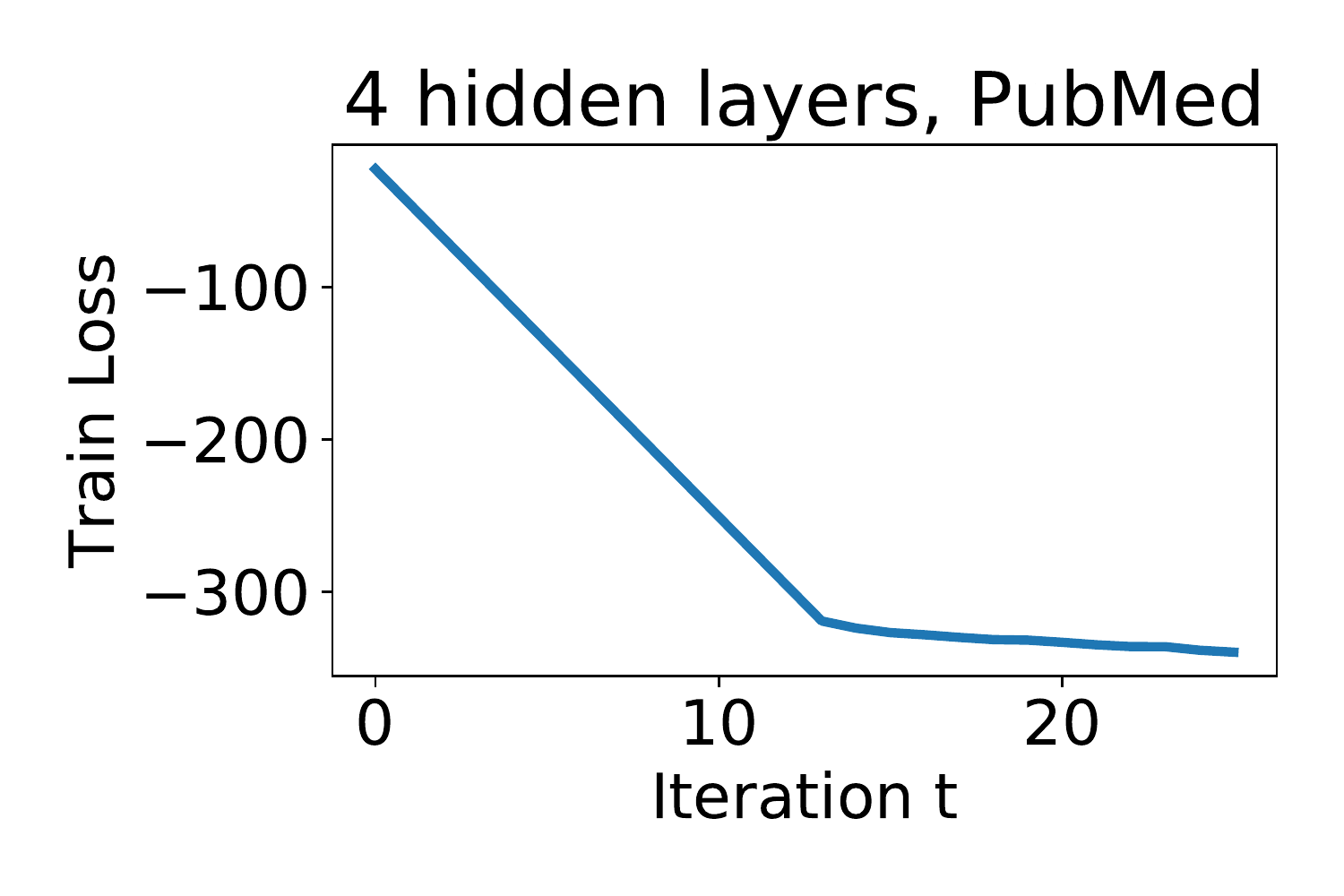}
  \caption{Train loss transition for the PubMed dataset.}\label{fig:train-loss-all-pubmed}
\end{figure}

%% file: image/test_loss_all.tex
\begin{figure}[p]
  \includegraphics[width=0.33\linewidth]{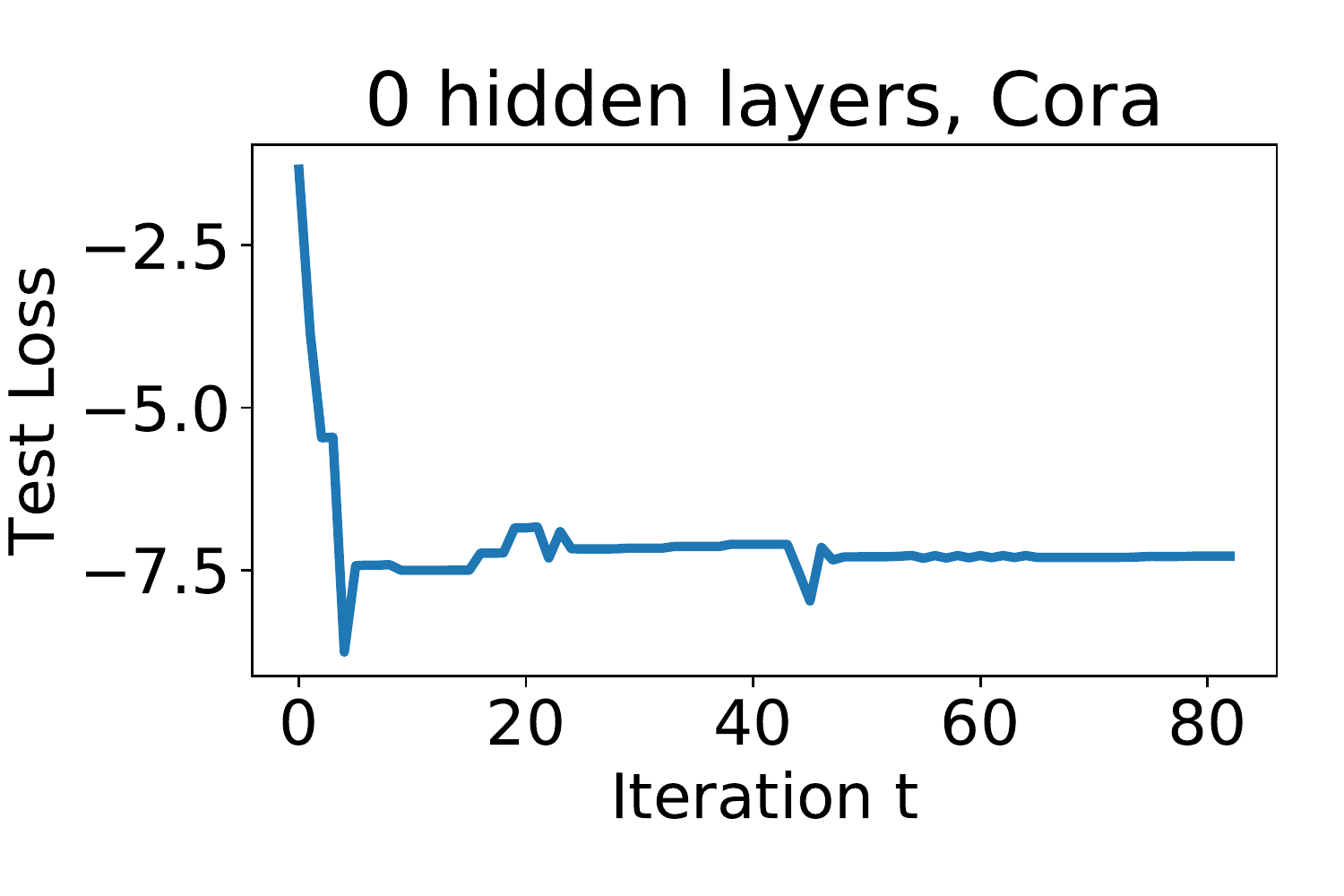} 
  \includegraphics[width=0.33\linewidth]{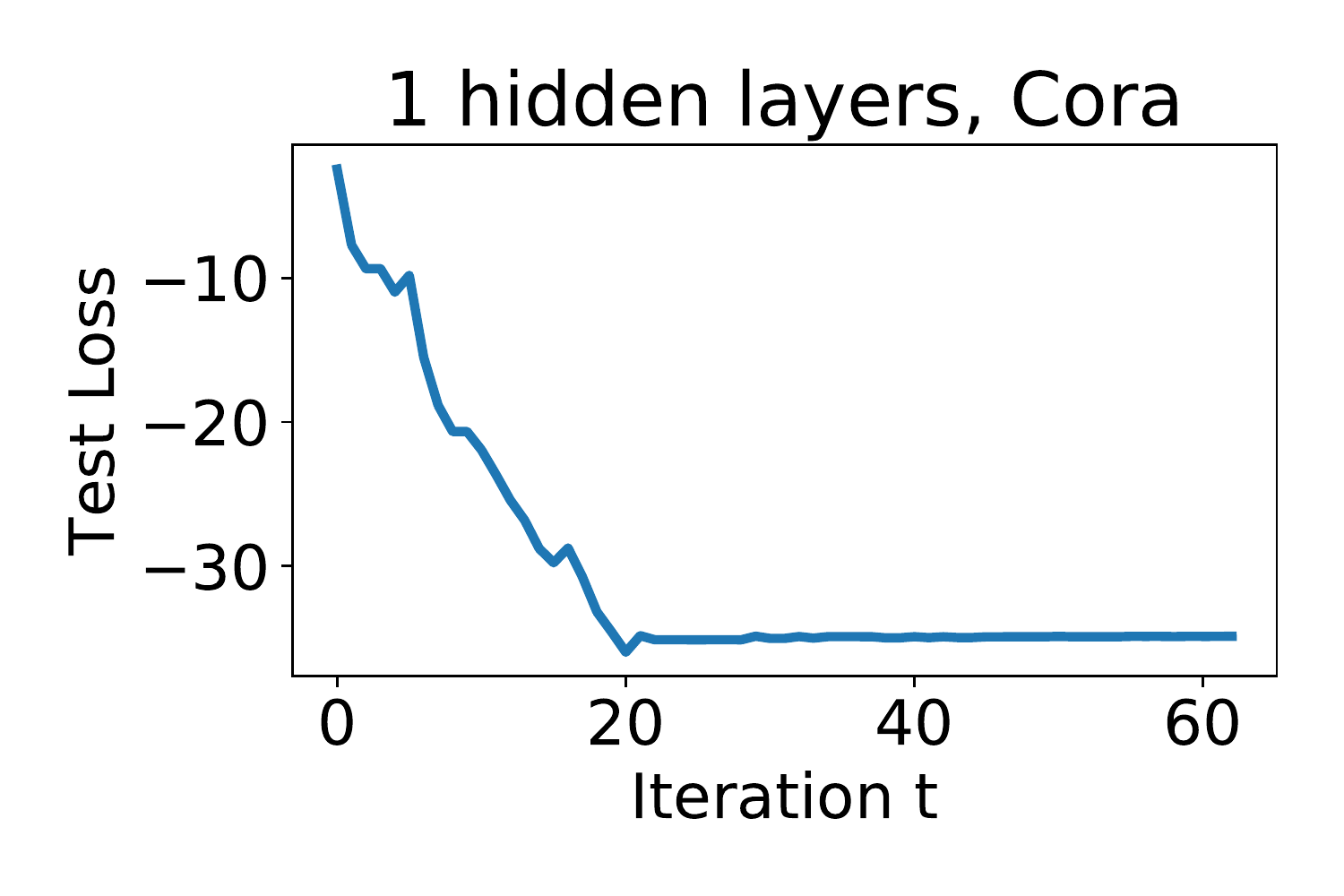}
  \includegraphics[width=0.33\linewidth]{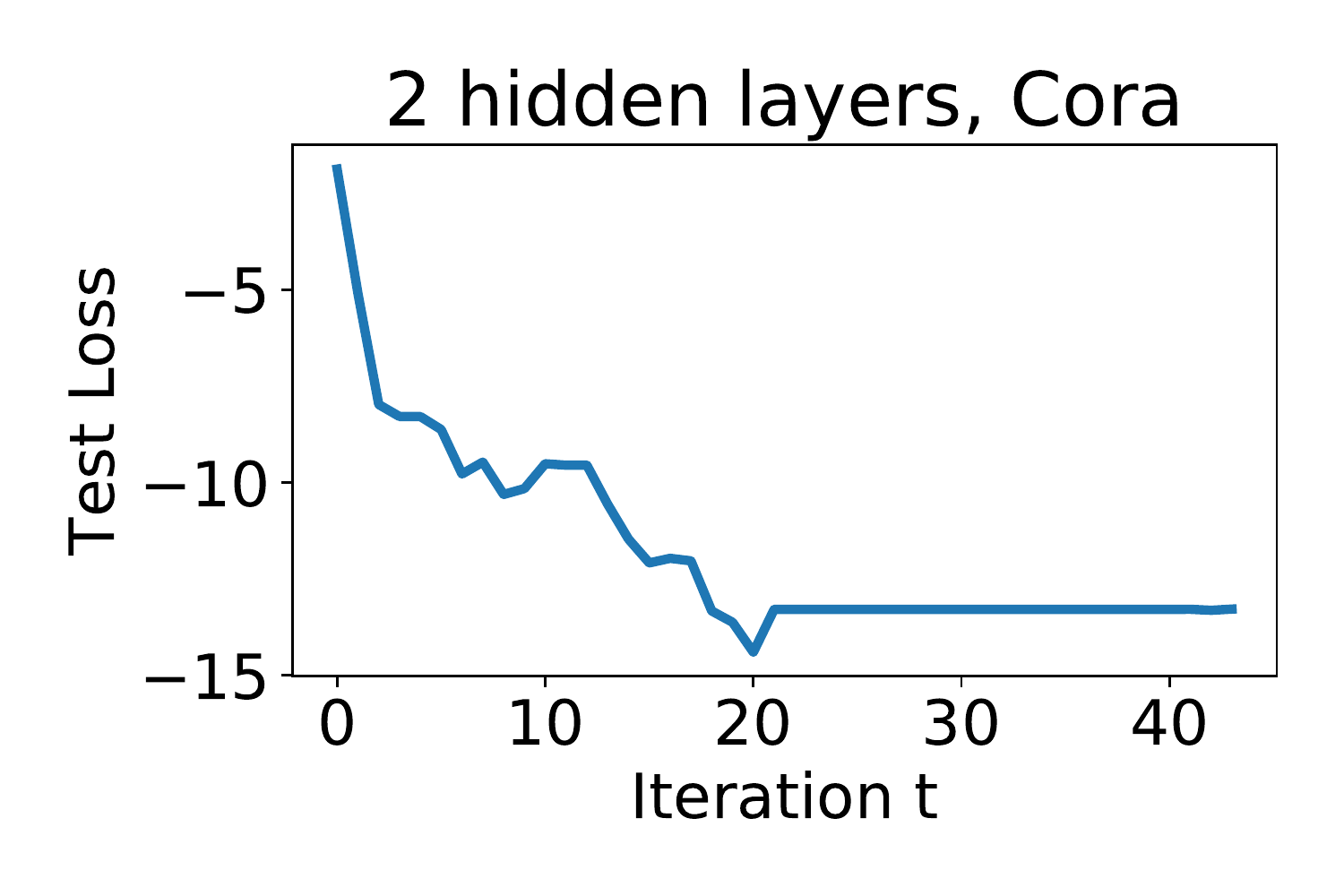}\\
  \includegraphics[width=0.33\linewidth]{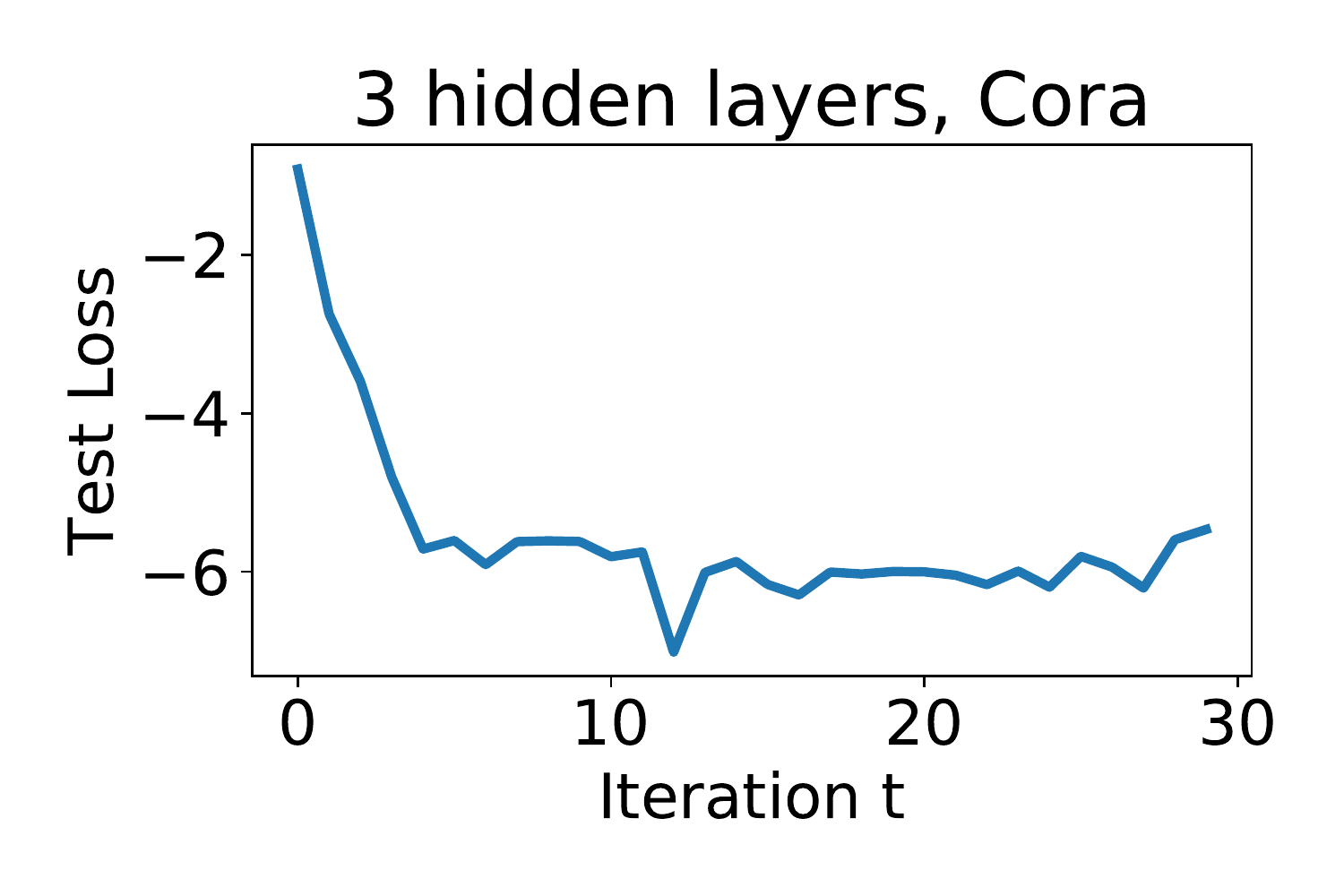}
  \includegraphics[width=0.33\linewidth]{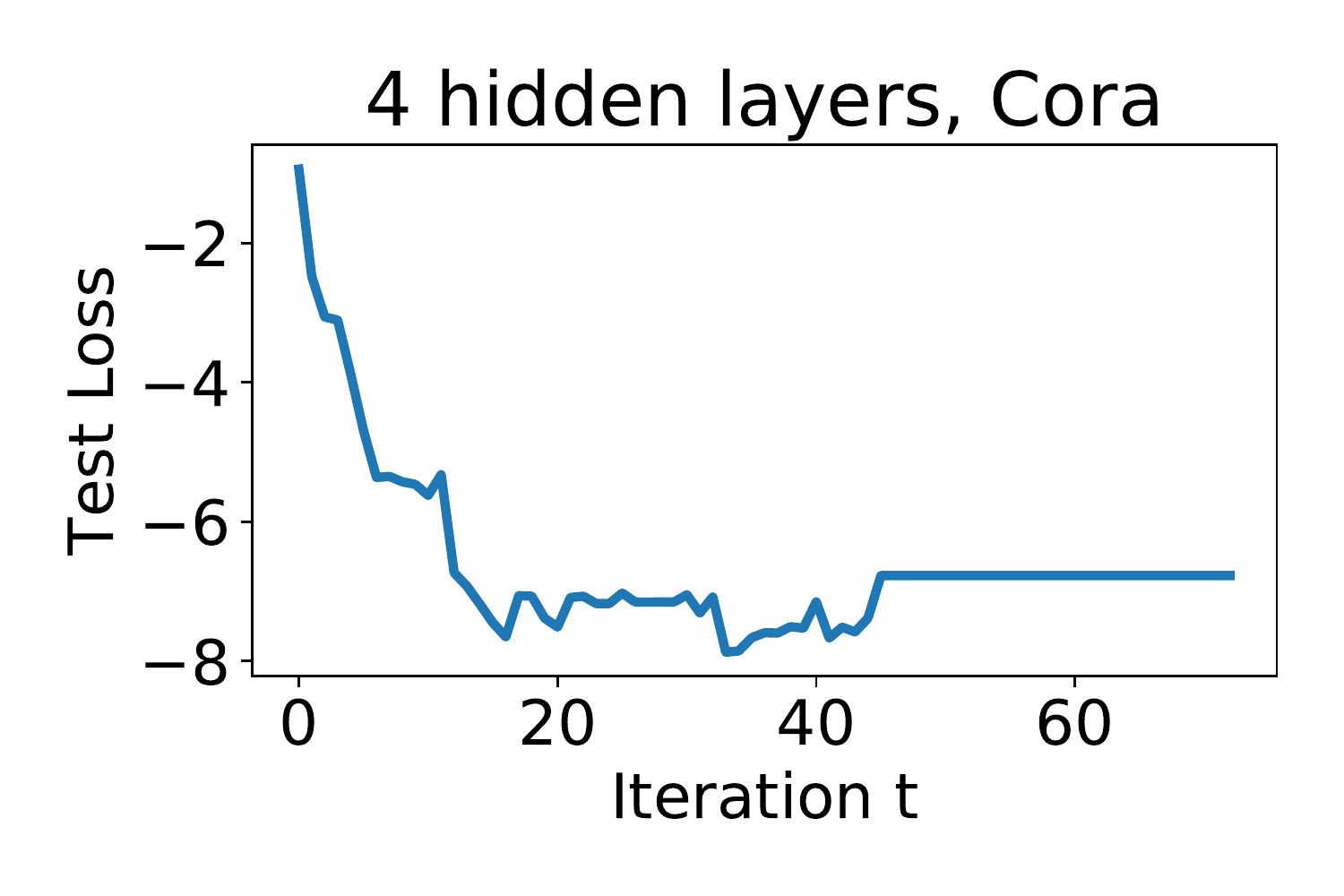}
  \caption{Test loss transition for the Cora dataset.}\label{fig:test-loss-all-cora}
\vspace{\baselineskip}
  \includegraphics[width=0.33\linewidth]{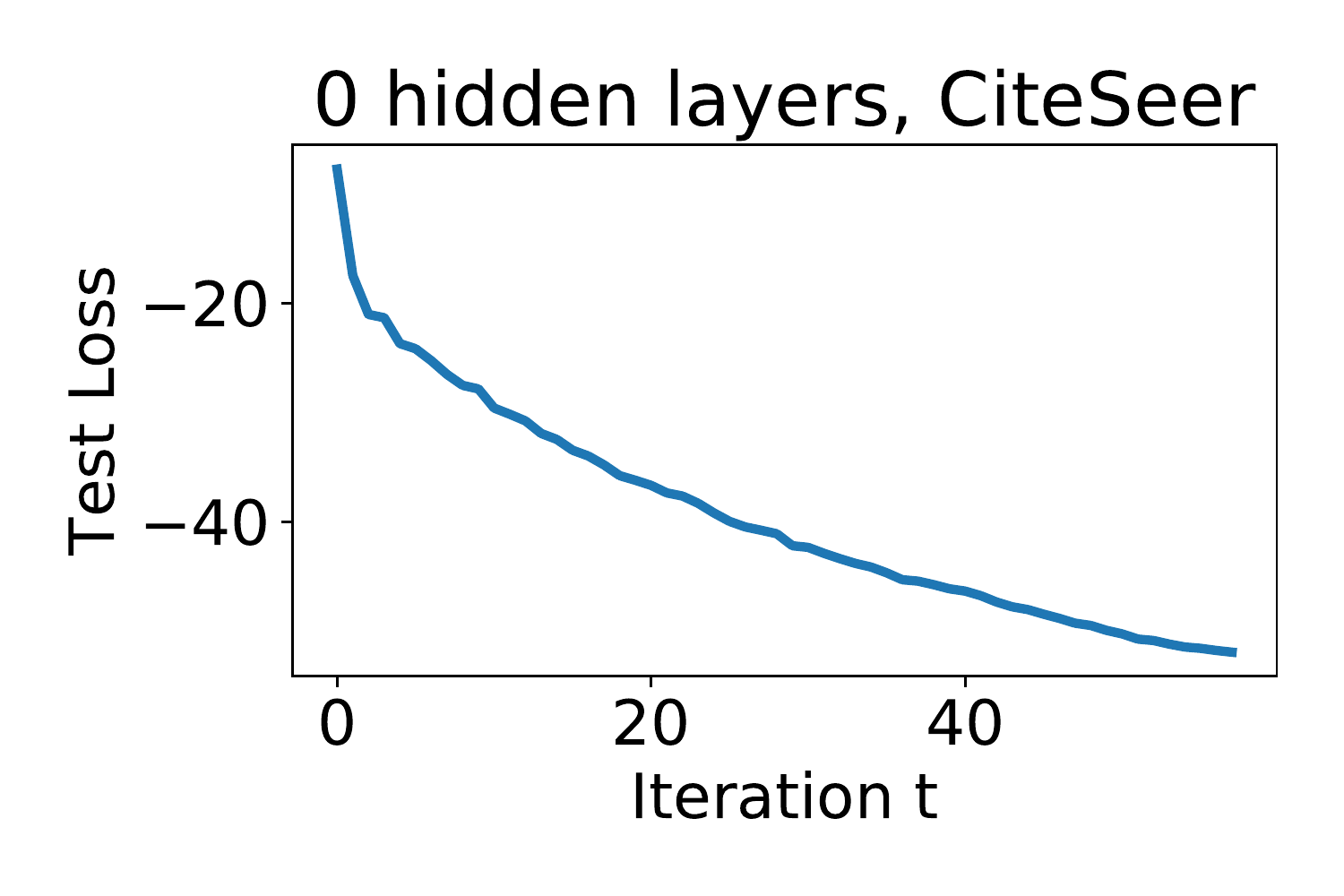} 
  \includegraphics[width=0.33\linewidth]{image/test_loss/citeseer/1.pdf}
  \includegraphics[width=0.33\linewidth]{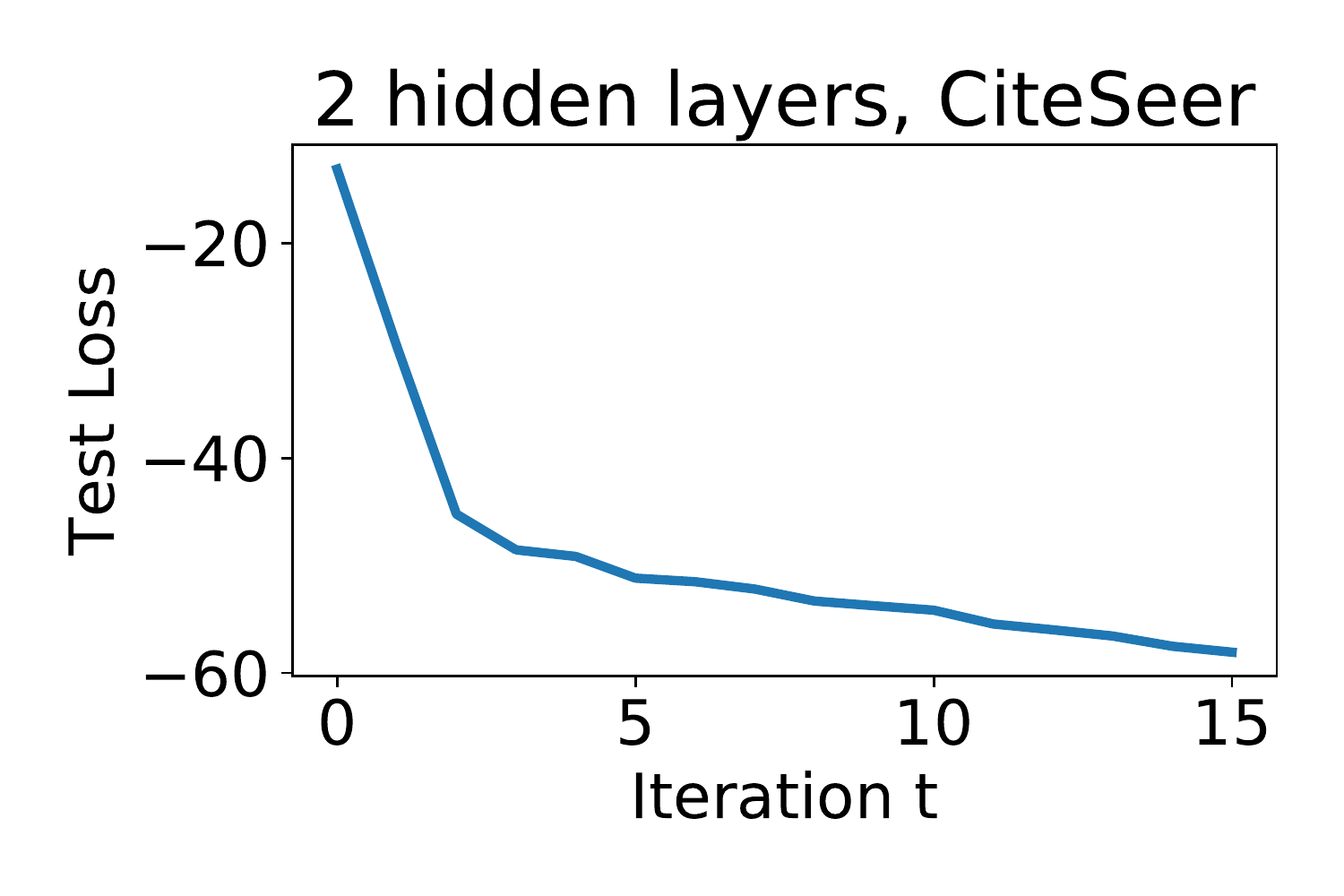}\\
  \includegraphics[width=0.33\linewidth]{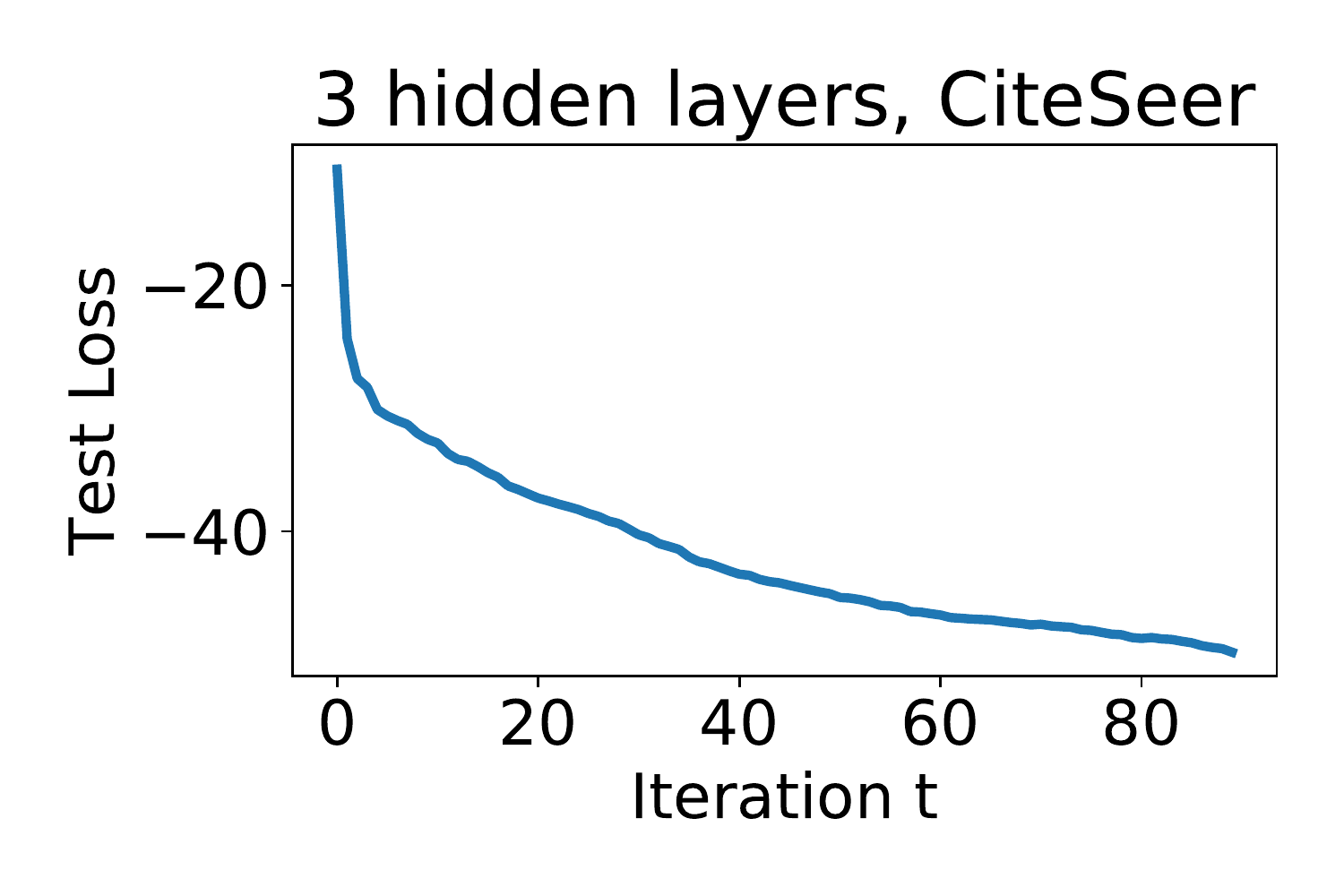}
  \includegraphics[width=0.33\linewidth]{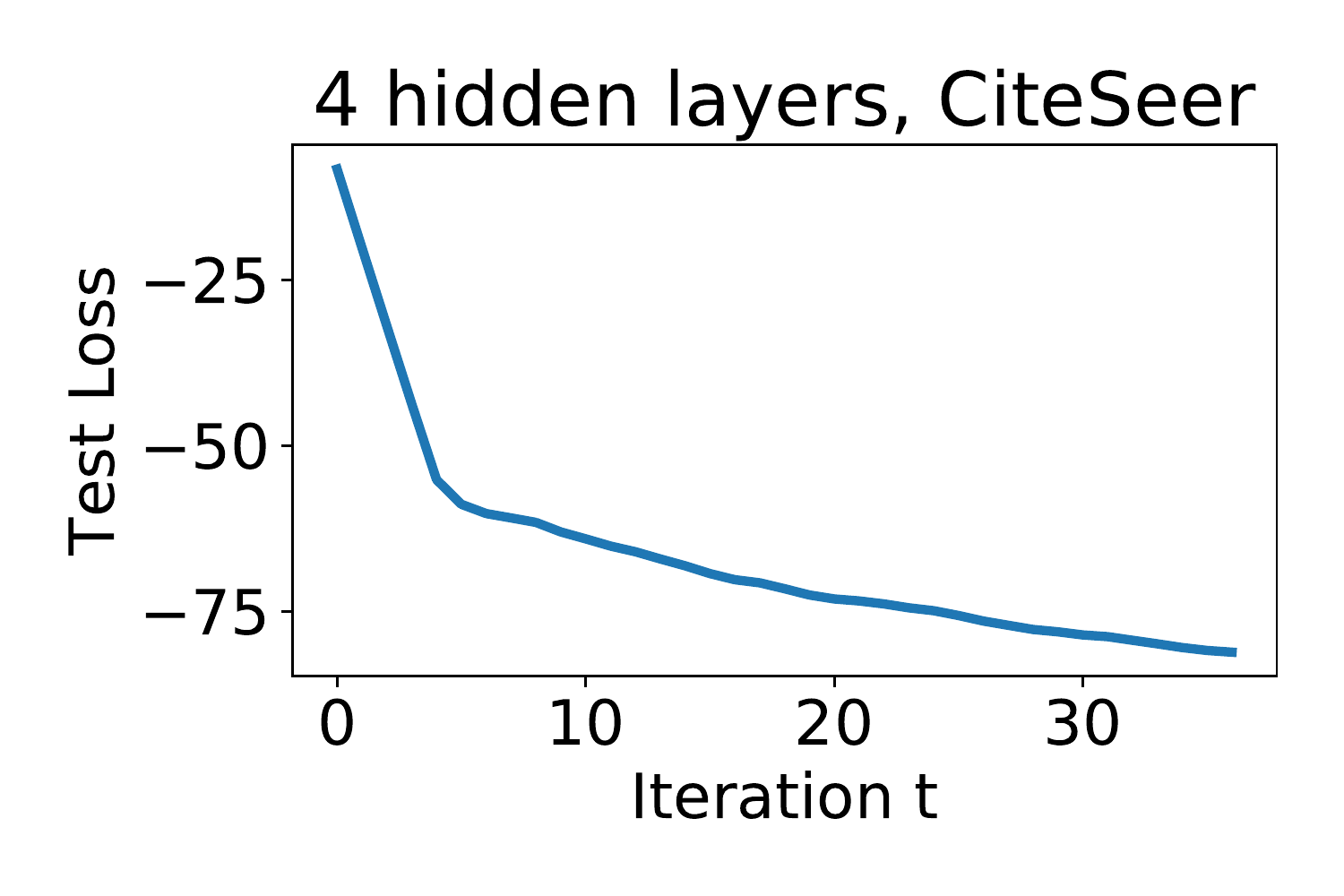}
  \caption{Test loss transition for the CiteSeer dataset.}\label{fig:test-loss-all-citeseer}
\vspace{\baselineskip}
  \includegraphics[width=0.33\linewidth]{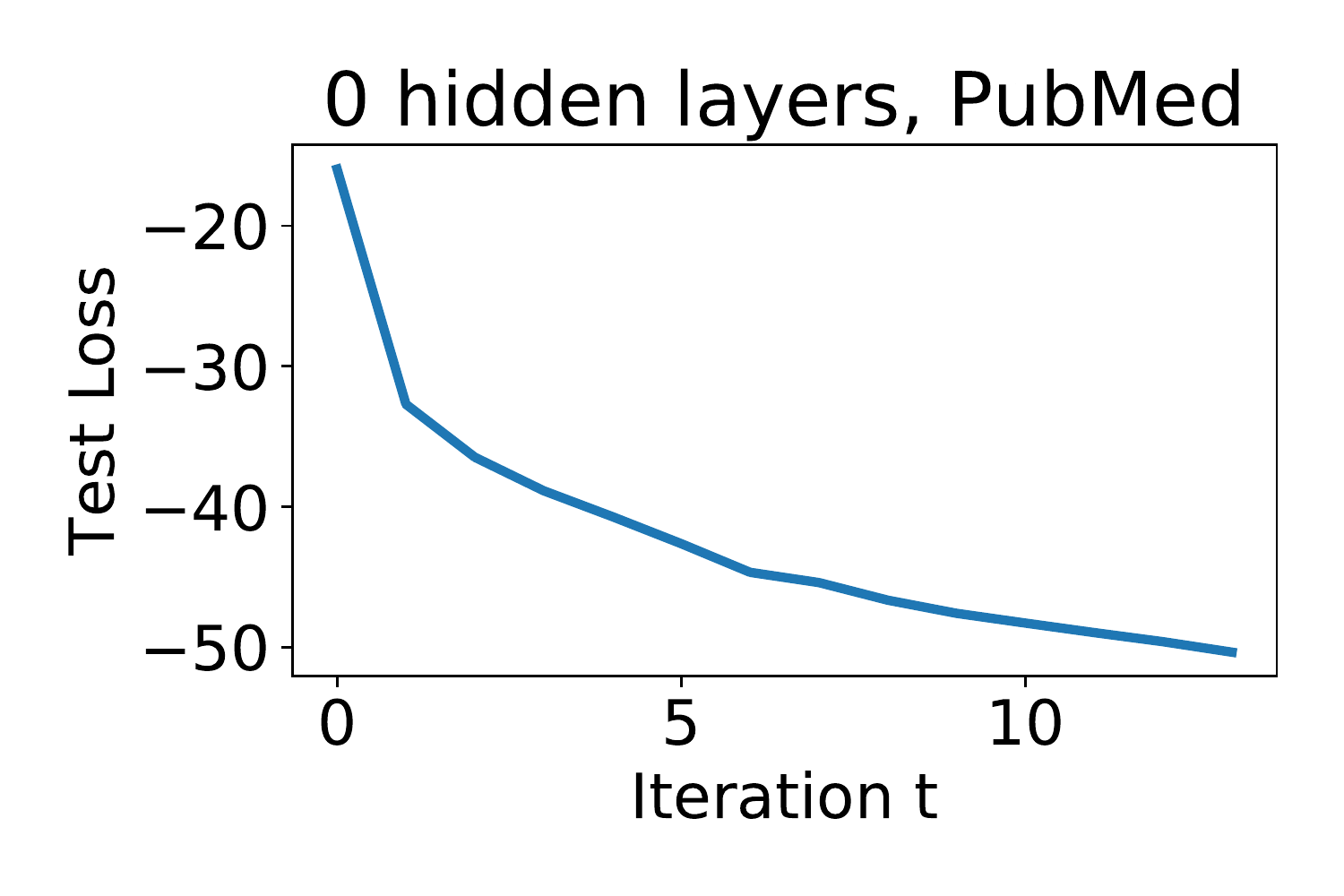} 
  \includegraphics[width=0.33\linewidth]{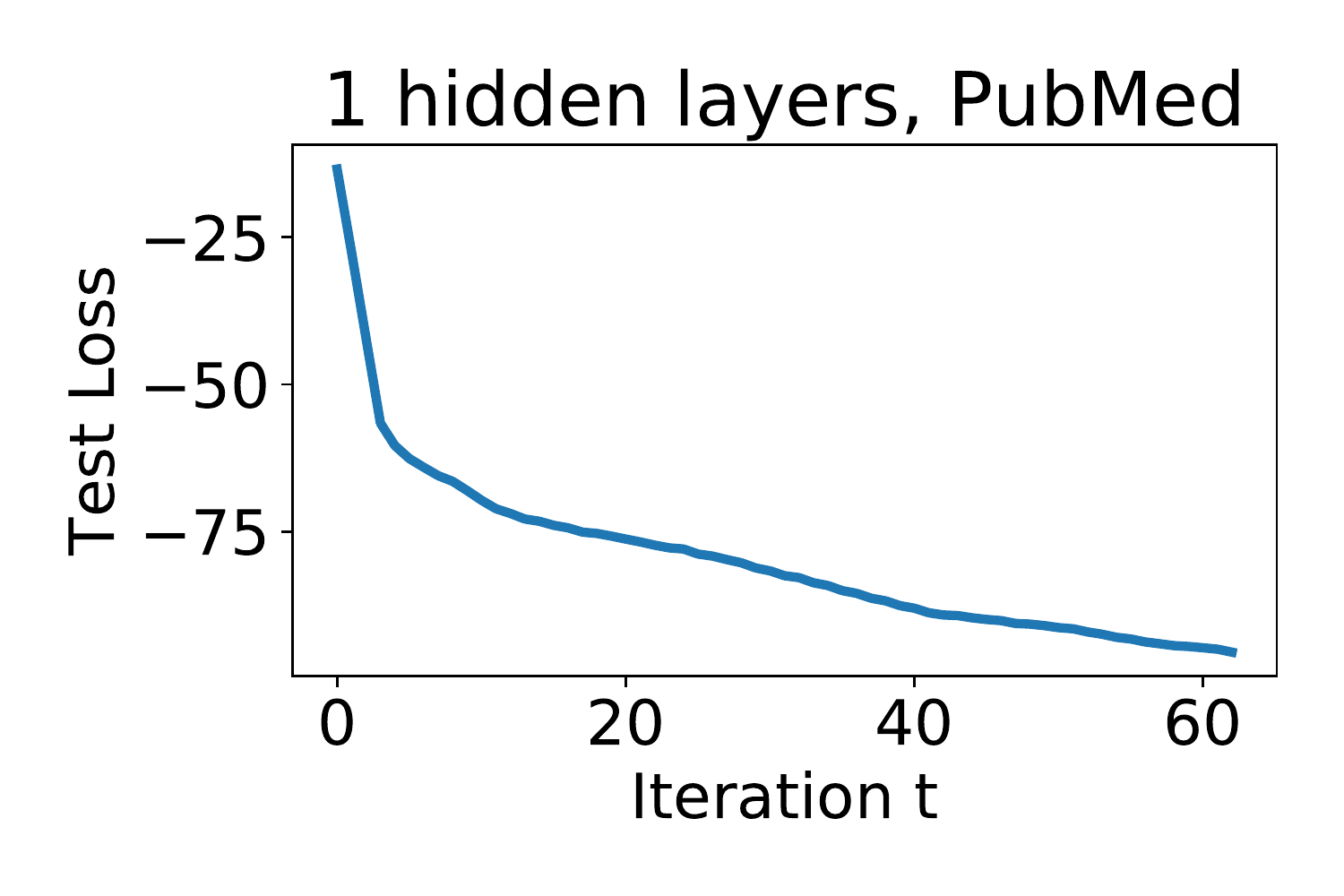}
  \includegraphics[width=0.33\linewidth]{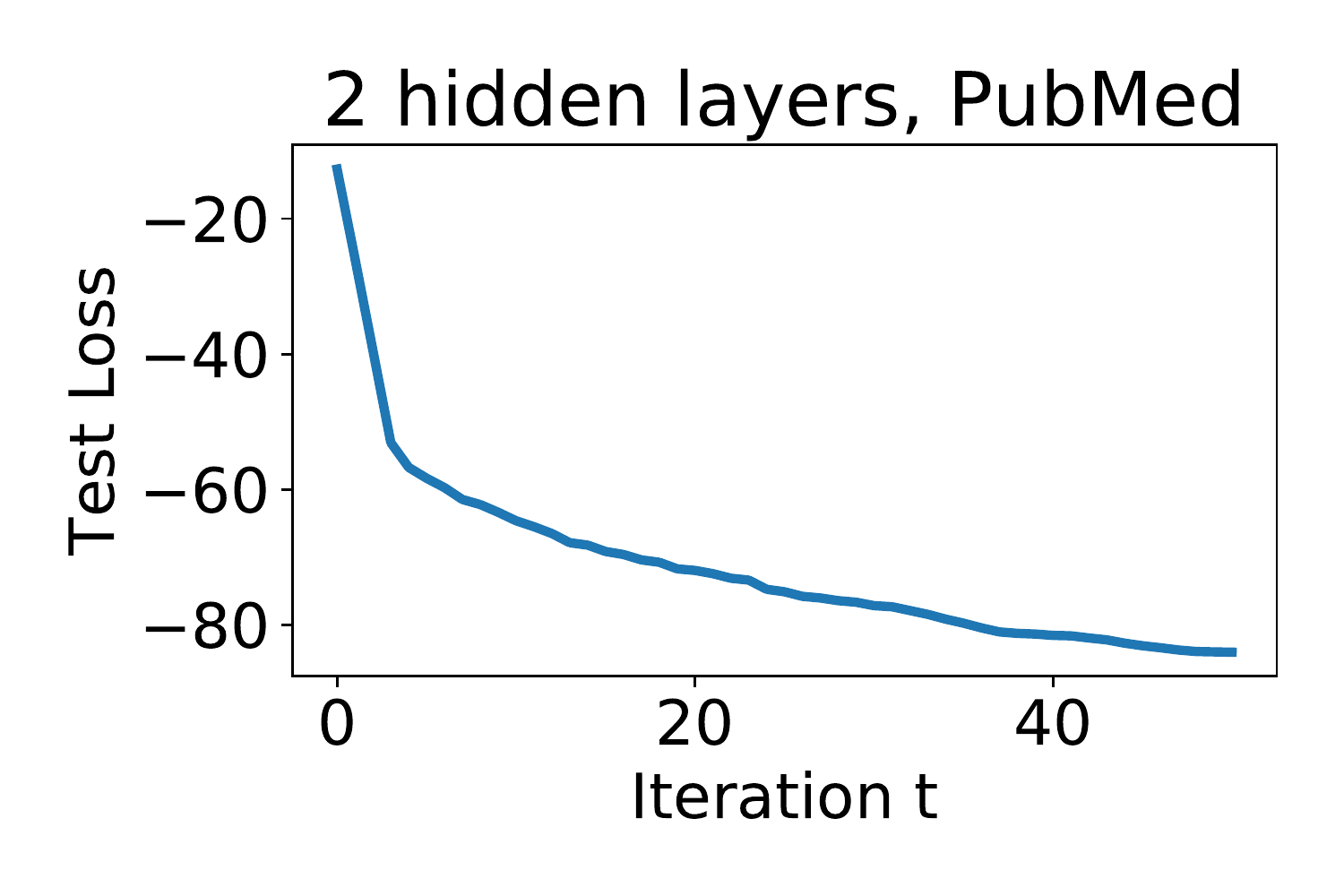}\\
  \includegraphics[width=0.33\linewidth]{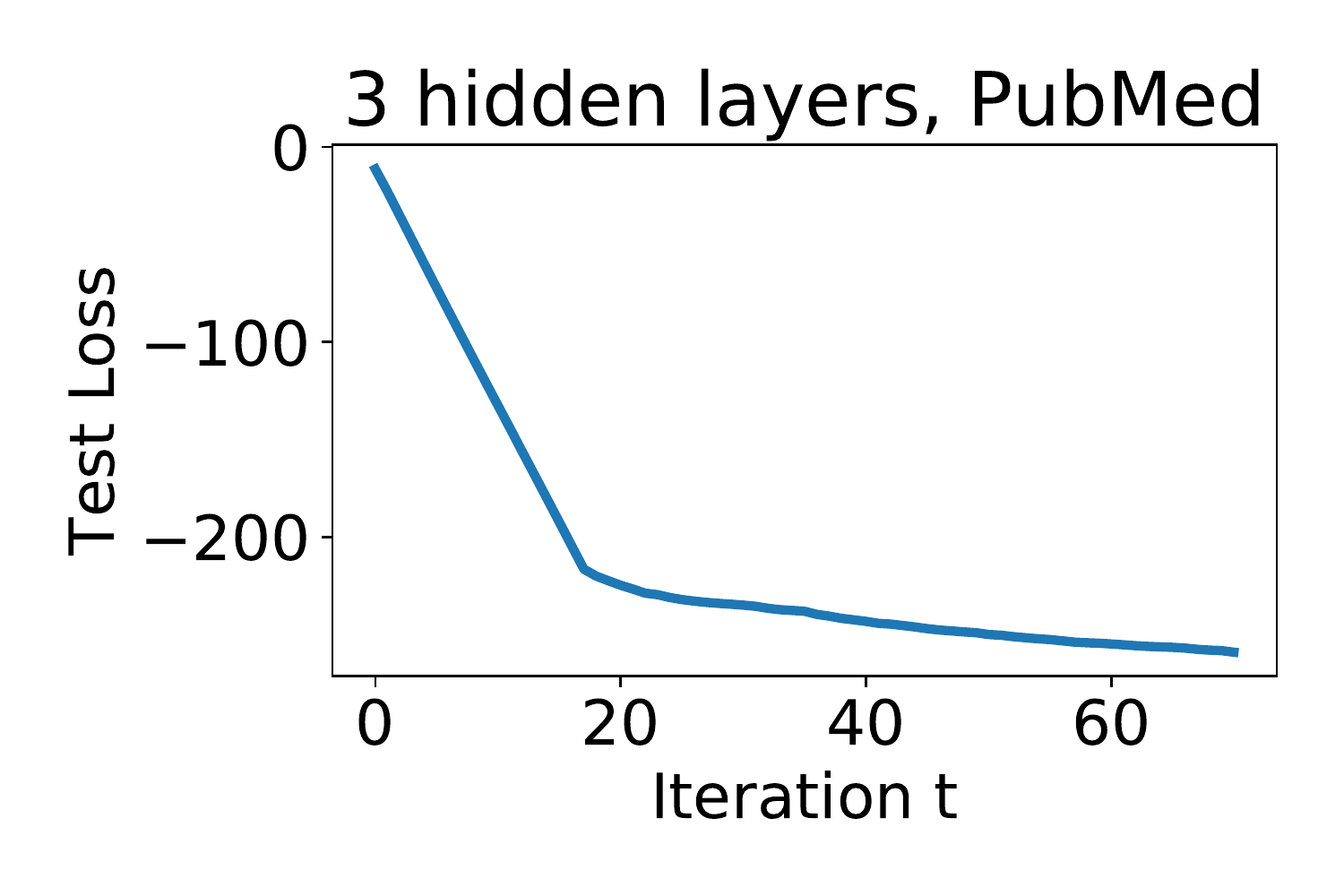}
  \includegraphics[width=0.33\linewidth]{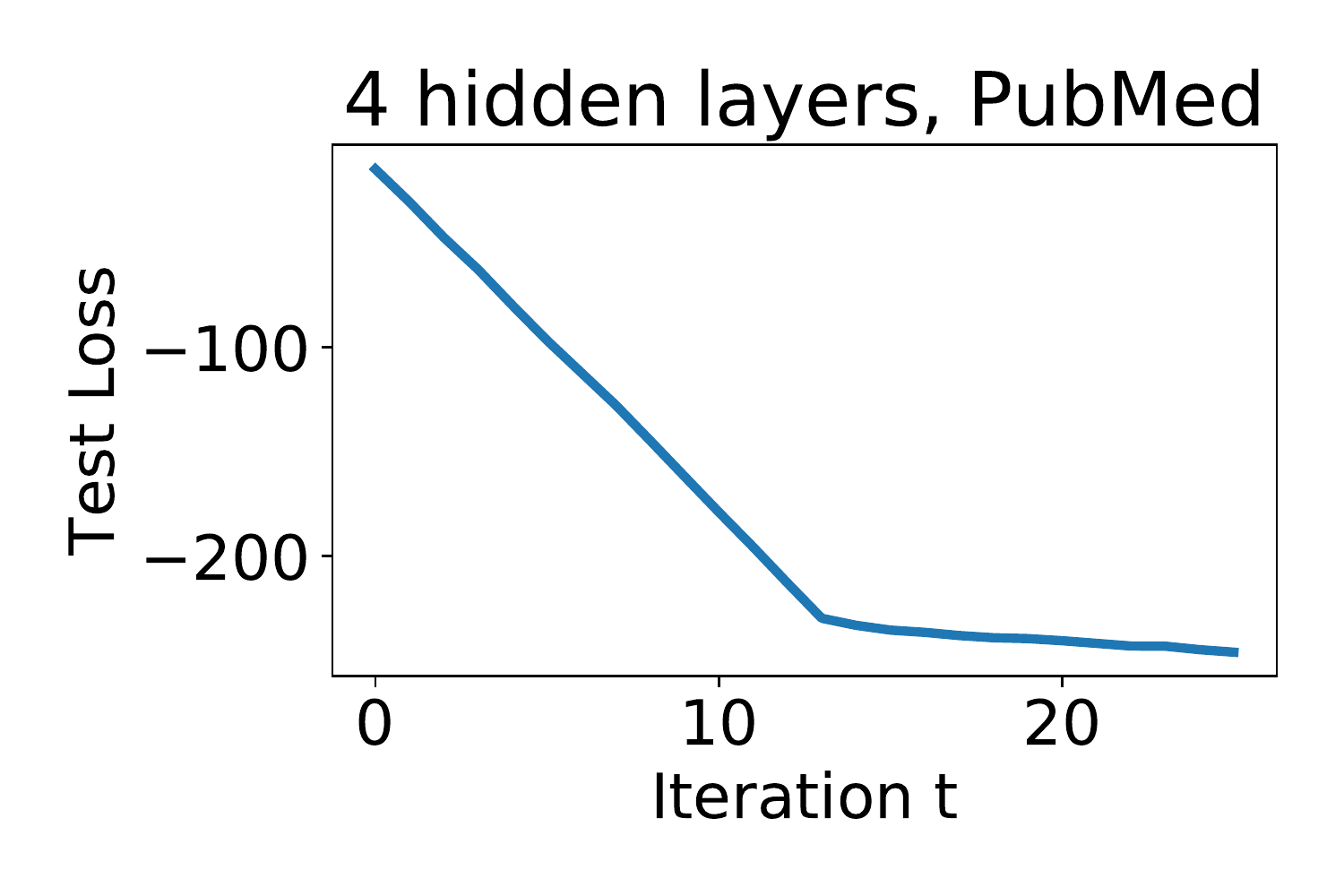}
  \caption{Test loss transition for the PubMed dataset.}\label{fig:test-loss-all-pubmed}
\end{figure}

%% file: image/cosine_all.tex
\begin{figure}[p]
  \includegraphics[width=0.33\linewidth]{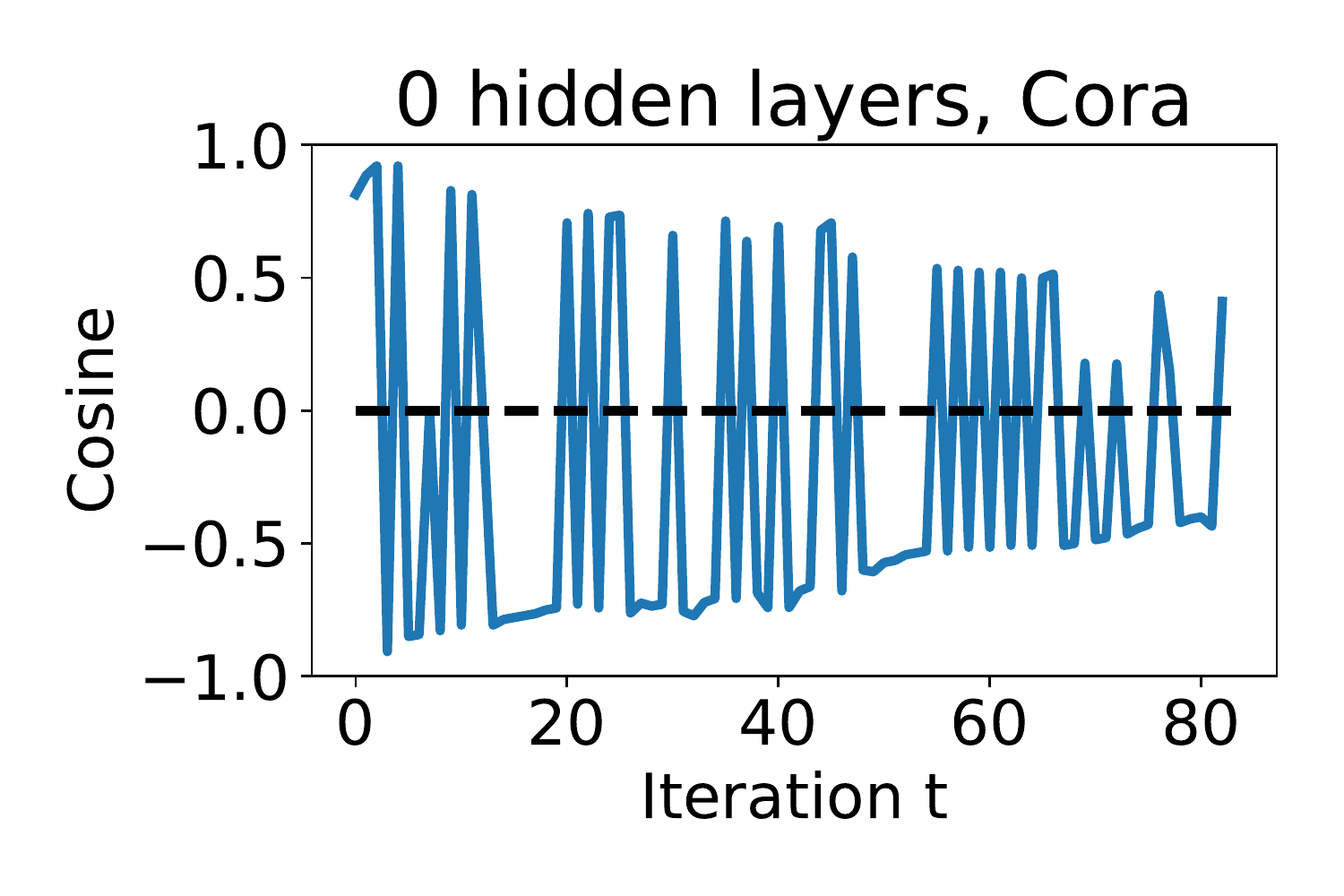}
  \includegraphics[width=0.33\linewidth]{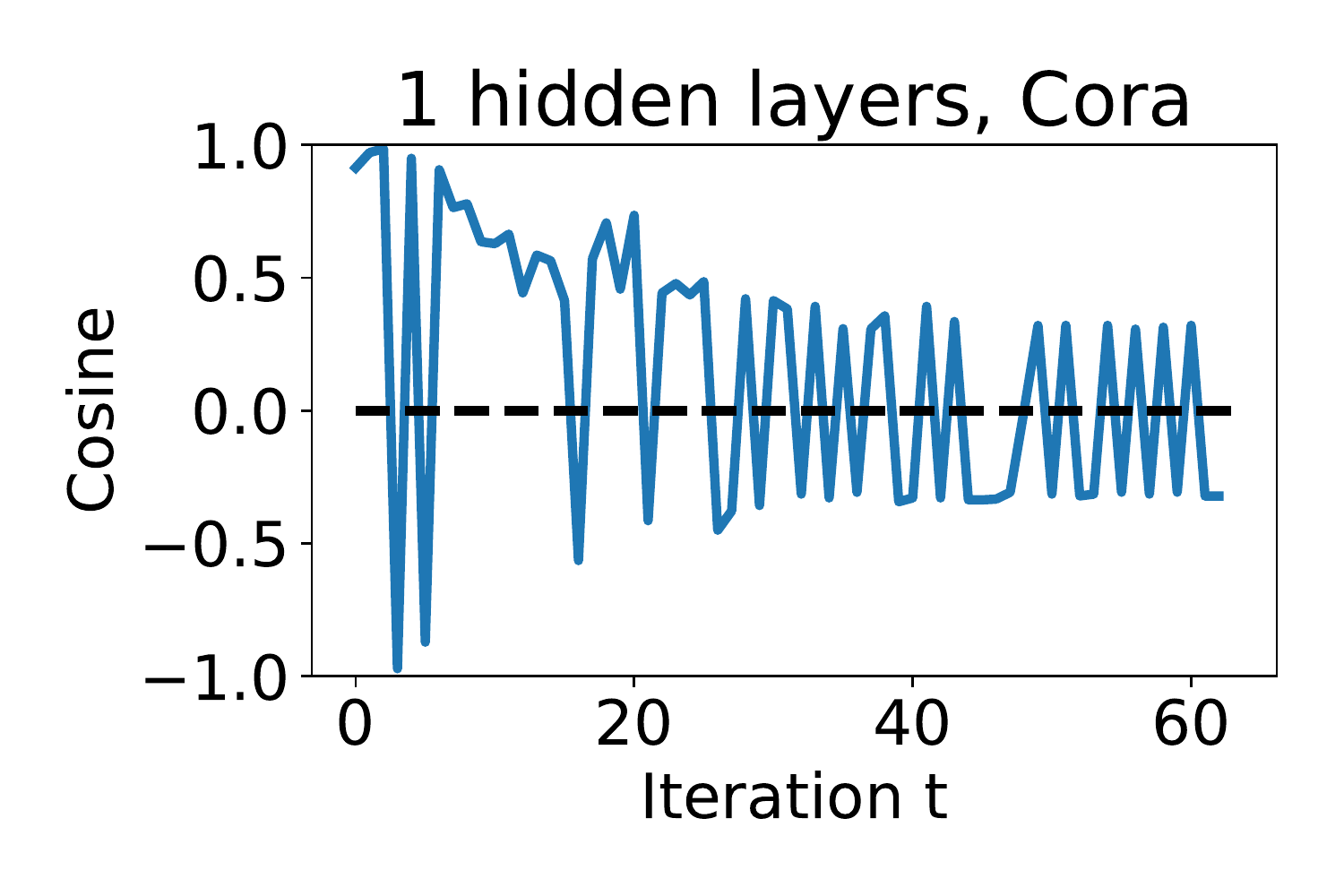}
  \includegraphics[width=0.33\linewidth]{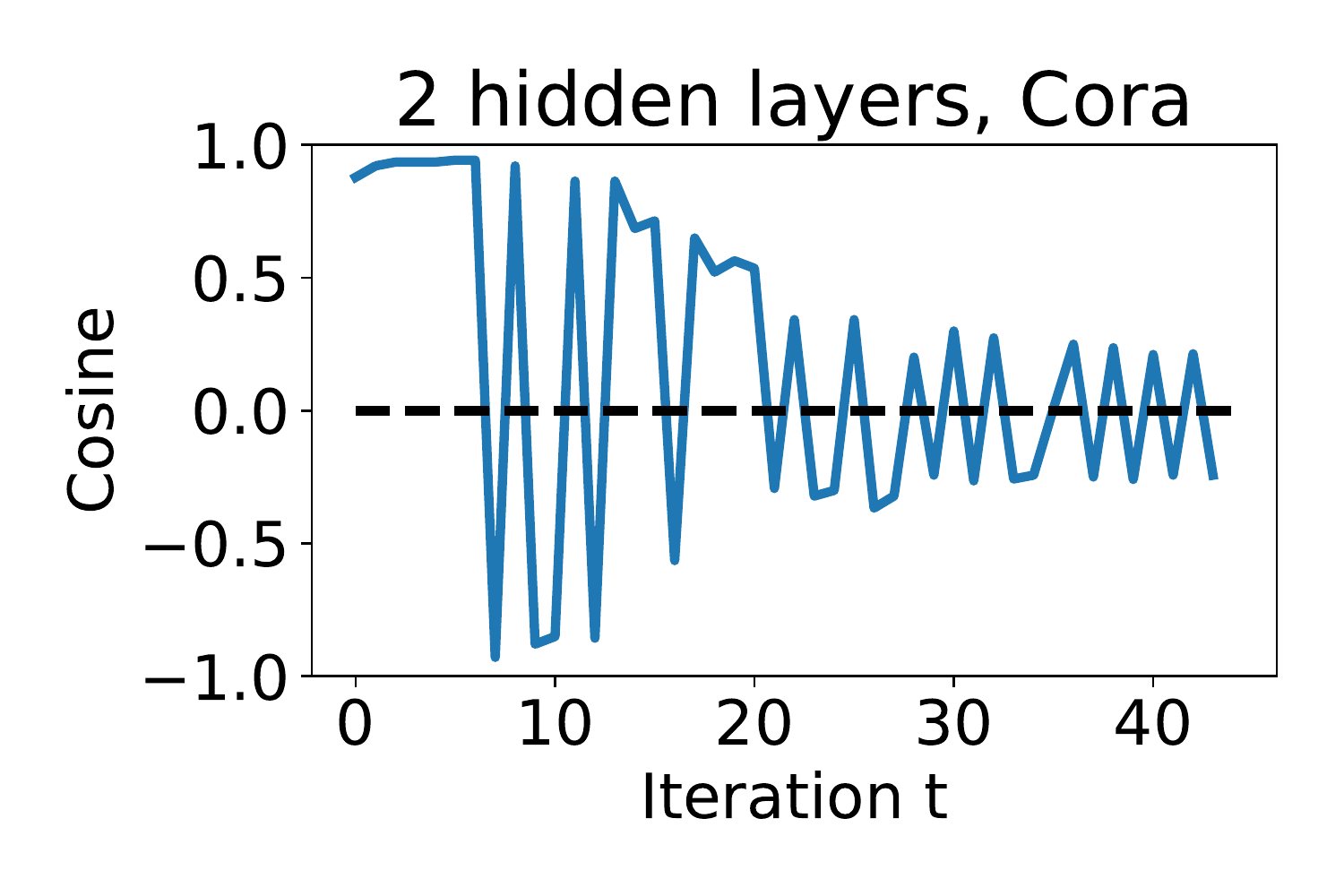}\\
  \includegraphics[width=0.33\linewidth]{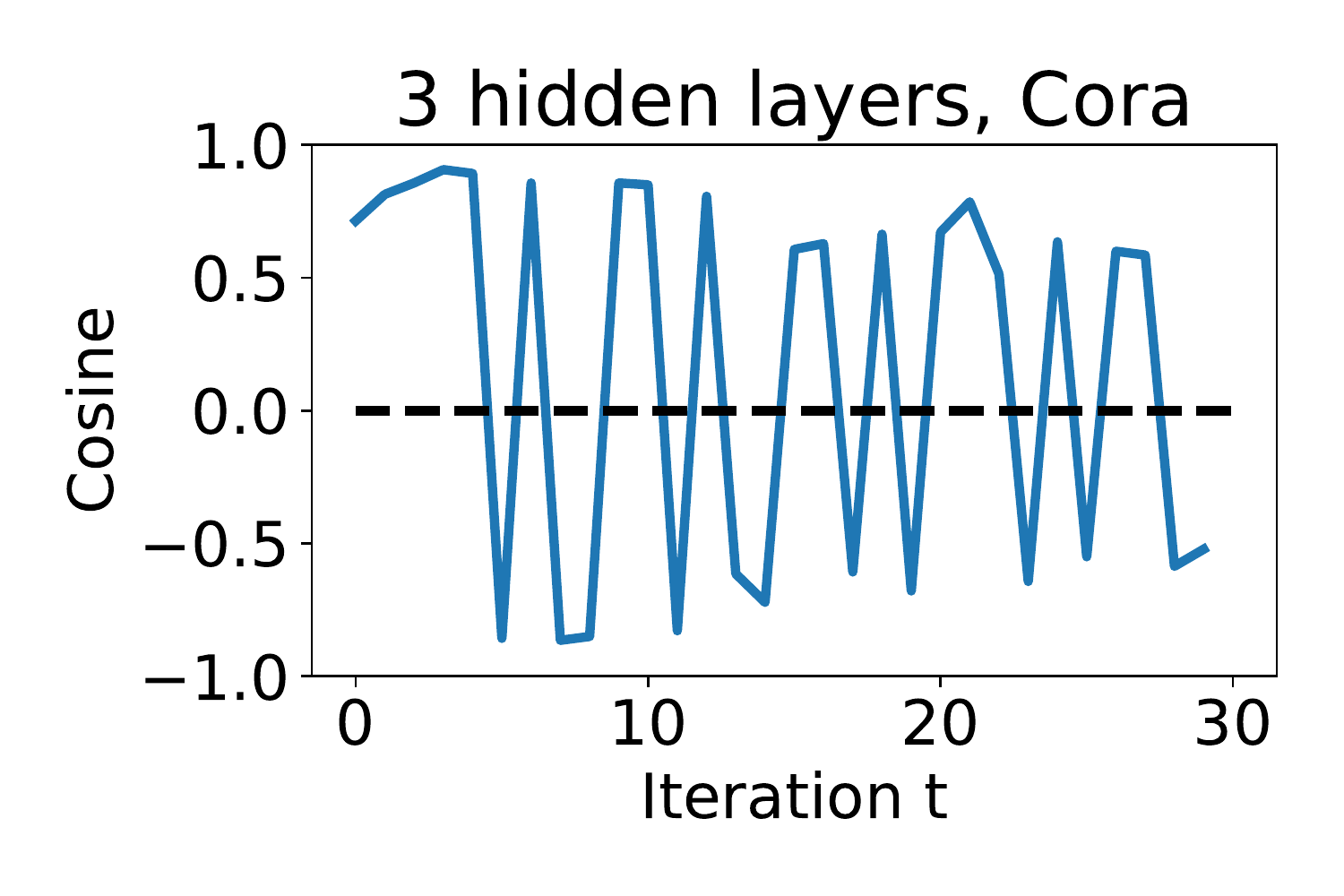}
  \includegraphics[width=0.33\linewidth]{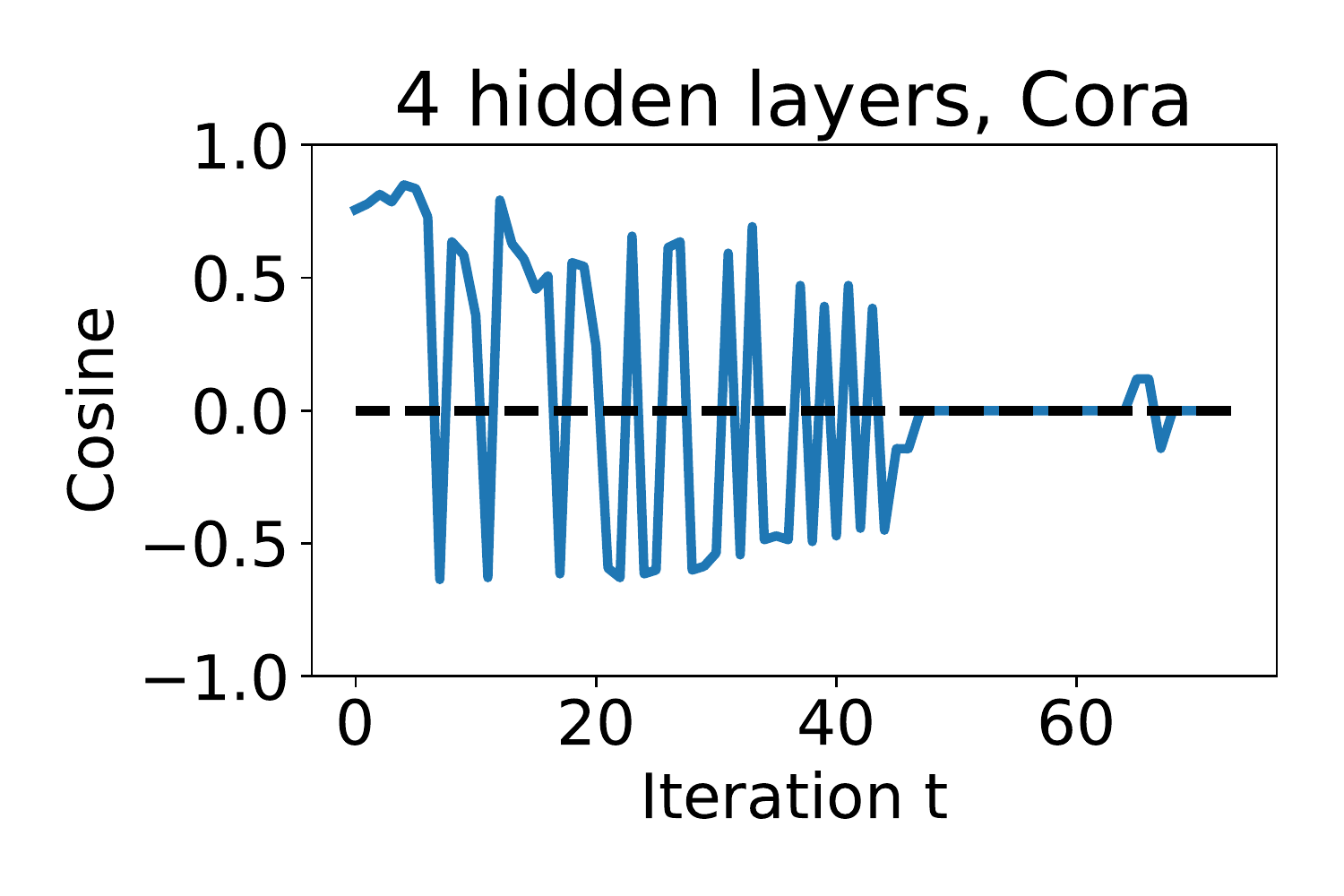}
  \caption{The similarity $\cos \theta^{(t)}$ between weak learners $f^{(t)}$ and the gradient $\nabla \widehat{\mathcal{L}}$ of the training loss for the Cora dataset.}\label{fig:cosine-all-cora}
\vspace{\baselineskip}
  \includegraphics[width=0.33\linewidth]{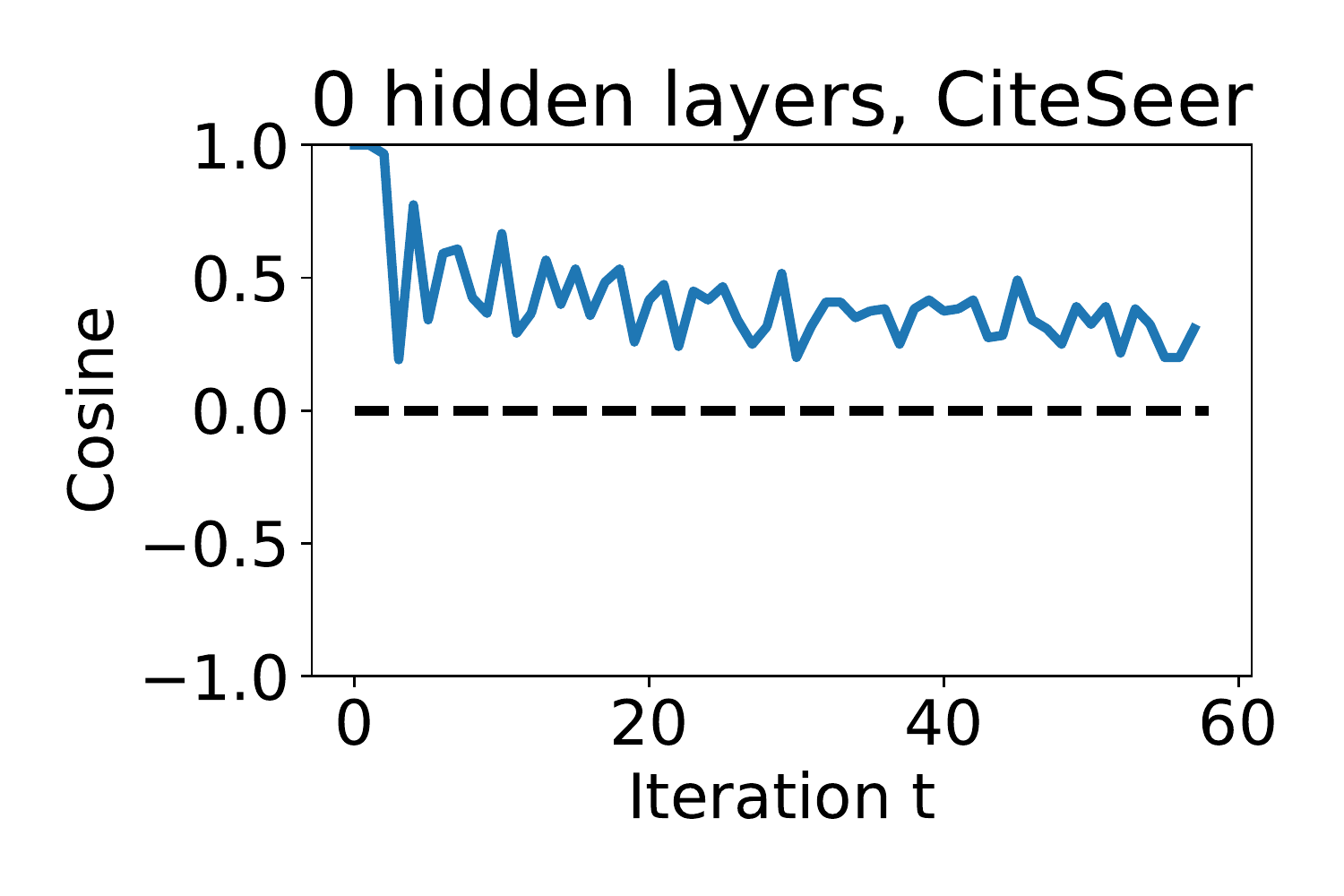}
  \includegraphics[width=0.33\linewidth]{image/cosine/citeseer/1.pdf}
  \includegraphics[width=0.33\linewidth]{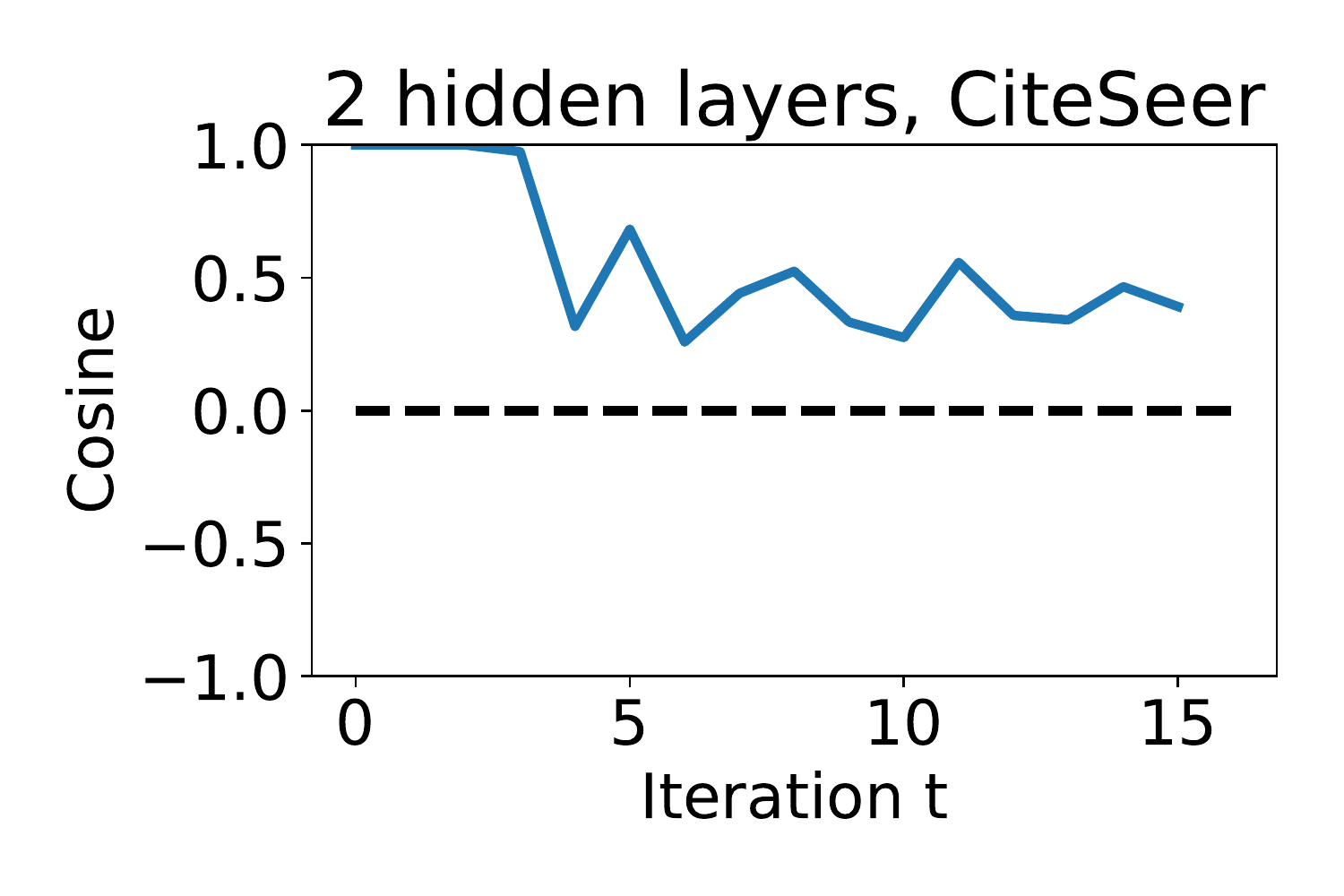}\\
  \includegraphics[width=0.33\linewidth]{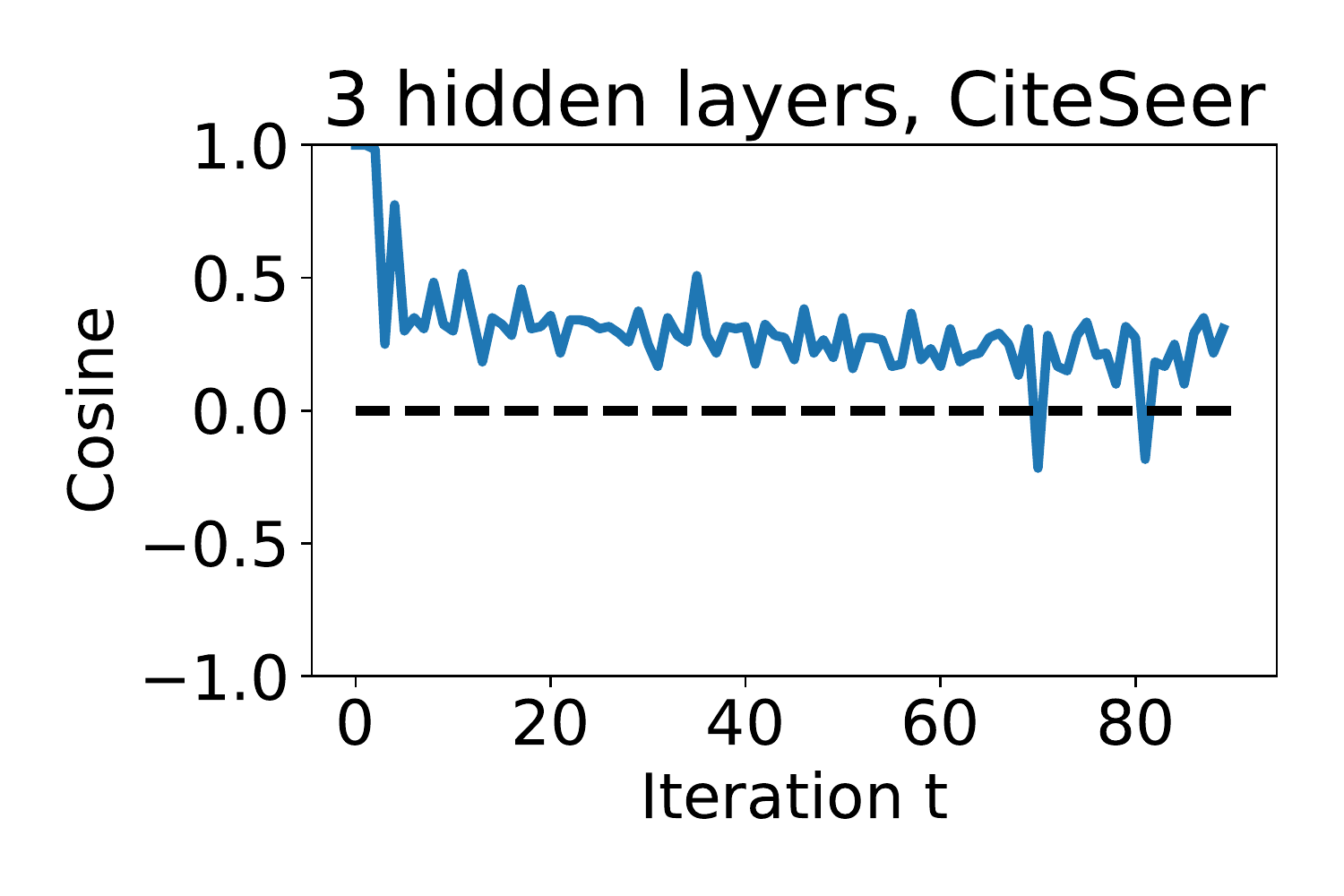}
  \includegraphics[width=0.33\linewidth]{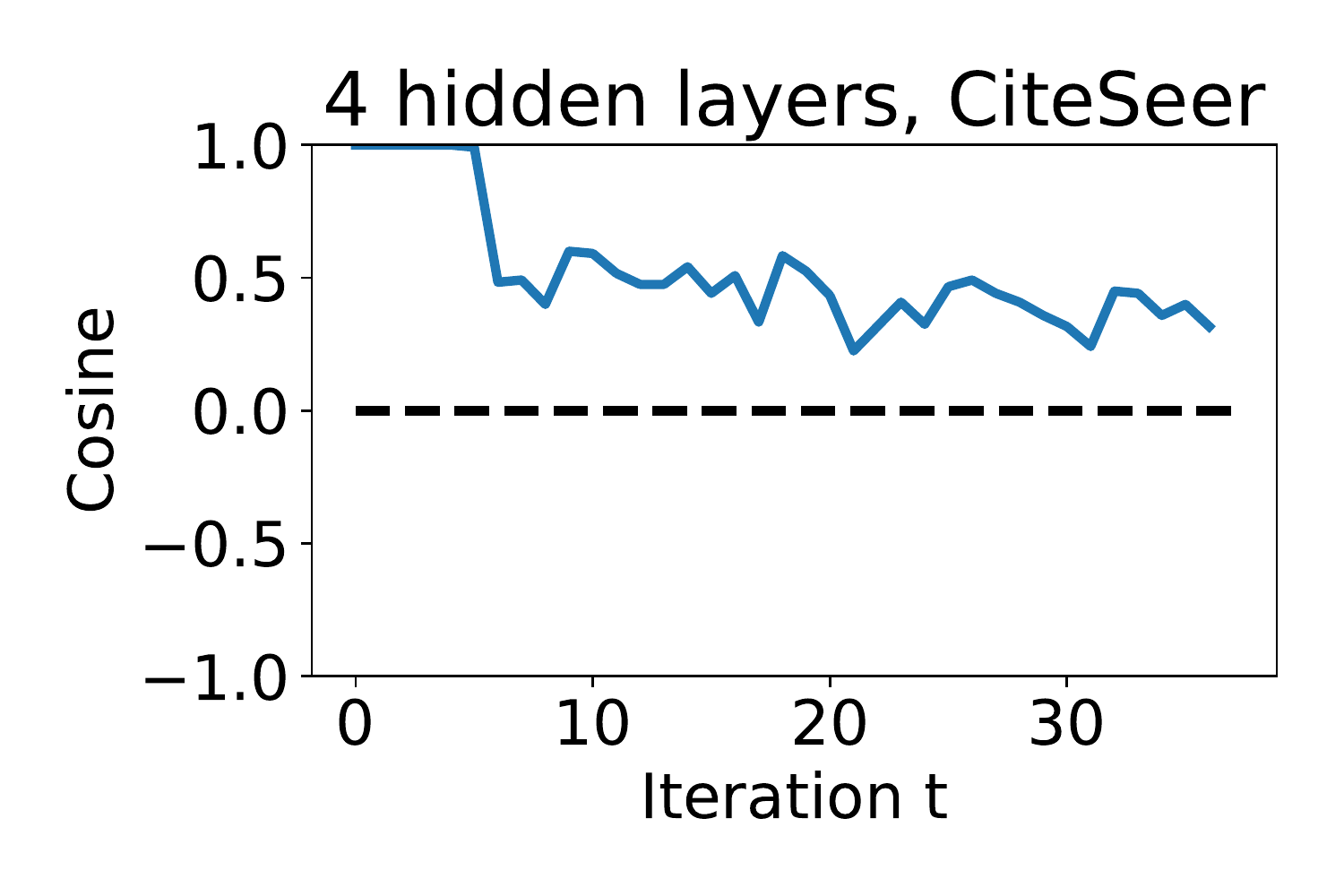}
  \caption{The similarity $\cos \theta^{(t)}$ between weak learners $f^{(t)}$ and the gradient $\nabla \widehat{\mathcal{L}}$ of the training loss for the CiteSeer dataset.}\label{fig:cosine-all-citeseer}
\vspace{\baselineskip}
  \includegraphics[width=0.33\linewidth]{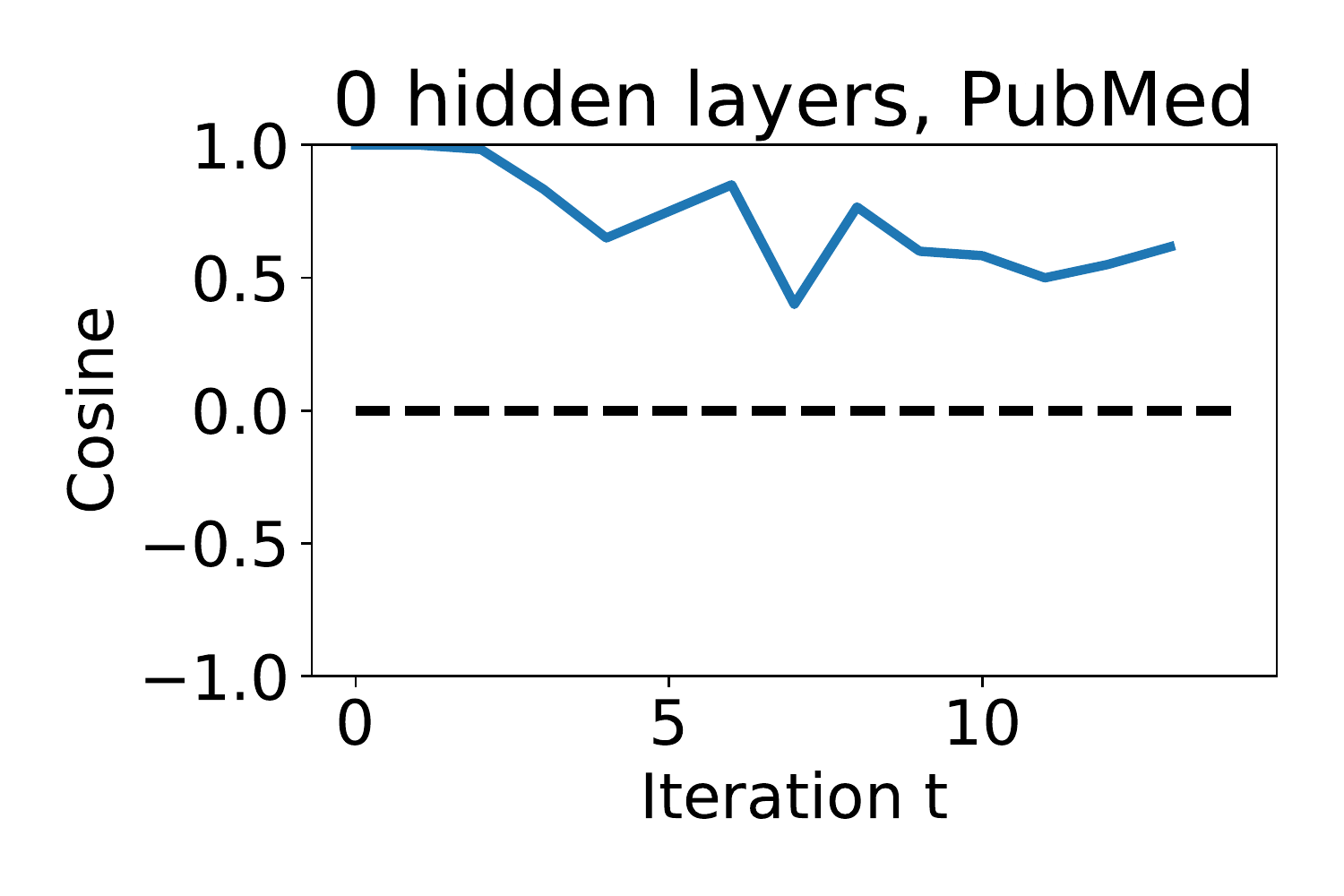}
  \includegraphics[width=0.33\linewidth]{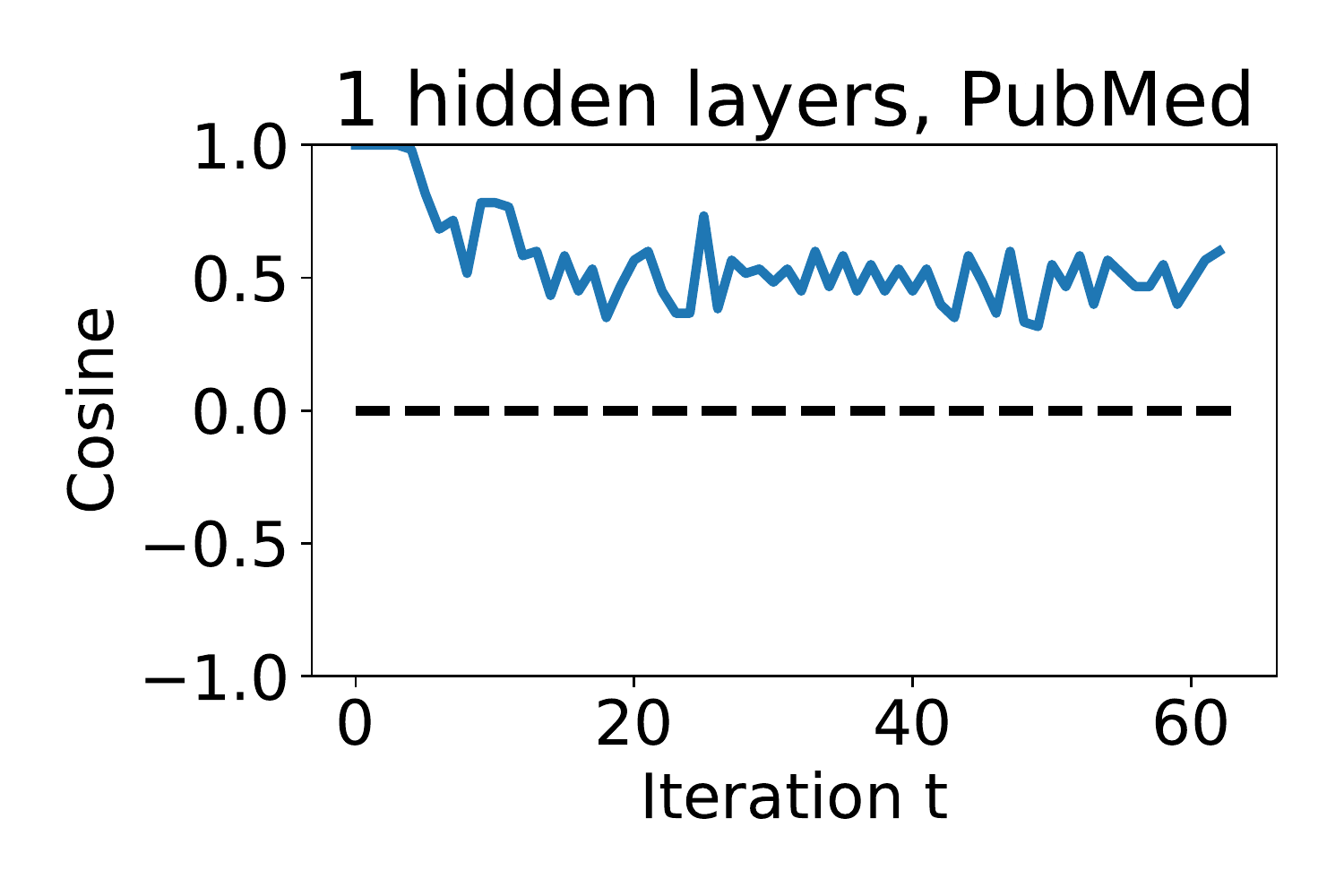}
  \includegraphics[width=0.33\linewidth]{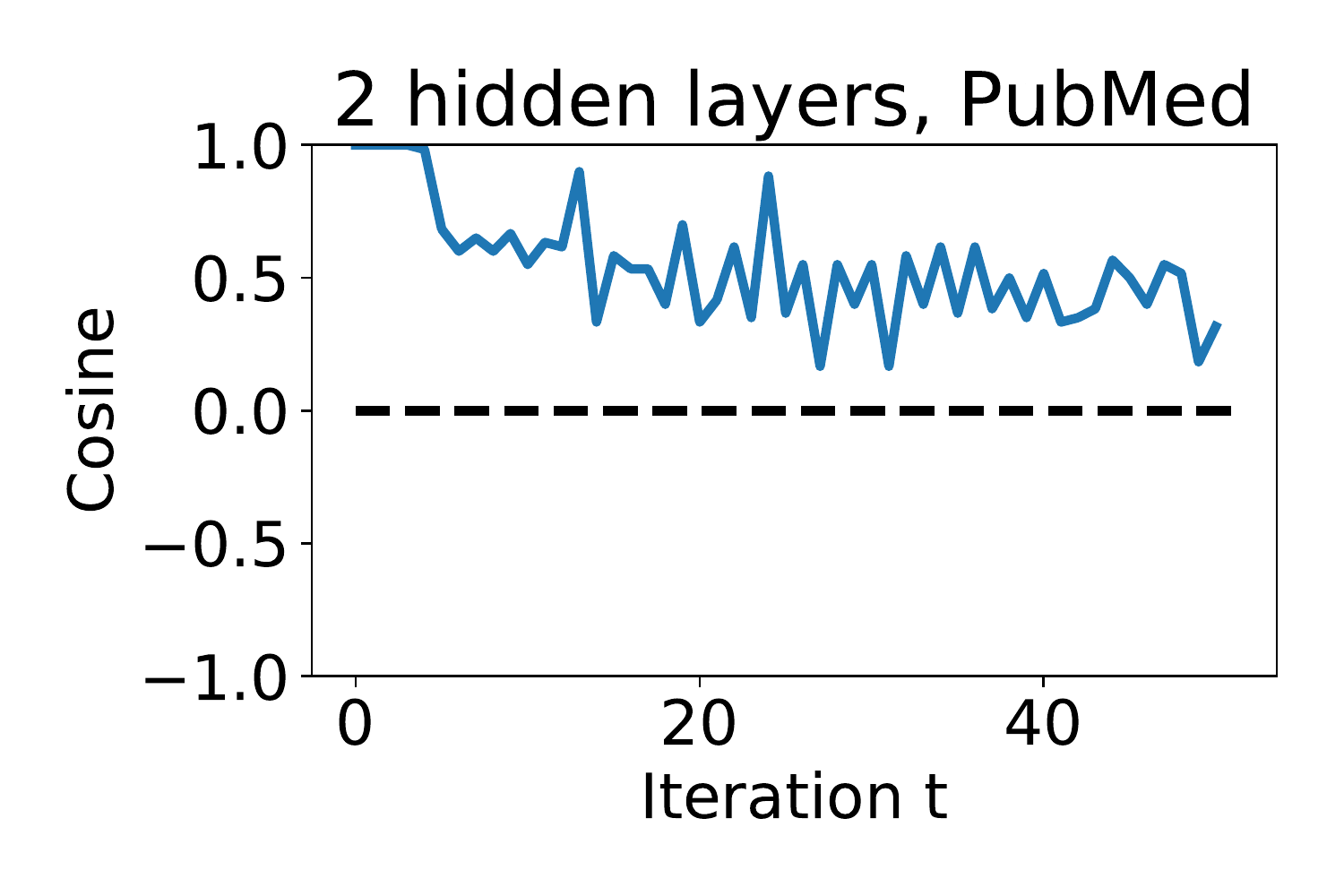}\\
  \includegraphics[width=0.33\linewidth]{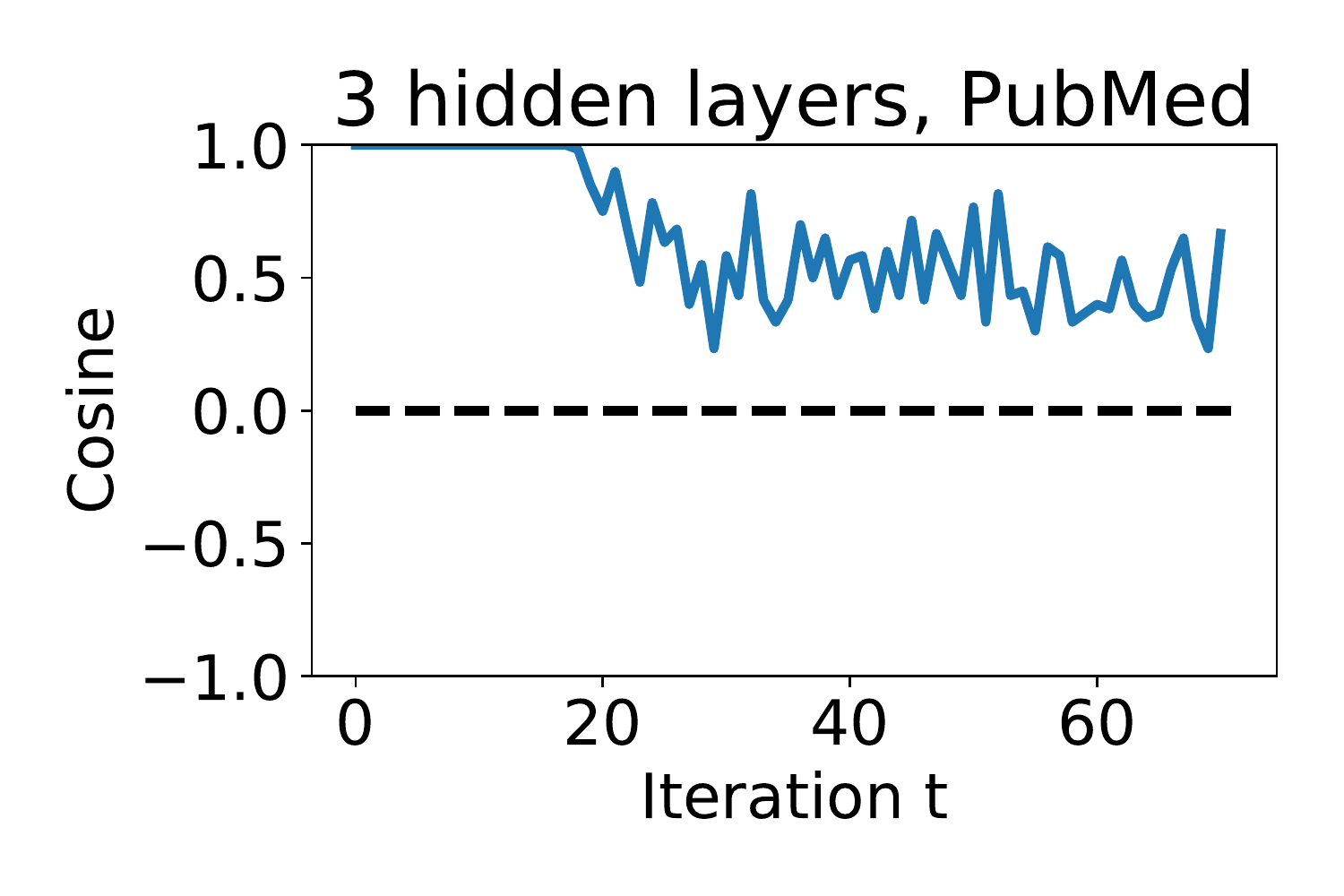}
  \includegraphics[width=0.33\linewidth]{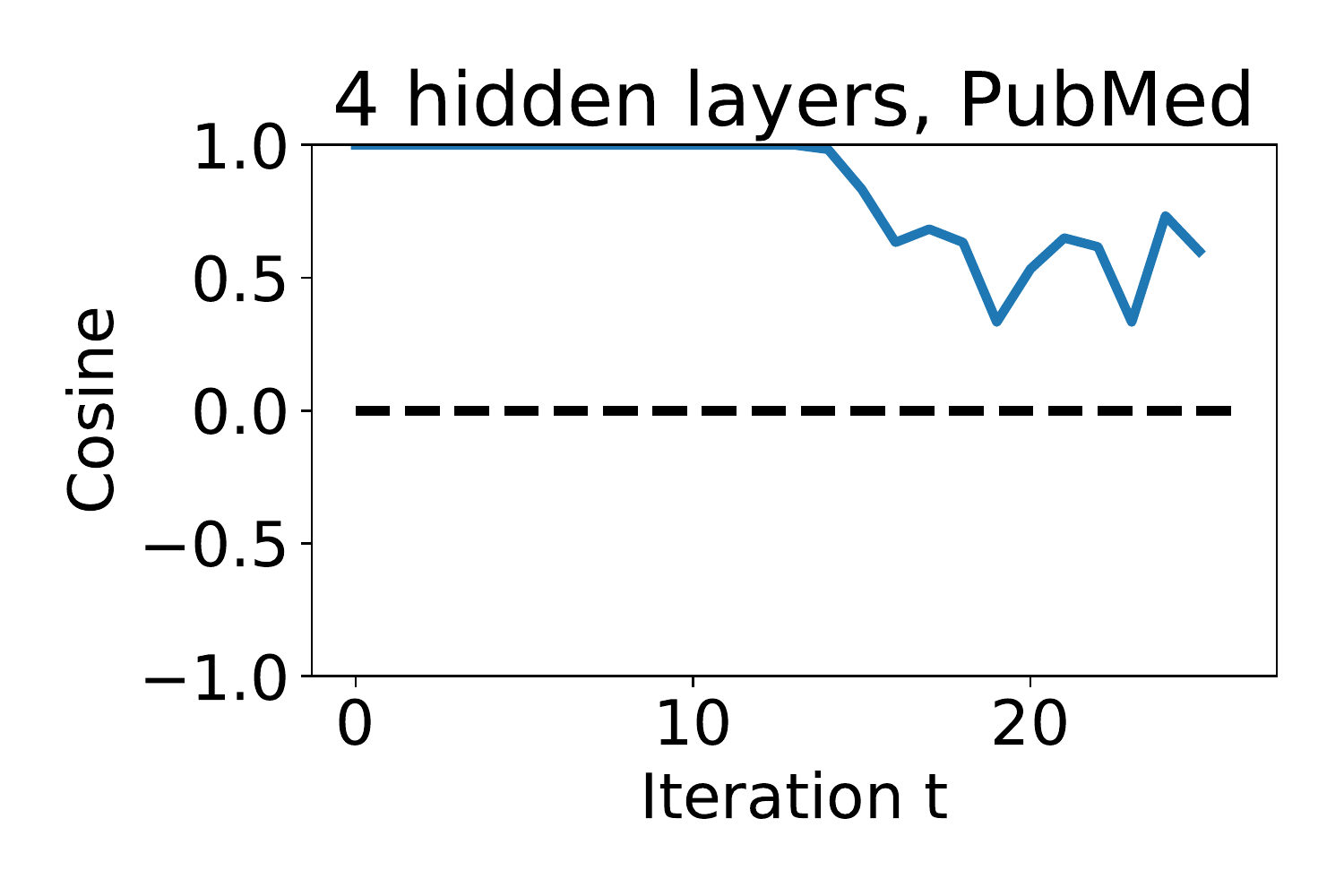}
  \caption{The similarity $\cos \theta^{(t)}$ between weak learners $f^{(t)}$ and the gradient $\nabla \widehat{\mathcal{L}}$ of the training loss for the PubMed dataset.}\label{fig:cosine-all-pubmed}
\end{figure}

%% file: table/citation_for_deep_mlps.tex
\begin{table}
  \caption{Accuracy of node classification tasks on citation networks. $L$ denotes the number of hidden layers. Numbers are $(\textrm{mean})\pm (\textrm{standard deviation})$ of ten runs. $(\ast)$ All runs failed due to GPU memory errors.}
  \label{tbl:node_classification_on_citation_network}
  \centering
  \begin{tabular}{lllll}
    \toprule
        & $L$ & Cora & CiteSeer & PubMed\\
    \midrule
GB-GNN-Adj
& 0 & $79.4 \pm 1.9$ & $70.3 \pm 0.5$ & $78.8 \pm 0.7$ \\
& 1 & $79.9 \pm 0.8$ & $70.5 \pm 0.8$ & $79.4 \pm 0.2$ \\
& 2 & $79.9 \pm 1.3$ & $68.5 \pm 1.3$ & $78.9 \pm 0.6$ \\
& 3 & $77.4 \pm 0.8$ & $64.4 \pm 1.5$ & $78.0 \pm 0.5$ \\
& 4 & $75.6 \pm 2.6$ & $60.7 \pm 1.7$ & $77.9 \pm 0.6$ \\
GB-GNN-Adj. + Fine Tuning
& 1 & $80.4 \pm 0.8$ & $70.8 \pm 0.8$ & $79.0 \pm 0.5$ \\
\midrule
GB-GNN-KTA
& 0 & $80.0 \pm 0.8$ & $70.0 \pm 1.8$ & $79.4 \pm 0.1$ \\
& 1 & $80.9 \pm 0.9$ & $73.1 \pm 1.1$ & $79.1 \pm 0.4$ \\
& 2 & $79.8 \pm 1.3$ & $68.8 \pm 1.1$ & $79.1 \pm 0.4$ \\
& 3 & $78.5 \pm 0.9$ & $65.2 \pm 1.5$ & $78.4 \pm 0.8$ \\
& 4 & $76.0 \pm 2.5$ & $65.6 \pm 1.7$ & $78.0 \pm 0.7$ \\
GB-GNN-KTA + Fine Tuning
& 1 & $82.3 \pm 1.1$ & $70.8 \pm 1.0$ & N.A.$^{(\ast)}$ \\
\midrule
GB-GNN-II
& 0 & $79.2 \pm 1.3$ & $71.4 \pm 0.3$ & $79.3 \pm 0.5$ \\
& 1 & $79.8 \pm 1.3$ & $71.3 \pm 0.5$ & $79.4 \pm 0.3$ \\
& 2 & $79.9 \pm 0.8$ & $69.8 \pm 1.1$ & $79.3 \pm 0.3$ \\
& 3 & $78.7 \pm 1.7$ & $66.7 \pm 1.9$ & $79.2 \pm 0.6$ \\
& 4 & $75.4 \pm 2.0$ & $65.1 \pm 2.5$ & $78.6 \pm 0.7$ \\
GB-GNN-II + Fine Tuning
& 1 & $80.8 \pm 1.3$ & $70.8 \pm 0.9$ & $79.2 \pm 0.8$ \\
\midrule
GB-GNN-SAMME.R
& 0 & $81.0 \pm 0.8$ & $70.4 \pm 0.7$ & $78.9 \pm 0.3$ \\
& 1 & $82.2 \pm 1.2$ & $71.6 \pm 0.5$ & $78.8 \pm 0.3$ \\
& 2 & $80.5 \pm 0.8$ & $67.4 \pm 1.1$ & $78.9 \pm 0.3$ \\
& 3 & $79.6 \pm 1.1$ & $64.4 \pm 1.4$ & $78.8 \pm 0.5$ \\
& 4 & $78.9 \pm 1.8$ & $64.6 \pm 1.3$ & $78.1 \pm 0.6$ \\
GB-GNN-SAMME.R + Fine Tuning
& 1 & $82.1 \pm 1.0$ & $71.3 \pm 0.8$ & $79.4 \pm 0.4$ \\
    \bottomrule
  \end{tabular}
\end{table}

%% file: source/appendix/experiment_additional_results/model_comparison.tex
\subsection{Performance Comparison with Existing GNN Models}\label{sec:model-comparison}

Table \ref{tbl:model-comparison} shows the accuracies of node prediction tasks on citation networks for various GNN models. We borrowed the results of the official repository of Deep Graph Library (GDL)~\cite{wang2019dgl} (\url{https://github.com/dmlc/dgl}), a package for deep learning on graphs.

\input{table/model_comparison}

%% file: table/model_comparison.tex
\begin{table}
  \caption{Comparison of accuracy of GNN models. Created from the official repository of DGL as of May 23rd, 2020. Adj.: Matrix multiplication model $\mathcal{G}_{\tilde{A}}$ by the normalized adjacency matrix $\tilde{A}$. KTA: Kernel target alignment model $\mathcal{G}_{\mathrm{KTA}}$. II: Input injection model $\mathcal{G}_{\mathrm{II}}$. FT: Fine Tuning. Paper: accuracies are cited from the paper in the Ref. column. DGL: accuracies are cited from the official implementation of DGL. $(\ast)$ Not available due to GPU memory errors. $(\ast\ast)$ Not available from the DGL repository.}
  \label{tbl:model-comparison}
  \centering
  \begin{tabular}{llllllll}
    \toprule
Model&&Ref.&Source&Framework&Cora&Citeseer&Pubmed\\
\midrule
GB-GNN&Adj&--&--&PyTorch&79.9&70.5&79.4\\
&Adj + FT&&&&80.4&70.7&79.0\\
&KTA&&&&80.9&73.1&79.4\\
&KTA + FT&&&&82.3&70.8&N.A.${}^{(\ast)}$\\
&II&&&&79.8&71.4&79.4\\
&II + FT&&&&80.8&70.8&79.2\\
&SAMME.R&&&&82.2&71.6&78.8\\
&SAMME.R + FT&&&&82.1&71.3&79.4\\
\midrule
SGC&&\cite{wu2019simplifying}&Paper&--&83.0&72.5&79.0\\
&&&DGL&PyTorch&84.2&70.9&78.5\\
GCN&&\cite{kipf2017iclr}&Paper&--&81.5&70.3&79.0\\
&&&DGL&PyTorch&81.0&70.2&78.0\\
&&&&TensorFlow&81.0&70.7&79.2\\
TAGCN&&\cite{du2017topology}&Paper&--&83.3&71.4&79.4\\
&&&DGL&PyTorch&81.2&71.5&79.4\\
&&&&MXNet&82.0&70.2&79.8\\
DGI&&\cite{velickovic2018deep}&Paper&--&82.3&71.8&76.8\\
&&&DGL&PyTorch&81.6&69.4&76.1\\
&&&&TensorFlow&81.6&70.2&77.2\\
GraphSAGE&&\cite{NIPS2017_6703}&DGL&PyTorch&83.3&71.1&78.3\\
&&&&MXNet&81.7&69.9&79.0\\
APPNP&&\cite{klicpera2018combining}&Paper&--&85.0&75.7&79.7\\
&&&DGL&PyTorch&83.7&71.5&79.3\\
&&&&MXNet&83.7&71.3&79.8\\
GAT&&\cite{velickovic2018graph}&Paper&--&83.0&72.5&79.0\\
&&&DGL&PyTorch&84.0&70.9&78.6\\
&&&&TensorFlow&84.2&70.9&78.5\\
MoNet&&\cite{Monti_2017_CVPR}&DGL&PyTorch&81.6&N.A.${}^{(\ast\ast)}$&76.3\\
&&&&MXNet&81.4&N.A.${}^{(\ast\ast)}$&74.8\\
    \bottomrule
  \end{tabular}
\end{table}

%% file: main.bbl
\begin{thebibliography}{10}

\bibitem{mixhop}
Sami Abu-El-Haija, Bryan Perozzi, Amol Kapoor, Nazanin Alipourfard, Kristina
  Lerman, Hrayr Harutyunyan, Greg~Ver Steeg, and Aram Galstyan.
\newblock {M}ix{H}op: {H}igher-order graph convolutional architectures via
  sparsified neighborhood mixing.
\newblock In {\em Proceedings of the 36th International Conference on Machine
  Learning (ICML)}, volume~97 of {\em Proceedings of Machine Learning
  Research}, pages 21--29. PMLR, 2019.

\bibitem{ngcn}
Sami Abu-El-Haija, Bryan Perozzi, Amol Kapoor, and Joonseok Lee.
\newblock {N}-{GCN}: {M}ulti-scale graph convolutionfor semi-supervised node
  classification.
\newblock In {\em Conference on Uncertainty in Artificial Intelligence (UAI)},
  2019.

\bibitem{optuna}
Takuya Akiba, Shotaro Sano, Toshihiko Yanase, Takeru Ohta, and Masanori Koyama.
\newblock Optuna: {A} next-generation hyperparameter optimization framework.
\newblock In {\em Proceedings of the 25th ACM SIGKDD International Conference
  on Knowledge Discovery \& Data Mining}, pages 2623--2631. ACM, 2019.

\bibitem{pmlr-v97-arora19a}
Sanjeev Arora, Simon Du, Wei Hu, Zhiyuan Li, and Ruosong Wang.
\newblock Fine-grained analysis of optimization and generalization for
  overparameterized two-layer neural networks.
\newblock In {\em Proceedings of the 36th International Conference on Machine
  Learning (ICML)}, volume~97 of {\em Proceedings of Machine Learning
  Research}, pages 322--332. PMLR, 2019.

\bibitem{pmlr-v80-arora18b}
Sanjeev Arora, Rong Ge, Behnam Neyshabur, and Yi~Zhang.
\newblock Stronger generalization bounds for deep nets via a compression
  approach.
\newblock In {\em Proceedings of the 35th International Conference on Machine
  Learning (ICML)}, volume~80 of {\em Proceedings of Machine Learning
  Research}, pages 254--263. PMLR, 2018.

\bibitem{belkin2004regularization}
Mikhail Belkin, Irina Matveeva, and Partha Niyogi.
\newblock Regularization and semi-supervised learning on large graphs.
\newblock In {\em International Conference on Computational Learning Theory},
  pages 624--638. Springer, 2004.

\bibitem{NIPS2011_4443}
James~S. Bergstra, R\'{e}mi Bardenet, Yoshua Bengio, and Bal\'{a}zs K\'{e}gl.
\newblock Algorithms for hyper-parameter optimization.
\newblock In {\em Advances in Neural Information Processing Systems 24}, pages
  2546--2554. Curran Associates, Inc., 2011.

\bibitem{breiman1998arcing}
Leo Breiman et~al.
\newblock Arcing classifier (with discussion and a rejoinder by the author).
\newblock {\em The Annals of Statistics}, 26(3):801--849, 1998.

\bibitem{busch2020pushnet}
Julian Busch, Jiaxing Pi, and Thomas Seidl.
\newblock {PushNet}: {E}fficient and adaptive neural message passing.
\newblock {\em arXiv preprint arXiv:2003.02228}, 2020.

\bibitem{pmlr-v33-begin14}
Luc Bégin, Pascal Germain, François Laviolette, and Jean-Francis Roy.
\newblock Pac-bayesian theory for transductive learning.
\newblock In {\em Proceedings of the Seventeenth International Conference on
  Artificial Intelligence and Statistics}, volume~33 of {\em Proceedings of
  Machine Learning Research}, pages 105--113. PMLR, 2014.

\bibitem{icml2020_2172}
Ming Chen, Zhewei Wei, Zengfeng Huang, Bolin Ding, and Yaliang Li.
\newblock Simple and deep graph convolutional networks.
\newblock In {\em Proceedings of Machine Learning and Systems 2020}, pages
  3730--3740. 2020.

\bibitem{Chen:2016:XST:2939672.2939785}
Tianqi Chen and Carlos Guestrin.
\newblock {XGBoost}: {A} scalable tree boosting system.
\newblock In {\em Proceedings of the 22nd ACM SIGKDD International Conference
  on Knowledge Discovery and Data Mining}, pages 785--794. ACM, 2016.

\bibitem{chung1997spectral}
Fan~RK Chung and Fan~Chung Graham.
\newblock {\em Spectral graph theory}.
\newblock Number~92 in CBMS Regional Conference Series in Mathematics. American
  Mathematical Soc., 1997.

\bibitem{cortes2008stability}
Corinna Cortes, Mehryar Mohri, Dmitry Pechyony, and Ashish Rastogi.
\newblock Stability of transductive regression algorithms.
\newblock In {\em Proceedings of the 25th international conference on Machine
  learning (ICML)}, pages 176--183, 2008.

\bibitem{JMLR:v13:cortes12a}
Corinna Cortes, Mehryar Mohri, and Afshin Rostamizadeh.
\newblock Algorithms for learning kernels based on centered alignment.
\newblock {\em Journal of Machine Learning Research}, 13(28):795--828, 2012.

\bibitem{NIPS2001_1946}
Nello Cristianini, John Shawe-Taylor, Andr\'{e} Elisseeff, and Jaz~S. Kandola.
\newblock On kernel-target alignment.
\newblock In T.~G. Dietterich, S.~Becker, and Z.~Ghahramani, editors, {\em
  Advances in Neural Information Processing Systems 14}, pages 367--373. MIT
  Press, 2002.

\bibitem{du2017topology}
Jian Du, Shanghang Zhang, Guanhang Wu, Jos{\'e}~MF Moura, and Soummya Kar.
\newblock Topology adaptive graph convolutional networks.
\newblock {\em arXiv preprint arXiv:1710.10370}, 2017.

\bibitem{NIPS2019_8809}
Simon~S Du, Kangcheng Hou, Russ~R Salakhutdinov, Barnabas Poczos, Ruosong Wang,
  and Keyulu Xu.
\newblock Graph neural tangent kernel: {F}using graph neural networks with
  graph kernels.
\newblock In {\em Advances in Neural Information Processing Systems 32}, pages
  5723--5733. Curran Associates, Inc., 2019.

\bibitem{du2018gradient}
Simon~S. Du, Xiyu Zhai, Barnabas Poczos, and Aarti Singh.
\newblock Gradient descent provably optimizes over-parameterized neural
  networks.
\newblock In {\em International Conference on Learning Representations (ICLR)},
  2019.

\bibitem{NIPS2015_5954}
David~K Duvenaud, Dougal Maclaurin, Jorge Iparraguirre, Rafael Bombarell,
  Timothy Hirzel, Alan Aspuru-Guzik, and Ryan~P Adams.
\newblock Convolutional networks on graphs for learning molecular fingerprints.
\newblock In {\em Advances in Neural Information Processing Systems 28}, pages
  2224--2232. Curran Associates, Inc., 2015.

\bibitem{el2006stable}
Ran El-Yaniv and Dmitry Pechyony.
\newblock Stable transductive learning.
\newblock In {\em International Conference on Computational Learning Theory},
  pages 35--49. Springer, 2006.

\bibitem{el2009transductive}
Ran El-Yaniv and Dmitry Pechyony.
\newblock Transductive rademacher complexity and its applications.
\newblock {\em Journal of Artificial Intelligence Research}, 35:193--234, 2009.

\bibitem{freund1995boosting}
Yoav Freund.
\newblock Boosting a weak learning algorithm by majority.
\newblock {\em Information and computation}, 121(2):256--285, 1995.

\bibitem{freund1995desicion}
Yoav Freund and Robert~E Schapire.
\newblock A desicion-theoretic generalization of on-line learning and an
  application to boosting.
\newblock In {\em European conference on computational learning theory}, pages
  23--37. Springer, 1995.

\bibitem{friedman2000additive}
Jerome Friedman, Trevor Hastie, Robert Tibshirani, et~al.
\newblock Additive logistic regression: a statistical view of boosting (with
  discussion and a rejoinder by the authors).
\newblock {\em The Annals of Statistics}, 28(2):337--407, 2000.

\bibitem{friedman2001greedy}
Jerome~H Friedman et~al.
\newblock Greedy function approximation: A gradient boosting machine.
\newblock {\em The Annals of Statistics}, 29(5):1189--1232, 2001.

\bibitem{garg2020generalization}
Vikas~K. Garg, Stefanie Jegelka, and Tommi Jaakkola.
\newblock Generalization and representational limits of graph neural networks.
\newblock {\em arXiv preprint arXiv:2002.06157}, 2020.

\bibitem{giles1998citeseer}
C~Lee Giles, Kurt~D Bollacker, and Steve Lawrence.
\newblock {CiteSeer}: {An} automatic citation indexing system.
\newblock In {\em Proceedings of the third ACM conference on Digital
  libraries}, pages 89--98. ACM, 1998.

\bibitem{gilmer17a}
Justin Gilmer, Samuel~S. Schoenholz, Patrick~F. Riley, Oriol Vinyals, and
  George~E. Dahl.
\newblock Neural message passing for quantum chemistry.
\newblock In {\em Proceedings of the 34th International Conference on Machine
  Learning (ICML)}, volume~70 of {\em Proceedings of Machine Learning
  Research}, pages 1263--1272. PMLR, 2017.

\bibitem{gori2005new}
Marco Gori, Gabriele Monfardini, and Franco Scarselli.
\newblock A new model for learning in graph domains.
\newblock In {\em Proceedings. 2005 IEEE International Joint Conference on
  Neural Networks, 2005.}, volume~2, pages 729--734. IEEE, 2005.

\bibitem{ICML2011Grubb_626}
Alexander Grubb and Drew Bagnell.
\newblock Generalized boosting algorithms for convex optimization.
\newblock In {\em Proceedings of the 28th International Conference on Machine
  Learning (ICML)}, pages 1209--1216, 2011.

\bibitem{SciPyProceedings_11}
Aric~A. Hagberg, Daniel~A. Schult, and Pieter~J. Swart.
\newblock Exploring network structure, dynamics, and function using {NetworkX}.
\newblock In Ga\"el Varoquaux, Travis Vaught, and Jarrod Millman, editors, {\em
  Proceedings of the 7th Python in Science Conference}, pages 11 -- 15,
  Pasadena, CA USA, 2008.

\bibitem{NIPS2017_6703}
Will Hamilton, Zhitao Ying, and Jure Leskovec.
\newblock Inductive representation learning on large graphs.
\newblock In {\em Advances in Neural Information Processing Systems 30}, pages
  1024--1034. Curran Associates, Inc., 2017.

\bibitem{hastie2009multi}
Trevor Hastie, Saharon Rosset, Ji~Zhu, and Hui Zou.
\newblock Multi-class {AdaBoost}.
\newblock {\em Statistics and its Interface}, 2(3):349--360, 2009.

\bibitem{he2016deep}
Kaiming He, Xiangyu Zhang, Shaoqing Ren, and Jian Sun.
\newblock Deep residual learning for image recognition.
\newblock In {\em The IEEE Conference on Computer Vision and Pattern
  Recognition (CVPR)}, pages 770--778, 2016.

\bibitem{pmlr-v80-huang18b}
Furong Huang, Jordan Ash, John Langford, and Robert Schapire.
\newblock Learning deep {R}es{N}et blocks sequentially using boosting theory.
\newblock In {\em Proceedings of the 35th International Conference on Machine
  Learning (ICML)}, volume~80 of {\em Proceedings of Machine Learning
  Research}, pages 2058--2067. PMLR, 2018.

\bibitem{Iscen_2019_CVPR}
Ahmet Iscen, Giorgos Tolias, Yannis Avrithis, and Ondrej Chum.
\newblock Label propagation for deep semi-supervised learning.
\newblock In {\em The IEEE Conference on Computer Vision and Pattern
  Recognition (CVPR)}, pages 5070--5079, 2019.

\bibitem{NIPS2018_8076}
Arthur Jacot, Franck Gabriel, and Clement Hongler.
\newblock Neural tangent kernel: {C}onvergence and generalization in neural
  networks.
\newblock In {\em Advances in Neural Information Processing Systems 31}, pages
  8571--8580. Curran Associates, Inc., 2018.

\bibitem{joachims2003transductive}
Thorsten Joachims.
\newblock Transductive learning via spectral graph partitioning.
\newblock In {\em Proceedings of the 20th International Conference on Machine
  Learning (ICML)}, pages 290--297, 2003.

\bibitem{NIPS2017_6907}
Guolin Ke, Qi~Meng, Thomas Finley, Taifeng Wang, Wei Chen, Weidong Ma, Qiwei
  Ye, and Tie-Yan Liu.
\newblock {LightGBM}: {A} highly efficient gradient boosting decision tree.
\newblock In {\em Advances in Neural Information Processing Systems 30}, pages
  3146--3154. Curran Associates, Inc., 2017.

\bibitem{kingma2014iclr}
Diederik~P Kingma and Jimmy Ba.
\newblock {Adam}: {A} method for stochastic optimization.
\newblock In {\em International Conference on Learning Representations (ICLR)},
  2015.

\bibitem{kipf2017iclr}
Thomas~N. Kipf and Max Welling.
\newblock Semi-supervised classification with graph convolutional networks.
\newblock In {\em International Conference on Learning Representations (ICLR)},
  2017.

\bibitem{klicpera2018combining}
Johannes Klicpera, Aleksandar Bojchevski, and Stephan Günnemann.
\newblock Combining neural networks with personalized pagerank for
  classification on graphs.
\newblock In {\em International Conference on Learning Representations (ICLR)},
  2019.

\bibitem{AAAI1816098}
Qimai Li, Zhichao Han, and Xiao-Ming Wu.
\newblock Deeper insights into graph convolutional networks for semi-supervised
  learning.
\newblock In {\em Thirty-Second AAAI Conference on Artificial Intelligence},
  2018.

\bibitem{liao2018lanczosnet}
Renjie Liao, Zhizhen Zhao, Raquel Urtasun, and Richard Zemel.
\newblock {LanczosNet}: {M}ulti-scale deep graph convolutional networks.
\newblock In {\em International Conference on Learning Representations (ICLR)},
  2019.

\bibitem{NIPS2019_9276}
Sitao Luan, Mingde Zhao, Xiao-Wen Chang, and Doina Precup.
\newblock Break the ceiling: {S}tronger multi-scale deep graph convolutional
  networks.
\newblock In {\em Advances in Neural Information Processing Systems 32}, pages
  10943--10953. Curran Associates, Inc., 2019.

\bibitem{NIPS1999_1766}
Llew Mason, Jonathan Baxter, Peter~L. Bartlett, and Marcus~R. Frean.
\newblock Boosting algorithms as gradient descent.
\newblock In {\em Advances in Neural Information Processing Systems 12}, pages
  512--518. MIT Press, 2000.

\bibitem{mccallum2000automating}
Andrew~Kachites McCallum, Kamal Nigam, Jason Rennie, and Kristie Seymore.
\newblock Automating the construction of internet portals with machine
  learning.
\newblock {\em Information Retrieval}, 3(2):127--163, 2000.

\bibitem{mohri2018foundations}
Mehryar Mohri, Afshin Rostamizadeh, and Ameet Talwalkar.
\newblock {\em Foundations of machine learning}.
\newblock MIT Press, 2018.

\bibitem{Monti_2017_CVPR}
Federico Monti, Davide Boscaini, Jonathan Masci, Emanuele Rodola, Jan Svoboda,
  and Michael~M. Bronstein.
\newblock Geometric deep learning on graphs and manifolds using mixture model
  cnns.
\newblock In {\em The IEEE Conference on Computer Vision and Pattern
  Recognition (CVPR)}, pages 5115--5124, 2017.

\bibitem{nagarajan2018deterministic}
Vaishnavh Nagarajan and Zico Kolter.
\newblock Deterministic {PAC}-bayesian generalization bounds for deep networks
  via generalizing noise-resilience.
\newblock In {\em International Conference on Learning Representations}, 2019.

\bibitem{nguyen2017semi}
Hai Nguyen, Shinichi Maeda, and Kenta Oono.
\newblock Semi-supervised learning of hierarchical representations of molecules
  using neural message passing.
\newblock {\em arXiv preprint arXiv:1711.10168}, 2017.

\bibitem{pmlr-v80-nitanda18a}
Atsushi Nitanda and Taiji Suzuki.
\newblock Functional gradient boosting based on residual network perception.
\newblock In {\em Proceedings of the 35th International Conference on Machine
  Learning (ICML)}, volume~80 of {\em Proceedings of Machine Learning
  Research}, pages 3819--3828. PMLR, 2018.

\bibitem{pmlr-v108-nitanda20a}
Atsushi Nitanda and Taiji Suzuki.
\newblock Functional gradient boosting for learning residual-like networks with
  statistical guarantees.
\newblock In {\em Proceedings of the Twenty Third International Conference on
  Artificial Intelligence and Statistics}, volume 108 of {\em Proceedings of
  Machine Learning Research}, pages 2981--2991. PMLR, 2020.

\bibitem{nt2019revisiting}
Hoang NT and Takanori Maehara.
\newblock Revisiting graph neural networks: All we have is low-pass filters.
\newblock {\em arXiv preprint arXiv:1905.09550}, 2019.

\bibitem{Oono2020Graph}
Kenta Oono and Taiji Suzuki.
\newblock Graph neural networks exponentially lose expressive power for node
  classification.
\newblock In {\em International Conference on Learning Representations (ICLR)},
  2020.

\bibitem{NEURIPS2019_9015}
Adam Paszke, Sam Gross, Francisco Massa, Adam Lerer, James Bradbury, Gregory
  Chanan, Trevor Killeen, Zeming Lin, Natalia Gimelshein, Luca Antiga, Alban
  Desmaison, Andreas Kopf, Edward Yang, Zachary DeVito, Martin Raison, Alykhan
  Tejani, Sasank Chilamkurthy, Benoit Steiner, Lu~Fang, Junjie Bai, and Soumith
  Chintala.
\newblock {PyTorch}: {A}n imperative style, high-performance deep learning
  library.
\newblock In {\em Advances in Neural Information Processing Systems 32}, pages
  8024--8035. Curran Associates, Inc., 2019.

\bibitem{pechyony2009theory}
Dmitry Pechyony and Ran El-Yaniv.
\newblock {\em Theory and Practice of Transductive Learning}.
\newblock PhD thesis, Computer Science Department, Technion, 2009.

\bibitem{NIPS2018_7898}
Liudmila Prokhorenkova, Gleb Gusev, Aleksandr Vorobev, Anna~Veronika Dorogush,
  and Andrey Gulin.
\newblock {CatBoost}: {U}nbiased boosting with categorical features.
\newblock In {\em Advances in Neural Information Processing Systems 31}, pages
  6638--6648. Curran Associates, Inc., 2018.

\bibitem{Rong2020DropEdge:}
Yu~Rong, Wenbing Huang, Tingyang Xu, and Junzhou Huang.
\newblock {DropEdge}: {T}owards deep graph convolutional networks on node
  classification.
\newblock In {\em International Conference on Learning Representations (ICLR)},
  2020.

\bibitem{scarselli2009graph}
Franco Scarselli, Marco Gori, Ah~Chung Tsoi, Markus Hagenbuchner, and Gabriele
  Monfardini.
\newblock The graph neural network model.
\newblock {\em IEEE Transactions on Neural Networks}, 20(1):61--80, 2009.

\bibitem{scarselli2018vapnik}
Franco Scarselli, Ah~Chung Tsoi, and Markus Hagenbuchner.
\newblock The {Vapnik--Chervonenkis} dimension of graph and recursive neural
  networks.
\newblock {\em Neural Networks}, 108:248--259, 2018.

\bibitem{schapire1990strength}
Robert~E Schapire.
\newblock The strength of weak learnability.
\newblock {\em Machine learning}, 5(2):197--227, 1990.

\bibitem{schapire1998boosting}
Robert~E Schapire, Yoav Freund, Peter Bartlett, Wee~Sun Lee, et~al.
\newblock Boosting the margin: A new explanation for the effectiveness of
  voting methods.
\newblock {\em The annals of statistics}, 26(5):1651--1686, 1998.

\bibitem{schapire1999improved}
Robert~E Schapire and Yoram Singer.
\newblock Improved boosting algorithms using confidence-rated predictions.
\newblock {\em Machine learning}, 37(3):297--336, 1999.

\bibitem{schlichtkrull2018modeling}
Michael Schlichtkrull, Thomas~N Kipf, Peter Bloem, Rianne Van Den~Berg, Ivan
  Titov, and Max Welling.
\newblock Modeling relational data with graph convolutional networks.
\newblock In {\em European Semantic Web Conference}, pages 593--607. Springer,
  2018.

\bibitem{sen2008collective}
Prithviraj Sen, Galileo Namata, Mustafa Bilgic, Lise Getoor, Brian Galligher,
  and Tina Eliassi-Rad.
\newblock Collective classification in network data.
\newblock {\em AI magazine}, 29(3):93--93, 2008.

\bibitem{sun2019adagcn}
Ke~Sun, Zhouchen Lin, and Zhanxing Zhu.
\newblock {AdaGCN}: Adaboosting graph convolutional networks into deep models.
\newblock {\em arXiv preprint arXiv:1908.05081}, 2019.

\bibitem{pmlr-v35-tolstikhin14}
Ilya Tolstikhin, Gilles Blanchard, and Marius Kloft.
\newblock Localized complexities for transductive learning.
\newblock In {\em Proceedings of The 27th Conference on Learning Theory},
  volume~35 of {\em Proceedings of Machine Learning Research}, pages 857--884.
  PMLR, 2014.

\bibitem{tolstikhin2015permutational}
Ilya Tolstikhin, Nikita Zhivotovskiy, and Gilles Blanchard.
\newblock Permutational rademacher complexity.
\newblock In {\em International Conference on Algorithmic Learning Theory},
  pages 209--223. Springer, 2015.

\bibitem{vapnik1982estimation}
Vladimir Vapnik.
\newblock {\em Estimation of Dependences Based on Empirical Data: Springer
  Series in Statistics (Springer Series in Statistics)}.
\newblock Springer-Verlag, 1982.

\bibitem{NIPS2016_6556}
Andreas Veit, Michael~J Wilber, and Serge Belongie.
\newblock Residual networks behave like ensembles of relatively shallow
  networks.
\newblock In {\em Advances in Neural Information Processing Systems 29}, pages
  550--558. Curran Associates, Inc., 2016.

\bibitem{velickovic2018graph}
Petar Veličković, Guillem Cucurull, Arantxa Casanova, Adriana Romero, Pietro
  Liò, and Yoshua Bengio.
\newblock Graph attention networks.
\newblock In {\em International Conference on Learning Representations (ICLR)},
  2018.

\bibitem{velickovic2018deep}
Petar Veličković, William Fedus, William~L. Hamilton, Pietro Liò, Yoshua
  Bengio, and R~Devon Hjelm.
\newblock Deep graph infomax.
\newblock In {\em International Conference on Learning Representations (ICLR)},
  2019.

\bibitem{verma2019stability}
Saurabh Verma and Zhi-Li Zhang.
\newblock Stability and generalization of graph convolutional neural networks.
\newblock In {\em Proceedings of the 25th ACM SIGKDD International Conference
  on Knowledge Discovery \& Data Mining}, pages 1539--1548. ACM, 2019.

\bibitem{2020SciPy-NMeth}
Pauli {Virtanen}, Ralf {Gommers}, Travis~E. {Oliphant}, Matt {Haberland}, Tyler
  {Reddy}, David {Cournapeau}, Evgeni {Burovski}, Pearu {Peterson}, Warren
  {Weckesser}, Jonathan {Bright}, St{\'e}fan~J. {van der Walt}, Matthew
  {Brett}, Joshua {Wilson}, K.~{Jarrod Millman}, Nikolay {Mayorov}, Andrew
  R.~J. {Nelson}, Eric {Jones}, Robert {Kern}, Eric {Larson}, CJ~{Carey},
  {\.I}lhan {Polat}, Yu~{Feng}, Eric~W. {Moore}, Jake {Vand erPlas}, Denis
  {Laxalde}, Josef {Perktold}, Robert {Cimrman}, Ian {Henriksen}, E.~A.
  {Quintero}, Charles~R {Harris}, Anne~M. {Archibald}, Ant{\^o}nio~H.
  {Ribeiro}, Fabian {Pedregosa}, Paul {van Mulbregt}, and SciPy 1.~0
  {Contributors}.
\newblock {SciPy 1.0}: Fundamental algorithms for scientific computing in
  {Python}.
\newblock {\em Nature Methods}, 17:261--272, 2020.

\bibitem{wang2019dgl}
Minjie Wang, Lingfan Yu, Da~Zheng, Quan Gan, Yu~Gai, Zihao Ye, Mufei Li,
  Jinjing Zhou, Qi~Huang, Chao Ma, Ziyue Huang, Qipeng Guo, Hao Zhang, Haibin
  Lin, Junbo Zhao, Jinyang Li, Alexander~J Smola, and Zheng Zhang.
\newblock {Deep Graph Library}: {T}owards efficient and scalable deep learning
  on graphs.
\newblock {\em ICLR Workshop on Representation Learning on Graphs and
  Manifolds}, 2019.

\bibitem{NIPS2019_9166}
Colin Wei and Tengyu Ma.
\newblock Data-dependent sample complexity of deep neural networks via
  lipschitz augmentation.
\newblock In {\em Advances in Neural Information Processing Systems 32}, pages
  9725--9736. Curran Associates, Inc., 2019.

\bibitem{wu2019simplifying}
Felix Wu, Tianyi Zhang, Amauri Holanda~de Souza~Jr, Christopher Fifty, Tao Yu,
  and Kilian~Q Weinberger.
\newblock Simplifying graph convolutional networks.
\newblock {\em arXiv preprint arXiv:1902.07153}, 2019.

\bibitem{xu2018how}
Keyulu Xu, Weihua Hu, Jure Leskovec, and Stefanie Jegelka.
\newblock How powerful are graph neural networks?
\newblock In {\em International Conference on Learning Representations (ICLR)},
  2019.

\bibitem{pmlr-v80-xu18c}
Keyulu Xu, Chengtao Li, Yonglong Tian, Tomohiro Sonobe, Ken-ichi Kawarabayashi,
  and Stefanie Jegelka.
\newblock Representation learning on graphs with jumping knowledge networks.
\newblock In {\em Proceedings of the 35th International Conference on Machine
  Learning (ICML)}, volume~80 of {\em Proceedings of Machine Learning
  Research}, pages 5453--5462. PMLR, 2018.

\bibitem{yang2018graph}
Jianwei Yang, Jiasen Lu, Stefan Lee, Dhruv Batra, and Devi Parikh.
\newblock Graph {R}-{CNN} for scene graph generation.
\newblock In {\em Proceedings of the European Conference on Computer Vision
  (ECCV)}, pages 670--685, 2018.

\bibitem{zhang2019gresnet}
Jiawei Zhang.
\newblock {GResNet}: Graph residuals for reviving deep graph neural nets from
  suspended animation.
\newblock {\em arXiv preprint arXiv:1909.05729}, 2019.

\bibitem{zhao2020pairnorm}
Lingxiao Zhao and Leman Akoglu.
\newblock {PairNorm}: {T}ackling oversmoothing in {GNN}s.
\newblock In {\em International Conference on Learning Representations (ICLR)},
  2020.

\bibitem{NIPS2003_2506}
Dengyong Zhou, Olivier Bousquet, Thomas~N. Lal, Jason Weston, and Bernhard
  Sch\"{o}lkopf.
\newblock Learning with local and global consistency.
\newblock In {\em Advances in Neural Information Processing Systems 16}, pages
  321--328. MIT Press, 2004.

\end{thebibliography}
